\documentclass{article}

\usepackage[preprint, nonatbib]{neurips_2025}
\usepackage{enumitem}
\usepackage{subcaption}

\usepackage{comment}
\usepackage[utf8]{inputenc} 
\usepackage[T1]{fontenc}    
\usepackage{hyperref}       
\usepackage{amssymb}
\usepackage{url}            
\usepackage{booktabs}       
\usepackage{amsfonts}       
\usepackage{nicefrac}       
\usepackage{microtype}      
\usepackage{xcolor}         

\usepackage{graphicx}
\usepackage{amsmath}
\usepackage{tabularx}
\usepackage{makecell}
\usepackage{array}
\usepackage{multirow}
\usepackage{ragged2e}
\usepackage{float}
\newcommand\rev[1]{\textcolor{black}{#1}}

\newcolumntype{M}[1]{>{\centering\arraybackslash}m{#1}}

\title{Data-Efficient Time-Dependent PDE Surrogates: Graph Neural Simulators vs. Neural Operators}

\author{%
  Dibyajyoti Nayak \\
  Department of Civil and Systems Engineering\\
  Johns Hopkins University\\
  Baltimore, MD, 21218 \\
  \texttt{dnayak2@jh.edu} \\
  \And
  Somdatta Goswami \\
  Department of Civil and Systems Engineering \\
  Johns Hopkins University\\
  Baltimore, MD, 21218 \\
  \texttt{somdatta@jhu.edu} \\
}

\begin{document}

\maketitle

\begin{abstract}
Developing accurate, data-efficient surrogate models is central to advancing AI for Science. Neural operators (NOs), which approximate mappings between infinite-dimensional function spaces using conventional neural architectures, have gained popularity as surrogates for systems driven by partial differential equations (PDEs). However, their reliance on large datasets and limited ability to generalize in low-data regimes hinder their practical utility. We argue that these limitations arise from their global processing of data, which fails to exploit the local, discretized structure of physical systems. To address this, we propose Graph Neural Simulators (GNS) as a principled surrogate modeling paradigm for time-dependent PDEs. GNS leverages message-passing combined with numerical time-stepping schemes to learn PDE dynamics by modeling the instantaneous time derivatives. This design mimics traditional numerical solvers, enabling stable long-horizon rollouts and strong inductive biases that enhance generalization. We rigorously evaluate GNS on \rev{five} canonical PDE systems: (1) 2D scalar Burgers’, (2) 2D coupled Burgers’, (3) 2D Allen–Cahn, (4) 2D nonlinear shallow-water equations, \rev{and (5) 2D flow past a cylinder}, comparing against state-of-the-art NOs including Deep Operator Network (DeepONet), Fourier Neural Operator (FNO), and \rev{Geometry-Informed Neural Operator (GINO)}. Results demonstrate that GNS is markedly more data-efficient, achieving $<1\%$ relative $L_2$ error using only 3\% of available trajectories, and exhibits dramatically reduced error accumulation over time (82.5\% lower autoregressive error than FNO, 99.9\% lower than DeepONet). To choose the training data, we introduce a PCA combined with KMeans clustering trajectory selection strategy. These findings provide compelling evidence that GNS, with its graph-based locality and solver-inspired design, is the most suitable and scalable surrogate modeling framework for AI-driven scientific discovery. Code and data are available at the project repository: \url{https://github.com/Centrum-IntelliPhysics/GNS_vs_NOs.git}.
\end{abstract}

\section{Introduction}
\label{sec:intro}
Time-dependent partial differential equations (PDEs) are central to the mathematical modeling of a wide range of physical phenomena, from fluid mechanics and material physics to acoustics and geosciences. Classical numerical solvers such as finite difference, finite volume, and finite element methods remain the workhorse for obtaining accurate solutions. However, their computational cost grows rapidly with problem size and complexity. Large-scale simulations can demand millions of core-hours on supercomputers, making them prohibitive in tasks such as design optimization, uncertainty quantification, inverse modeling, or real-time control, which require repeated evaluations under varying conditions. In such scenarios, relying solely on traditional solvers is infeasible.

\begin{figure}[htbp!]
    \centering
    \includegraphics[width=\linewidth]{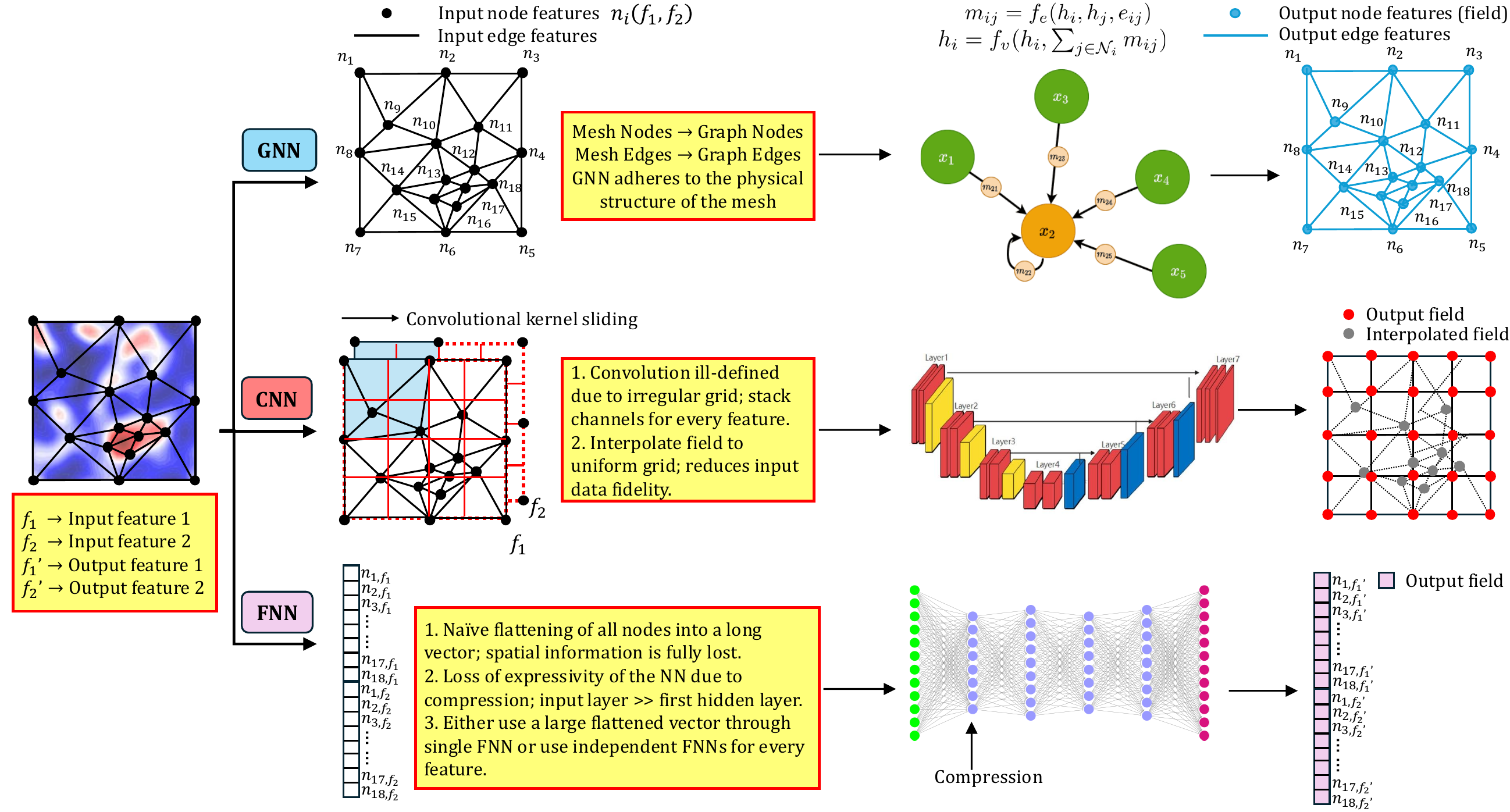}
    \caption{Comparison of neural network architectures for neural operators in scientific applications. Fully connected neural networks (FCNNs) flatten mesh data, discarding spatial structure and compressing high-dimensional inputs into narrow hidden layers, which limits expressivity. Convolutional neural networks (CNNs) require interpolation to uniform grids at both input and output, sacrificing fidelity on unstructured (adaptive) meshes and introducing errors tied to interpolation order. Graph neural networks (GNNs), by contrast, natively represent mesh nodes and edges as graphs, preserving geometry, topology, and physical interactions. Their intrinsic alignment with mesh discretizations, ability to exploit locality, and data efficiency make GNNs the natural choice for Scientific Machine Learning, where irregular geometries and fine discretizations are the norm.}
    \label{fig:GNS_vs_CNN_vs_ANN}
\end{figure}

To address this issue, surrogate modeling has emerged as a powerful alternative. By approximating the behavior of high-fidelity simulations, surrogates can deliver rapid and reasonably accurate solutions at a fraction of the computational cost. Early approaches to surrogate modeling were primarily physics-based reduced-order methods, which achieved success in simpler problem settings but struggled with nonlinear, high-dimensional, or complex systems~\cite{benner2015survey, quarteroni2015reduced, fresca2021comprehensive, fresca2022pod, lu2021review, carlberg2013gnat}. More recently, machine learning approaches, in particular, neural operators (NOs)~\cite{lu2021learning, goswami2023physics, raonic2023convolutional, bonev2023spherical, cao2024laplace, li2020multipole, li2023geometry, hao2023gnot} have gained traction for learning mappings between infinite-dimensional input–output function spaces. Popular architectures among NOs such as the deep operator network (DeepONet)~\cite{lu2021learning} and Fourier neural operator (FNO)~\cite{li2020fourier} offer mesh-independent PDE solvers that are scalable to high-dimensional systems, representing a paradigm shift away from traditional numerical pipelines. While NOs can provide rapid inference across a parametric sweep of input conditions, they often require large amounts of training data and can therefore exhibit poor generalization in data-scarce or limited data regimes. One approach to mitigate this limitation is to incorporate physics-informed strategies \cite{wang2021learning,mandl2025separable,goswami2023physics, mandl2026physics, wang2025time} which explicitly enforce that the network’s predictions satisfy the underlying PDE structure, along with initial and boundary conditions. Such physics-informed neural operators have shown promise, but they come with their own set of challenges. In particular, they require explicit knowledge of the governing PDE and accurate specification of initial and boundary conditions -- information that may not always be available in realistic or complex systems. Moreover, the optimization of physics-informed loss terms often demands careful hyperparameter tuning, making training a non-trivial task~\cite{mcclenny2023self, goswami2025neural}. Furthermore, architectures like FNNs~\cite{hornik1991approximation} or CNNs~\cite{krizhevsky2012imagenet}, which NOs often build upon, exacerbate these issues: fully connected networks treat the domain as flattened vectors, discarding spatial structure, while CNNs assume uniform discretizations and rely on kernels that aggregate information in ways that do not reflect physical locality (see Fig.~\ref{fig:GNS_vs_CNN_vs_ANN}). As a result, such models can fail to capture the inherently local interactions \textit{e.g.}, flux exchanges between neighboring elements that drive PDE-governed dynamics. This mismatch limits their utility in real-world physical applications like fluid flow around complex geometries, deformable materials, phase boundaries, or environmental modeling.

Graph-based learning offers a compelling solution for leveraging local interactions by discretizing computational domains as graphs, where nodes correspond to spatial locations and edges encode local connectivity. This representation naturally handles non-uniform meshes and irregular geometries, while aligning with the local interactions inherent in physical systems (see Fig.~\ref{fig:GNS_vs_CNN_vs_ANN}). Such non-uniform structures are ubiquitous in scientific computing, appearing in mesh-based simulations (finite element and adaptive meshes), particle-based methods (molecular dynamics and smoothed particle hydrodynamics), and point cloud data from geophysical or astronomical observations. In these applications, simulation domains are often defined on unstructured or dynamically deforming meshes, making traditional convolutional architectures poorly suited for direct data processing. Graph Neural Networks (GNNs) have emerged as a powerful framework for learning on graph-structured data through iterative message-passing mechanisms that aggregate information from neighboring nodes, effectively capturing both local dependencies and global structural patterns~\cite{scarselli2008graph, kipf2016semi, velivckovic2017graph, wu2020comprehensive}. This inherent flexibility enables GNNs to model physical systems with adaptive meshes or spatially heterogeneous properties by learning dynamics directly from data, without requiring structured grids or explicit knowledge of governing PDEs. Unlike CNNs or FNNs, which rely on regular data structures, GNNs propagate information through explicit connectivity patterns, allowing them to capture fine-grained local gradients while maintaining sensitivity to long-range dependencies. \rev{This local coupling has a direct algebraic counterpart: discretizing a differential operator with local stencils yields sparse, Laplacian-structured system matrices whose
connectivity reflects the locality of the operator~\cite{conejero2023pre}, and message passing on a graph inherits the same structure, with the graph adjacency encoding this local coupling that GNS extends to a learned, nonlinear update.} This natural alignment with PDE physics makes GNN-based surrogates remarkably data-efficient, capable of learning solution operators from relatively small datasets while generalizing across varying resolutions and scaling effectively to high-dimensional systems.
\begin{figure}[htbp!]
    \centering
    \includegraphics[width=\linewidth, trim={0 1.5cm 0 1.5cm}, clip=True]{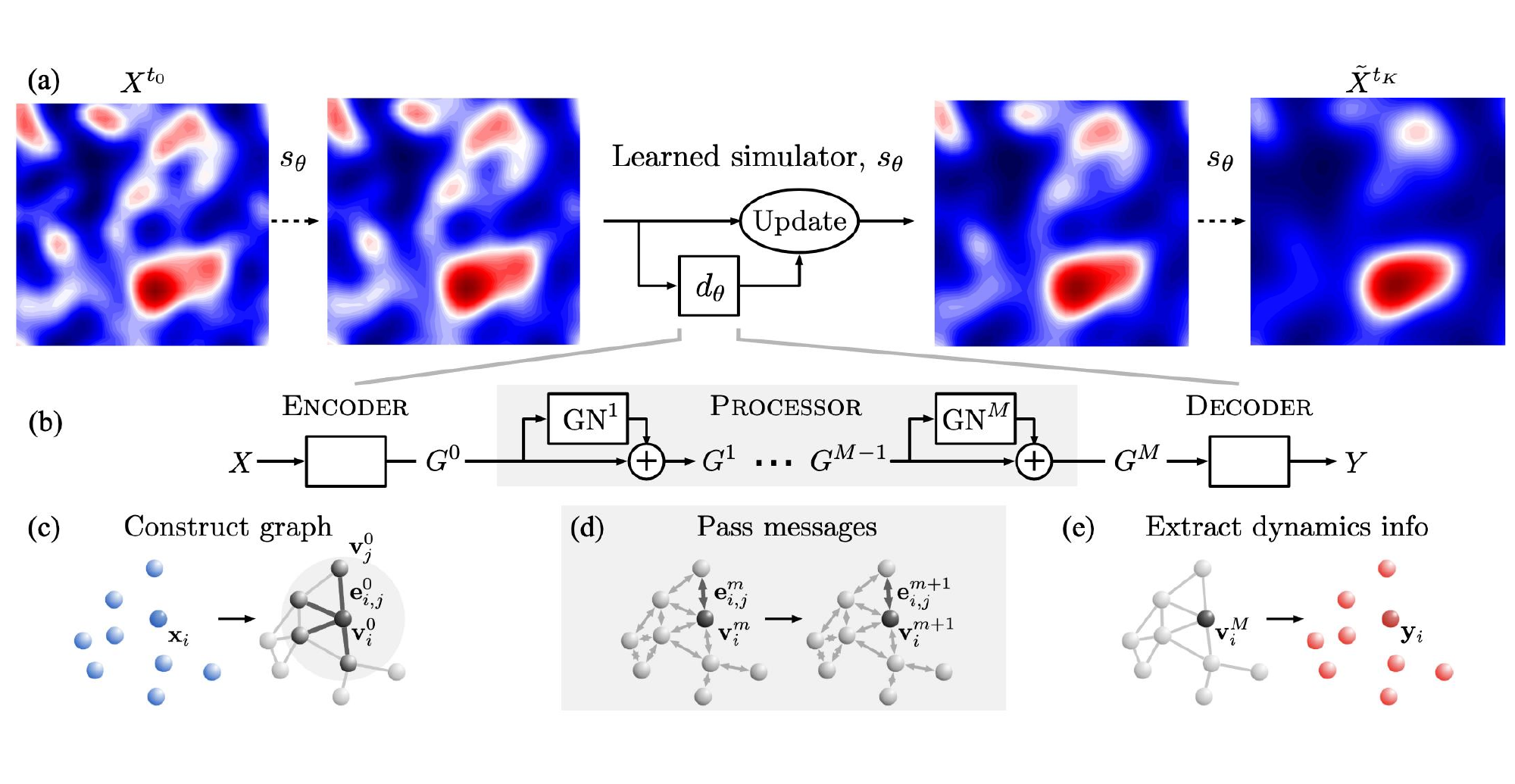}
    \caption{A schematic of the Graph Neural Simulator (GNS) architecture (adapted from \cite{sanchez2020learning}).}
    \label{fig:gns_architecture}
\end{figure}

Building on the flexibility of GNNs, Graph Neural Simulators (GNS) have been proposed as powerful data-driven surrogates for time-dependent physical systems~\cite{sanchez2020learning, pfaff2020learning}. For completeness, a schematic of the GNS architecture, adapted from \cite{sanchez2020learning}, is presented in Fig.~\ref{fig:gns_architecture}. By leveraging message passing~\cite{gilmer2017neural} to encode local temporal updates, GNS models excel in capturing the evolution of PDE-governed systems on irregular domains, particularly in settings with unstructured meshes, particle-based discretizations, or evolving geometries. This locality-driven formulation enables improved generalization across domains, a challenge where NOs often struggle. Unlike NOs that directly learn global solution operators across function spaces, GNS emphasizes local dynamics by propagating information through graph neighborhoods, offering a more physically intuitive representation. Thus, neural operators and GNN-based simulators represent two complementary paradigms: global operator learning versus local dynamics learning, each with distinct strengths and limitations for modeling complex physical phenomena.

Building on recent work that introduced a DeepONet-based surrogate model for accurate temporal extrapolation in time-dependent PDEs~\cite{nayak2026ti, mandl2026physics}, we employ a GNS framework that learns to approximate the instantaneous time derivative of autonomous PDE systems (see Fig.~\ref{fig:gns_architecture}). By reformulating the temporal evolution problem as learning the right-hand side (RHS) function of the PDE, our approach leverages the inherent spatial locality of GNS architectures to model derivative terms while enabling stable long-term integration through established numerical time-stepping schemes. This methodology combines the geometric flexibility of graph-based representations with the temporal stability of classical numerical methods. In this regard, we make the following key contributions:

\begin{itemize}
\item We develop a GNS-based surrogate that serves as an accurate and reliable forward simulator for time-dependent PDEs by learning to model time-derivative fields.
\item We rigorously evaluate the generalization performance of the proposed GNS model against the popular NO-based surrogate baselines in the challenging regime of limited training data.
\item We introduce a systematic training sample selection strategy (specifically for low data regime) that combines principal component analysis (PCA)~\cite{wold1987principal} with KMeans clustering~\cite{hartigan1979algorithm} to ensure sufficient representation of the full dataset through the selected training samples, demonstrating improved performance over pure random sampling approaches.
\end{itemize}
The remainder of the paper is organized as follows. Section~\ref{sec:GNNs} begins with an introduction to graphs in Subsection~\ref{sec:Graphs}, explaining how these data structures provide a more natural representation of physical systems. We then present the mathematical background of the message-passing mechanism in Subsection~\ref{sec:message_passing}, which forms the foundation of all GNN architectures. Building on this foundation, Subsection~\ref{subsec:GNS} details the complete GNS architecture, followed by Subsection~\ref{sec:GNS_sim_PDE}, which describes how we leverage GNS to construct accurate forward models for PDE systems. Finally, we present comprehensive experimental results in Section~\ref{sec:results}, summarize our findings in Section~\ref{sec:conclusions}, and outline limitations and future research directions in Section~\ref{sec:future_work}.

\section{Graph Neural Networks (GNNs)}
\label{sec:GNNs}
Graphs provide a flexible and expressive data structure for representing spatial relationships and interactions in non-Euclidean domains~\cite{bronstein2017geometric, bronstein2021geometric}. By non-Euclidean data, we refer to data that cannot be conveniently embedded into regular, uniformly spaced grids such as those used in standard image processing. Instead, the physical domain is denoted using irregular meshes where connectivity between points is determined by physical relationships, topology, solution gradient or geometry rather than fixed spatial coordinates. In the following subsections, we outline the motivation for using graphs as an alternative data structure, define key terminologies, introduce the mathematical background of message passing, present a thorough overview of Graph Neural Simulators including their encoder, processor, and decoder components, and finally demonstrate how GNS can be leveraged to build accurate, reliable, and stable surrogate models for solving time-dependent PDE systems.

\subsection{Graphs}
\label{sec:Graphs}
A graph $G = (\mathcal{V}, \mathcal{E})$ broadly consists of two entities:
\begin{enumerate}
    \item Vertices (nodes): $\mathbf{v}_i \in \mathcal{V}$, representing individual elements in the system - such as mesh nodes or particles. Each node can store features describing its physical state (\textit{e.g.}, velocity, pressure, position, temperature).
    \item Edges: $\mathbf{e}_{ij} \in \mathcal{E}$ represent relationships or interactions between node pairs $\mathbf{v}_i$ and $\mathbf{v}_j$. Edges may be defined by mesh connectivity, spatial proximity (\textit{e.g.}, $k$-nearest neighbors), or physical constraints, and can store edge features such as relative distances, relative velocities, or interaction coefficients. They may be either directed or undirected, depending on the flow of information between the two nodes of interest.
\end{enumerate}

Graphs offer a \textit{permutation-invariant} representation of data, meaning that the interaction between nodes is independent of their ordering or their absolute positions in the Euclidean space. Furthermore, since graphs encode interactions between any two nodes via edge connections, the \textit{relative} positional information between nodes is often sufficient to model such interactions, rather than requiring absolute coordinates. As a result, graphs provide an efficient and physically consistent framework for modeling interaction behavior in a wide range of systems, particularly those where local or pairwise relationships dictate the dynamics.

In GNNs, learning over the constructed graph is typically performed through a \textit{message-passing} mechanism~\cite{gilmer2017neural}. At each iteration (or layer), nodes exchange information with their neighbors via the edges, aggregating the incoming messages to update their own states. This paradigm naturally captures local interactions, long-range dependencies through multiple hops, and physical invariances such as translational and rotational symmetry. Formally, a GNN takes a graph $G = (\mathcal{V}, \mathcal{E})$ as input, computes the interaction behavior by iteratively updating the properties of its nodes and edges, and returns an updated graph $G' = (\mathcal{V}', \mathcal{E}')$, where the connectivity structure remains unchanged but the node features $\mathbf{v}_i'$ and edge features $\mathbf{e}_{ij}'$ are updated.

\subsection{Message Passing on Graphs}
\label{sec:message_passing}
As mentioned previously, the interaction dynamics between the nodes of a graph are learned through message passing, which provides a principled way of updating node and (optionally) edge feature vectors by incorporating the influence of both immediate and multi-hop neighbors. A single message-passing step generally consists of three stages:
\begin{enumerate}
    \item Message Construction:
    \[
        \mathbf{e}'_{i,j} = \mathbf{m}_{i,j} = f_{\theta}(\mathbf{v}_i, \mathbf{v}_j, \mathbf{e}_{i,j}).
    \]
    \item Message Aggregation (\textit{e.g.}, sum):
    \[
        \bar{\mathbf{v}}_i = \sum_{j \in \mathcal{N}(i)} \mathbf{e}'_{i,j}.
    \]
    \item Node Update:
    \[
        \mathbf{v}'_i = f_{\phi}(\mathbf{v}_i, \bar{\mathbf{v}}_i).
    \]
\end{enumerate}

Here, $f_{\theta}$ and $f_{\phi}$ are functions (typically neural networks) with learnable parameters $\theta$ and $\phi$, respectively. The aggregation operator, $\bar{\mathbf{v}}_i$, can be any permutation-invariant function (\textit{e.g.}, sum, mean, max), ensuring that the result is independent of the ordering of neighbors.

In the message construction step, $f_{\theta}$ takes as input the concatenated feature vectors of the sender node $\mathbf{v}_j$, the receiver node $\mathbf{v}_i$, and the edge connecting them $\mathbf{e}_{i,j}$, producing an updated edge feature vector $\mathbf{e}'_{i,j}$, also referred to as the message $\mathbf{m}_{i,j}$. This operation is performed for all neighbors $j \in \mathcal{N}(i)$ of the target (receiver) node $i$. Next, in the message aggregation step, all incoming messages $\mathbf{e}'_{i,j}$ from the neighbors of node $i$ are combined via a permutation-invariant aggregation function to produce an aggregated message vector $\bar{\mathbf{v}}_i$. Finally, in the node update step, $f_{\phi}$ takes both the current feature vector $\mathbf{v}_i$ and the aggregated message $\bar{\mathbf{v}}_i$ to produce the updated node feature vector $\mathbf{v}'_i$.

At the end of one message-passing step (or one GNN layer), the node and edge features $(\mathbf{v}_i, \mathbf{e}_{i,j})$ are updated to $(\mathbf{v}'_i, \mathbf{e}'_{i,j})$ based on the 1-hop neighborhood. To capture information from more distant nodes (\textit{\textit{i.e.}}, $k$-hop neighbors with $k > 1$), multiple message-passing steps are applied sequentially, allowing the influence of far-reaching interactions to propagate through the graph. Building on the concept of message passing, in the next section we discuss a widely adopted GNN-based architecture for simulating complex physical dynamics: the \emph{Graph Neural Simulator} (GNS)~\cite{sanchez2020learning, pfaff2020learning}.

\subsection{Graph Neural Simulators (GNS)}
\label{subsec:GNS}
At a high level, the GNS framework uses a GNN as the backbone with three main components: (1) Encoder, (2) Processor, and (3) Decoder. The encoder takes the physical state of the system and embeds it into a graph representation by defining nodes according to a chosen connectivity rule and assigning node features that represent physical quantities (\textit{e.g.}, position, velocity, or material properties). These features are then mapped into a latent space via learnable transformations. The processor performs multiple rounds of message passing to update the latent node features, propagating information across the graph and capturing both local and long-range interactions. The decoder takes the updated graph representation and projects the latent features back into the physical space to produce dynamic quantities of interest, such as velocities or accelerations at each node. In the following sections, we describe each of these components in detail along with the creation of inputs to the GNS.

\begin{enumerate}[leftmargin=*]
\item \textbf{Encoder}: 
The encoder maps the physical state of the system into a latent graph representation suitable for message passing.  
Formally, let the physical system be represented as a graph $G = (\mathcal{V}, \mathcal{E})$,
where $\mathcal{V} = \{1, \dots, N\}$ denotes the set of nodes (\textit{e.g.}, grid points, particles) and $\mathcal{E} \subseteq \mathcal{V} \times \mathcal{V}$ denotes the set of edges defined according to a chosen connectivity rule. Note that the edges can be both directed or undirected. 

In structured mesh-based domains, the grid naturally forms a graph with each grid point serving as a node. The edges follow standard finite-difference stencils, such as 4-nearest or 8-nearest neighbors in two-dimensional meshes. For unstructured or irregular domains, connectivity is often determined by a radius-based rule: an edge $(i, j)$ is created if $\| \mathbf{x}_i - \mathbf{x}_j \|_2 \le r_c$, where $r_c$ is a predefined connectivity radius. Alternatively, $k$-nearest neighbor graphs are used, where each node $i$ is connected to its $k$ closest nodes in Euclidean space. Each node $i \in \mathcal{V}$ is associated with a feature vector $\mathbf{v}_i^{(0)} \in \mathbb{R}^{d_v}$, which may include physical quantities of interest (\textit{e.g.}, velocity, temperature), static material properties (\textit{e.g.}, density, elasticity), positional coordinates $\mathbf{x}_i \in \mathbb{R}^d$, and boundary condition flags or positional encodings. Similarly, each edge $(i, j) \in \mathcal{E}$ has a feature vector $\mathbf{e}_{i,j}^{(0)} \in \mathbb{R}^{d_e}$, which can encode relative displacement $\mathbf{x}_j - \mathbf{x}_i$, Euclidean distance $\| \mathbf{x}_j - \mathbf{x}_i \|_2$, or material-specific interaction parameters.

The encoder transforms these raw features into a latent representation through learnable mappings:
\[
\mathbf{h}_i^{(0)} = \hat{f}_{\text{node}}\big( \mathbf{v}_i^{(0)} \big), 
\quad
\mathbf{z}_{i,j}^{(0)} = \hat{f}_{\text{edge}}\big( \mathbf{e}_{i,j}^{(0)} \big),
\]
where $\hat{f}_{\text{node}}$ and $\hat{f}_{\text{edge}}$ are typically neural networks with trainable parameters. Here, the subscripts $i$ and $j$ denote node indices, while the superscript $(0)$ indicates the initial layer (pre-processing stage). The resulting latent graph $\big(\{\mathbf{h}_i^{(0)}\}, \{\mathbf{z}_{i,j}^{(0)}\}\big)$ serves as input to the \emph{Processor} stage, which performs the message passing operations.

\item \textbf{Processor}: 
The processor is the core computational module of the GNS, responsible for propagating information across the graph through multiple rounds of message passing. Given the latent graph from the encoder, $\big(\{\mathbf{h}_i^{(0)}\}, \{\mathbf{z}_{i,j}^{(0)}\}\big)$, the processor updates node and edge features iteratively over $L$ message-passing layers. In each layer $\ell = 0, 1, \dots, L-1$, messages are first constructed for every edge $(i, j) \in \mathcal{E}$ as:
\[
\mathbf{m}_{i,j}^{(\ell)} = f_{\theta}\big(\mathbf{h}_i^{(\ell)}, \mathbf{h}_j^{(\ell)}, \mathbf{z}_{i,j}^{(\ell)}\big),
\]
where $f_{\theta}$ is a neural network with learnable parameters $\theta$ that generates messages based on connected node features and edge attributes. These messages are then aggregated at each receiving node $i$ using a permutation-invariant operator:
\[
\bar{\mathbf{m}}_i^{(\ell)} = \mathrm{AGG}\big(\{\mathbf{m}_{i,j}^{(\ell)} : j \in \mathcal{N}(i)\}\big),
\]
where $\mathcal{N}(i)$ denotes the set of neighbors of node $i$, and $\mathrm{AGG}$ may be a permutation-invariant operation such as sum, mean, or max. The aggregated message is then used to update the node features via a separate node update function:
\[
\mathbf{h}_i^{(\ell+1)} = f_{\phi}\big(\mathbf{h}_i^{(\ell)}, \bar{\mathbf{m}}_i^{(\ell)}\big),
\]
where $f_{\phi}$ is another neural network with learnable parameters $\phi$ that combines the current node state with incoming messages. Note that $\theta$ and $\phi$ represent different sets of trainable parameters for the message generation and node update functions, respectively, allowing each component to learn independent specialized transformations.

If edge features are also updated, a learnable edge update function with parameters $\psi$:
\[
\mathbf{z}_{i,j}^{(\ell+1)} = f_{\psi}\big(\mathbf{h}_i^{(\ell)}, \mathbf{h}_j^{(\ell)}, \mathbf{z}_{i,j}^{(\ell)}\big)
\]
can be applied in parallel. After $L$ such layers, the processor outputs the updated latent graph $\big(\{\mathbf{h}_i^{(L)}\}, \{\mathbf{z}_{i,j}^{(L)}\}\big)$, which encodes both local and long-range interactions.

\item \textbf{Decoder}: 
The decoder maps the processed latent graph representation back to the physical space to produce the target quantities of interest. Given the final node embeddings $\{\mathbf{h}_i^{(L)}\}$ from the processor, the decoder applies a learnable mapping
\[
\mathbf{y}_i = g_{\omega}\big(\mathbf{h}_i^{(L)}\big),
\]
where $g_{\omega}$ is a neural network with parameters $\omega$, and $\mathbf{y}_i \in \mathbb{R}^{d_y}$ represents the predicted physical quantities at node $i$ (\textit{e.g.}, velocity, acceleration, pressure). When the task requires edge-level predictions, an analogous mapping
\[
\mathbf{y}_{i,j} = g_{\eta}\big(\mathbf{z}_{i,j}^{(L)}\big)
\]
can be applied to the final edge embeddings. The decoder thus produces simulation outputs in the desired physical space, completing the encoder--processor--decoder pipeline of the GNS architecture.
\end{enumerate}

\subsection{GNS as a forward simulator for PDEs}
\label{sec:GNS_sim_PDE}
Consider a general time-dependent PDE system governing the evolution of a spatiotemporal field $u(\mathbf{x}, t)$, defined over a temporal domain $t \in [0, T]$ and a spatial domain $\mathbf{x} = [x_1, x_2, \dots, x_n]$, with $\mathbf{x} \in \mathcal{X} \subseteq \mathbb{R}^n$. The dynamics of this system can be expressed as
\begin{equation}
    \frac{\partial u}{\partial t} = \mathcal{G}\big(t, \mathbf{x}, u, u_{\mathbf{x}}, u_{\mathbf{xx}}, u_{\mathbf{xxx}}, \dots \big),
\end{equation}
where $\mathcal{G}$ denotes the PDE operator relating the temporal derivative of the field to its spatial derivatives.

As discussed previously, recent work~\cite{nayak2026ti} introduced a DeepONet-based surrogate model for solving time-dependent PDEs by reformulating the operator learning task to predict instantaneous time-derivative fields conditioned on the current system state. This is equivalent to learning a parametric approximation $\mathcal{G} \approx \mathcal{G}_{\theta}$ for the right-hand side of the PDE. Inspired by this formulation, we adopt a GNS framework to model the time derivative, \emph{i.e.}, the PDE RHS. The motivation is that the GNS architecture inherently captures spatial interactions between discretization points, making it well-suited for approximating the spatial derivative terms appearing in $\mathcal{G}$. In this work, we focus on \emph{autonomous systems}, where the right-hand side does not explicitly depend on time.

Once trained to approximate the instantaneous time derivative, the GNS can be integrated forward in time using any stable numerical time-stepping method. For instance, with the explicit Euler scheme, a one-step update from state $u^t$ to $u^{t+1}$ is given by
\begin{equation}
    u^{t+1} = u^t + \Delta t \cdot \mathrm{GNS}(u^t),
\end{equation}
where $\Delta t$ is the time step size. At inference time, the solution can be evolved from an initial condition by repeatedly applying the numerical scheme using the GNS-predicted time derivative. In this way, the GNS effectively serves as a forward simulator for the PDE system. Note that while we employ the first-order explicit Euler scheme, the same formulation holds for any higher-order numerical integration scheme.

Graph construction, along with the definition of appropriate node and edge feature vectors, forms the fundamental first step in the GNN learning process. In this work, we construct the graph by treating spatial grid points in the domain as nodes. For instance, in a two-dimensional grid, each node corresponds to a spatial location $x_{(i,j)} \in \mathbb{R}^2$ where $i \in \{1,2,\dots,N_x\}$ and $j \in \{1,2,\dots,N_y\}$, such that the total number of nodes is $N = N_x \times N_y$. To define edge connectivity, we employ a finite-difference-like stencil and connect each node to its $k$-nearest neighbors, with $k=8$ in our case. Therefore, a target node's neighborhood includes the left, right, top, bottom, and four diagonal neighbors. The edge set is thus defined as $\mathcal{E} = \{ (u,v) \mid v \in \mathcal{N}_k(u) \}$, where $\mathcal{N}_k(u)$ refers to the $k$-nearest neighboring nodes of target node $u$.

To reiterate, while solving physical systems using GNNs, it is imperative to assign a node feature vector that captures both the physical fields and the geometric embedding of the node. Specifically, in this work, the node features include: (i) the physical field values of interest at the node (\textit{e.g.}, velocity, pressure, or material parameters), (ii) the absolute positional coordinates $(x_u, y_u)$, and (iii) Fourier positional encodings of the form $\big[ \sin(\omega x_u), \; \cos(\omega x_u), \; \sin(\omega y_u), \; \cos(\omega y_u) \big]$. The idea behind including additional static node features, \textit{\textit{i.e.}}, features that are not updated at every message-passing step (\textit{e.g.}, Fourier embeddings in this case), is to provide a richer context and enhance the representation capacity by embedding the spatial coordinates into a better feature space, thereby improving the learning process. The edge feature vectors between any pair of connected nodes $(u,v) \in \mathcal{E}$ are designed to encode both (relative) geometric information and physical interactions. For our work, we consider an edge feature vector that includes the following quantities: \vspace{-4pt}
\begin{enumerate}
    \item Relative displacement vector between the target node $u$ and sender node $v$: $\Delta \mathbf{x}_{(u,v)} = \mathbf{x}_v - \mathbf{x}_u$ \vspace{-4pt}
    \item Euclidean distance between the nodes $u$ and $v$: $d_{(u,v)} = \|\Delta \mathbf{x}_{(u,v)}\|_2$ \vspace{-4pt}
    \item Difference in the physical fields between the two nodes: $\Delta f_{(u,v)} = f_v - f_u$ \vspace{-4pt}
    \item Norm of the difference in physical fields between the two nodes: $\|\Delta f_{(u,v)}\|_2$ \vspace{-4pt}
\end{enumerate}

\rev{The graph construction and feature definitions described above pertain to the four PDE systems defined on uniform structured grids, where the $k$-nearest-neighbor stencil recovers the standard finite-difference neighborhood. For the 2D flow past cylinder case, which is defined on unstructured, variable-geometry triangular meshes, the graph instead inherits the native mesh topology (mesh nodes and edges map directly to graph nodes and edges), and the node and edge feature vectors are adapted to encode the irregular discretization and the boundary-node types. The complete specification for this case is provided in Section~\ref{sec:cylinder_flow}.}

\section{Results}
\label{sec:results}
The key objective of this study is to evaluate the generalization capabilities of GNS-based surrogate models in comparison with neural operator (NO)-based surrogates. We focus on representative examples from two broad classes of NOs: (1) universal approximation theorem-based methods (\textit{e.g.}, deep operator networks (DeepONet)) and (2) kernel integral operator-based methods (\textit{e.g.}, Fourier Neural Operators (FNO), \rev{Geometry-Informed Neural Operator (GINO)}). Specifically, we examine four DeepONet-based and two FNO-based frameworks: (1) DeepONet Full Rollout (DON-FR), (2) DeepONet Autoregressive (DON-AR), (3) Time-Integrated DeepONet (TI-DON), (4) Time-Integrated DeepONet with Learned Integration (TI(L)-DON), (5) FNO Full Rollout (FNO-FR), and (6) FNO Autoregressive (FNO-AR). \rev{These six frameworks are evaluated on the four PDE systems defined on uniform grids. For the flow past cylinder case, which is posed on unstructured meshes and therefore incompatible with the fixed-grid FNO and DeepONet baselines, we instead benchmark GNS against the Geometry-Informed Neural Operator with time integration (GINO-TI), a kernel-integral operator specifically designed to handle unstructured input geometries.} For brevity, we do not reproduce the architectural details of all the neural operators and their variants here; a detailed discussion and comparative study of these NO-based models can be found in~\cite{nayak2026ti}. \rev{The specific architectural configurations used in this work for all frameworks, including the number of layers, hidden dimensions, activation functions, and other hyperparameters, are reported in Appendix~\ref{sec:architecture_details}. Additionally, we provide a detailed explanation of the functioning of the TI-based neural operators, i.e., TI-DON and TI(L)-DON, in Appendix~\ref{sec:TI-based-NOs_details}.}

To ensure a fair and comprehensive evaluation, we benchmark both GNS and NO frameworks on \rev{five} canonical PDE systems that cover a diverse range of complex, nonlinear, high-dimensional, and spatiotemporal dynamics: (1) the 2D Burgers' equation with a scalar field solution, (2) the 2D coupled Burgers' equation with a vector field solution, (3) the 2D Allen–Cahn equation, (4) the 2D nonlinear shallow water equations (SWE), \rev{and (5) the 2D flow past cylinder}. \rev{The first four benchmarks are posed on uniform grids, since most NO architectures (e.g., FNO) require a structured grid domain, enabling a controlled comparison across all frameworks on a common discretization. The fifth benchmark (flow past cylinder) is instead posed on unstructured, variable-geometry meshes to directly assess the geometric flexibility of graph-based surrogates, where GNS is compared against the geometry-aware GINO-TI baseline.} For the comparison of inference errors, we consider a relative $L_2$ error metric, defined as
\begin{equation}
\varepsilon_{L_2} = \dfrac{||u_{\text{pred}} - u_{\text{truth}}||_2}{||u_{\text{truth}}||_2},
\end{equation}
where $u_{\text{pred}}$ and $u_{\text{truth}}$ denote the predicted and ground truth solutions, respectively.

\begin{figure}[htbp!]
    \centering
    \includegraphics[width=\linewidth]{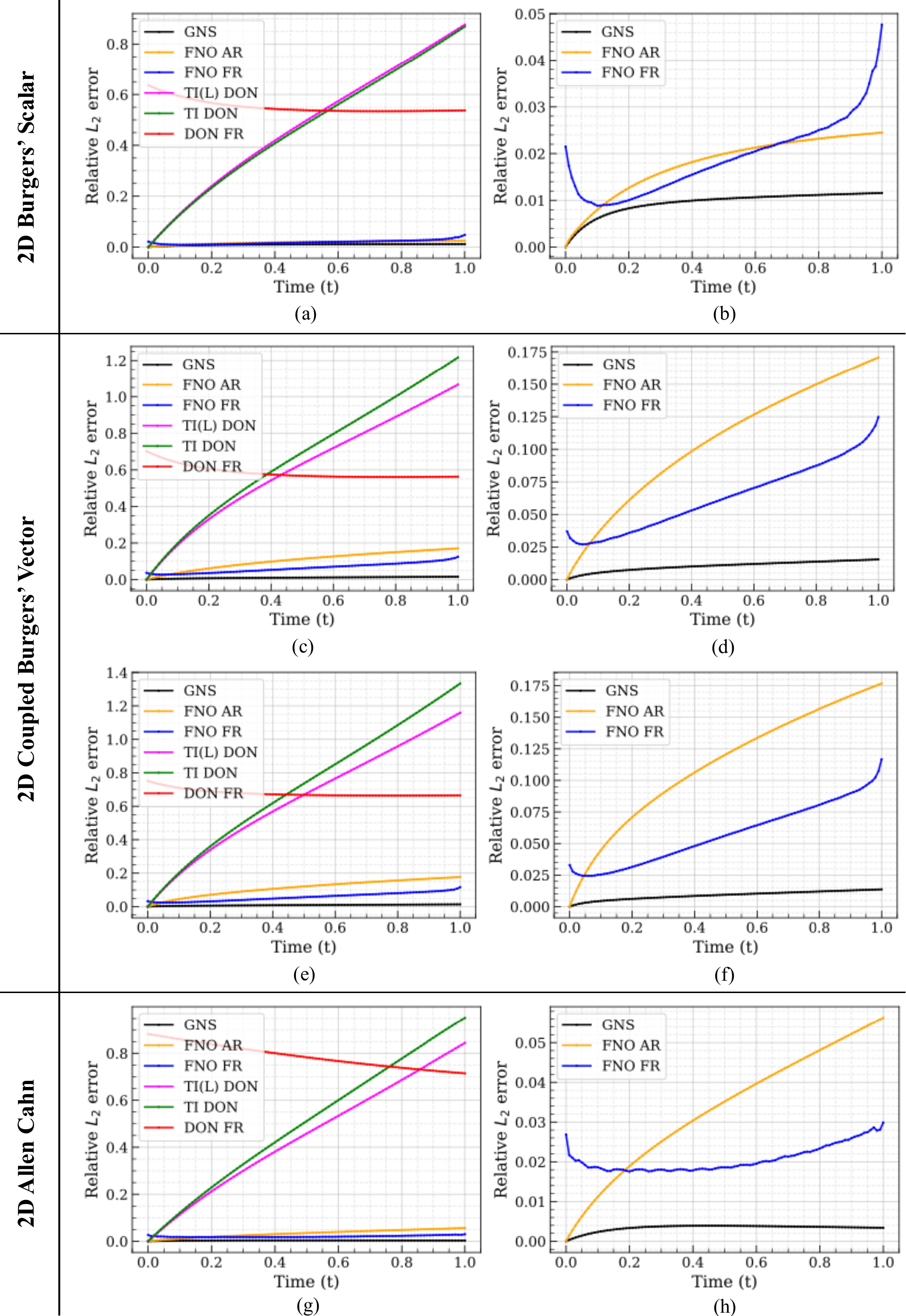}
    \caption{Temporal evolution of the relative $L_2$ error across different frameworks for the 2D Burgers' scalar, 2D Burgers' vector, and 2D Allen-Cahn equations, trained on 30 trajectories. Left column: comparison of all neural operator (NO) baselines with the GNS surrogate model. Right column: comparison of the GNS surrogate model with the FNO variants (FNO FR and FNO AR). For the 2D Burgers' vector case, (c) and (d) correspond to the $u$-velocity field, while (e) and (f) correspond to the $v$-velocity field. For brevity, DON AR is excluded as it exhibits unstable behavior and accumulates errors orders of magnitude higher than the other NO baselines.}
    \label{fig:err_acc}
\end{figure}

Figure~\ref{fig:err_acc} depicts the temporal error accumulation observed across all frameworks and the first three PDE examples considered in this study. Across these PDE systems, GNS (black) consistently exhibits the lowest error growth with a noticeable plateauing of errors, indicating that the GNS surrogate demonstrates accurate, reliable, and stable performance over extended temporal horizons. This is followed by FNO-FR (blue) and FNO-AR (orange), with FNO-FR performing better than FNO-AR for longer timesteps except in the 2D Burgers' scalar case, where FNO-AR surpasses FNO-FR beyond $t=0.7$. Note that FNO-FR incurs appreciable errors for the initial few timesteps across all three cases, beyond which it outperforms FNO-AR. Finally, DeepONet-based frameworks incur substantially higher errors than their counterparts, \textit{i.e.}, FNO and GNS. The primary reason is that DeepONet generally requires significantly more training data than the other methods to construct an accurate set of basis functions and coefficients. Among the DeepONet frameworks, the time-integrator variants, TI(L)-DON (magenta) and TI-DON (green), perform better than DON-FR (red) for the initial timesteps until $t=0.45$ or $t=0.5$. Beyond this point, DON-FR surpasses these frameworks. The aforementioned observations translate directly to the comparison plots in subsequent sections where we discuss individual cases separately and examine this phenomenon through both qualitative and quantitative assessment of generalization performance via relative $L_2$ test errors during inference. Details regarding the computational costs incurred during training and inference are provided in Table~\ref{tab:training-inference-times} in the Supplementary Information (SI). Additional information on the network architectures, as well as further results on predicted solution contours, is provided in SI~\ref{sec:architecture_details} and SI~\ref{sec:additional_results}, respectively. 
The code to reproduce the experiments is publicly available at \url{https://github.com/Centrum-IntelliPhysics/GNS_vs_NOs.git}.
\begin{table}[htbp!]
    \centering
    \renewcommand{\arraystretch}{1.2}
    \setlength{\tabcolsep}{3pt}
    \caption{Ground truth dataset generation details for \rev{four} PDE cases\rev{, excluding the 2D flow past cylinder case}. \rev{The dataset for the 2D flow past cylinder case is taken from \cite{pfaff2020learning}.} The data generation scripts for the 2D Burgers' and 2D nonlinear SWE examples were adapted from~\cite{Rosofsky_2023}. The generalization error of all frameworks for each PDE is evaluated using training dataset sizes of 30, 50, 70, and 100 samples. Training trajectories are selected using a combined PCA and KMeans clustering approach.}
    \begin{tabular}{|c|c|c|c|c|c|c|c|c|c|c|}
        \hline
        Case & GRF Kind & BCs & $l$ & $\sigma$ & $N_s$ & $N_{train}$ & $N_{test}$ & $N_x \times N_y$ & $N_t$ \\ \hline
        \multirow{4}{*}{\makecell{2D Burgers' \\ Scalar}} & \multirow{4}{*}{Mat\'ern} & \multirow{4}{*}{Periodic} & \multirow{4}{*}{0.125} & \multirow{4}{*}{0.15} & \multirow{4}{*}{1000} & 30 & \multirow{4}{*}{\rev{200}} & \multirow{4}{*}{$32 \times 32$} & \multirow{4}{*}{101} \\ \cline{7-7}
         &  &  &  &  &  & 50 &  &  &  \\ \cline{7-7}
         &  &  &  &  &  & 70 &  &  &  \\ \cline{7-7}
         &  &  &  &  &  & 100 &  &  &  \\ \hline
        \multirow{4}{*}{\makecell{2D Burgers' \\ Vector}} & \multirow{4}{*}{Mat\'ern} & \multirow{4}{*}{Periodic} & \multirow{4}{*}{0.1} & \multirow{4}{*}{0.2} & \multirow{4}{*}{500} & 30 & \multirow{4}{*}{\rev{100}} & \multirow{4}{*}{$64 \times 64$} & \multirow{4}{*}{101} \\ \cline{7-7}
         &  &  &  &  &  & 50 &  &  &  \\ \cline{7-7}
         &  &  &  &  &  & 70 &  &  &  \\ \cline{7-7}
         &  &  &  &  &  & 100 &  &  &  \\ \hline
        \multirow{4}{*}{2D Allen-Cahn} & \multirow{4}{*}{\makecell{Squared- \\ Exponential}}  & \multirow{4}{*}{Periodic} & \multirow{4}{*}{0.05} & \multirow{4}{*}{\makecell{Normalized \\ to [-1,1]}} & \multirow{4}{*}{1000} & 30 & \multirow{4}{*}{\rev{200}} & \multirow{4}{*}{$32 \times 32$} & \multirow{4}{*}{101} \\ \cline{7-7}
         &  &  &  &  &  & 50 &  &  &  \\ \cline{7-7}
         &  &  &  &  &  & 70 &  &  &  \\ \cline{7-7}
         &  &  &  &  &  & 100 &  &  &  \\ \hline
         \multirow{4}{*}{\makecell{2D nonlinear \\ SWE}} & \multirow{4}{*}{Mat\'ern} & \multirow{4}{*}{Periodic} & \multirow{4}{*}{0.1} & \multirow{4}{*}{0.2} & \multirow{4}{*}{500} & 30 & \multirow{4}{*}{\rev{100}} & \multirow{4}{*}{$64 \times 64$} & \multirow{4}{*}{101} \\ \cline{7-7}
         &  &  &  &  &  & 50 &  &  &  \\ \cline{7-7}
         &  &  &  &  &  & 70 &  &  &  \\ \cline{7-7}
         &  &  &  &  &  & 100 &  &  &  \\ \hline
    \end{tabular}
    \label{tab:dataset_generation_details}
\end{table}

\subsection{Two-Dimensional Burgers’ Equation with Scalar Output Field}
\label{subsec:2d-burgers-scalar}
In the first example, we consider the two-dimensional viscous Burgers’ equation with a scalar output field. This PDE, often regarded as a simplified form of the Navier–Stokes equations, captures essential features of fluid transport by balancing nonlinear advection with viscous diffusion. Despite its relative simplicity, it exhibits rich spatiotemporal dynamics and is widely used as a benchmark for surrogate modeling. The governing equation is:
\begin{equation}
    \frac{\partial u}{\partial t} + u \frac{\partial u}{\partial x} + u \frac{\partial u}{\partial y} = \nu \left( \frac{\partial^2 u}{\partial x^2} + \frac{\partial^2 u}{\partial y^2} \right), ~~~~ \forall ~ (x, y, t) \in [0, 1]^2 \times [0, 1],
    \label{eq:2d_burgers}
\end{equation}
where $u(x, y, t)$ is a scalar field and $\nu = 0.01$ is the kinematic viscosity. The IC, $u(x, y, 0) = u_0(x, y)$ is sampled from 2D Gaussian random field, and periodic Dirichlet and Neumann BCs are applied in both $x$ and $y$ directions. 

Mathematically, this is a nonlinear parabolic PDE in which convective terms steepen the gradients while diffusion smoothens them, leading to phenomena such as viscous shock formation and spatiotemporal mixing. For surrogate modeling, two challenges arise: (i) the nonlinearity amplifies approximation errors during long-horizon rollouts, and (ii) the relative influence of convection and diffusion, controlled by $\nu$, requires the model to generalize across regimes ranging from diffusion-dominated to advection-dominated dynamics.

To select training trajectories, we adopt a strategy combining principal component analysis (PCA) with KMeans clustering. Specifically, the full dataset is projected onto its first 20 principal components, after which KMeans clustering is performed with the number of clusters equal to the desired number of training trajectories. From each cluster, we then select the trajectory closest to the centroid in terms of Euclidean distance. This procedure ensures that the chosen training set captures the diversity of the full dataset in a more systematic manner than pure random sampling. While many sophisticated sampling strategies exist in the literature, we employ this relatively simple yet intuitive approach as a principled way of trajectory selection.

\rev{We note that this PCA combined with KMeans clustering graning data selection strategy is an offline dataset curation method: the principal components and cluster assignments are computed from the full pool of candidate output trajectories, which must therefore have been generated in advance. Its purpose is thus to construct representative training subsets of a prescribed size from an existing dataset, both as a controlled means of studying performance versus training budget and as a way to reduce the training cost of the surrogate itself, since the number of gradient updates per epoch and the memory footprint scale with the size of the training set. It does not reduce the cost of the initial data generation; settings in which generating even the candidate pool is prohibitive would call for adaptive online strategies such as active learning~\cite{cohn1996active}, which we leave as a direction for future work.}
\begin{figure}[htbp!]
    \centering
    \includegraphics[width=\linewidth]{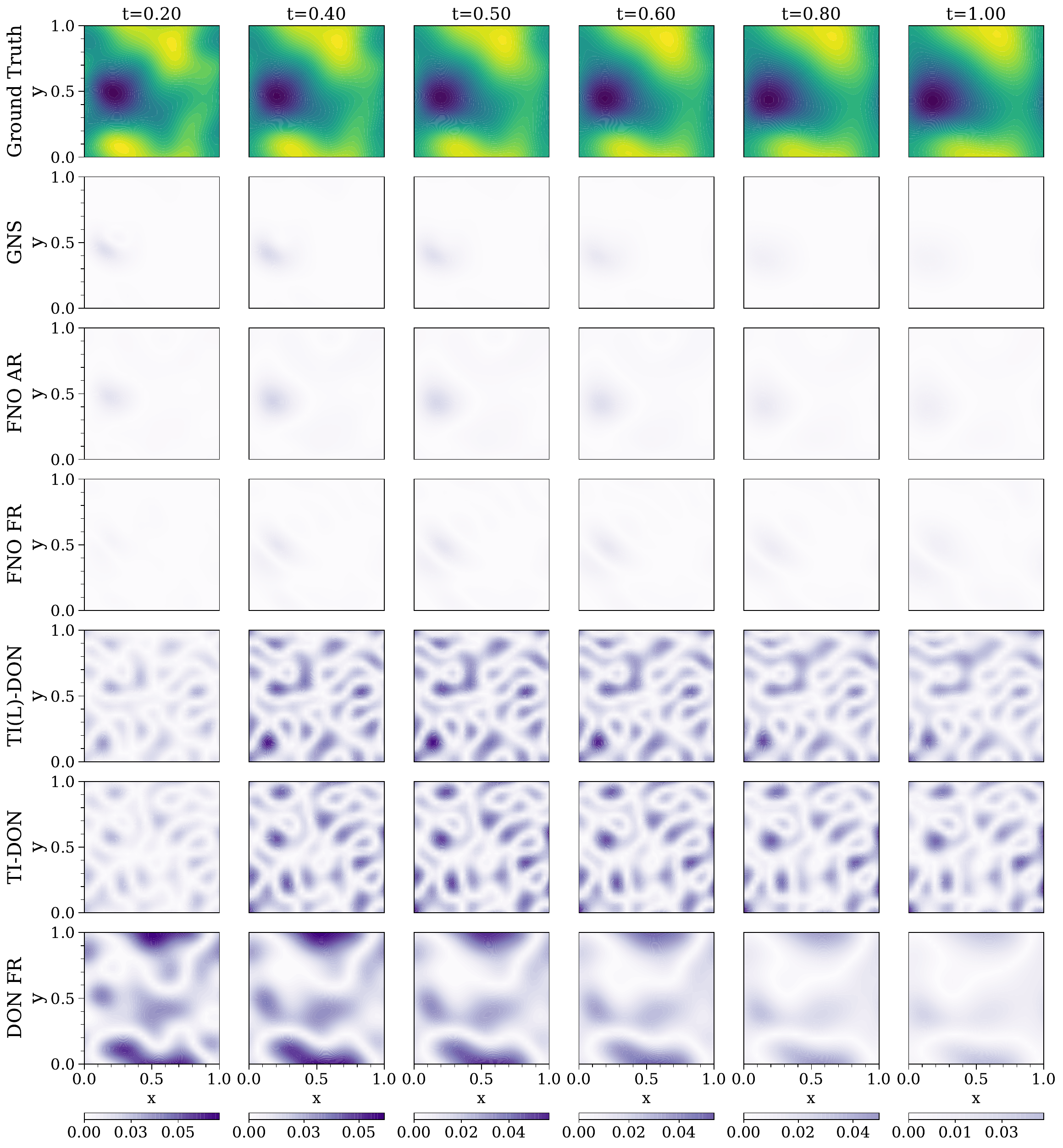}
    \includegraphics[width=\linewidth]{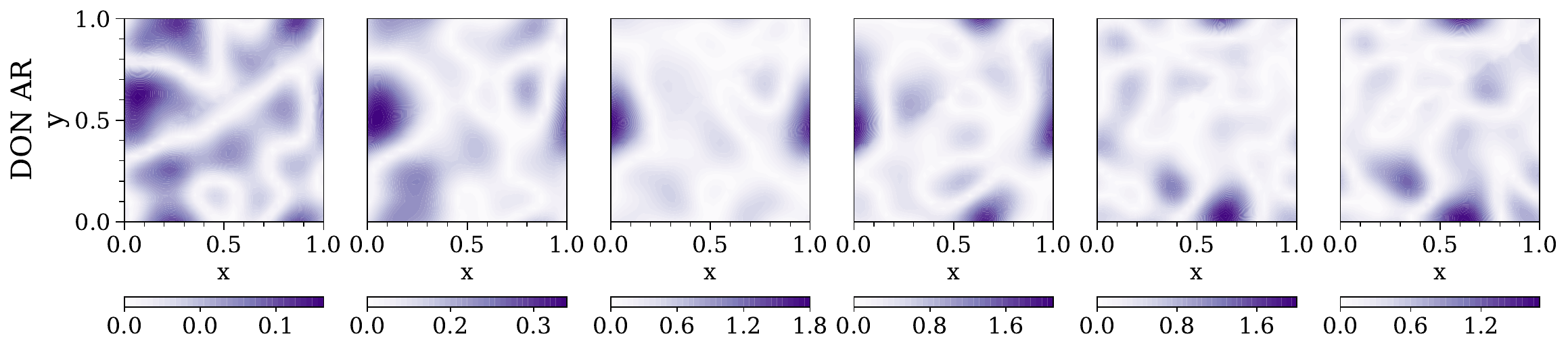}
    \caption{\rev{2D Burgers' equation with scalar output field: Performance of all frameworks trained on 30 trajectories and evaluated on a held-out set of 200 trajectories. Top row: ground truth solution contours. Subsequent rows: absolute error contours for each framework, shown for a representative sample.}}
    \label{fig:2D_Burgers_contours}
\end{figure}
\begin{figure}[htbp!]
    \centering
    \includegraphics[width=\linewidth]{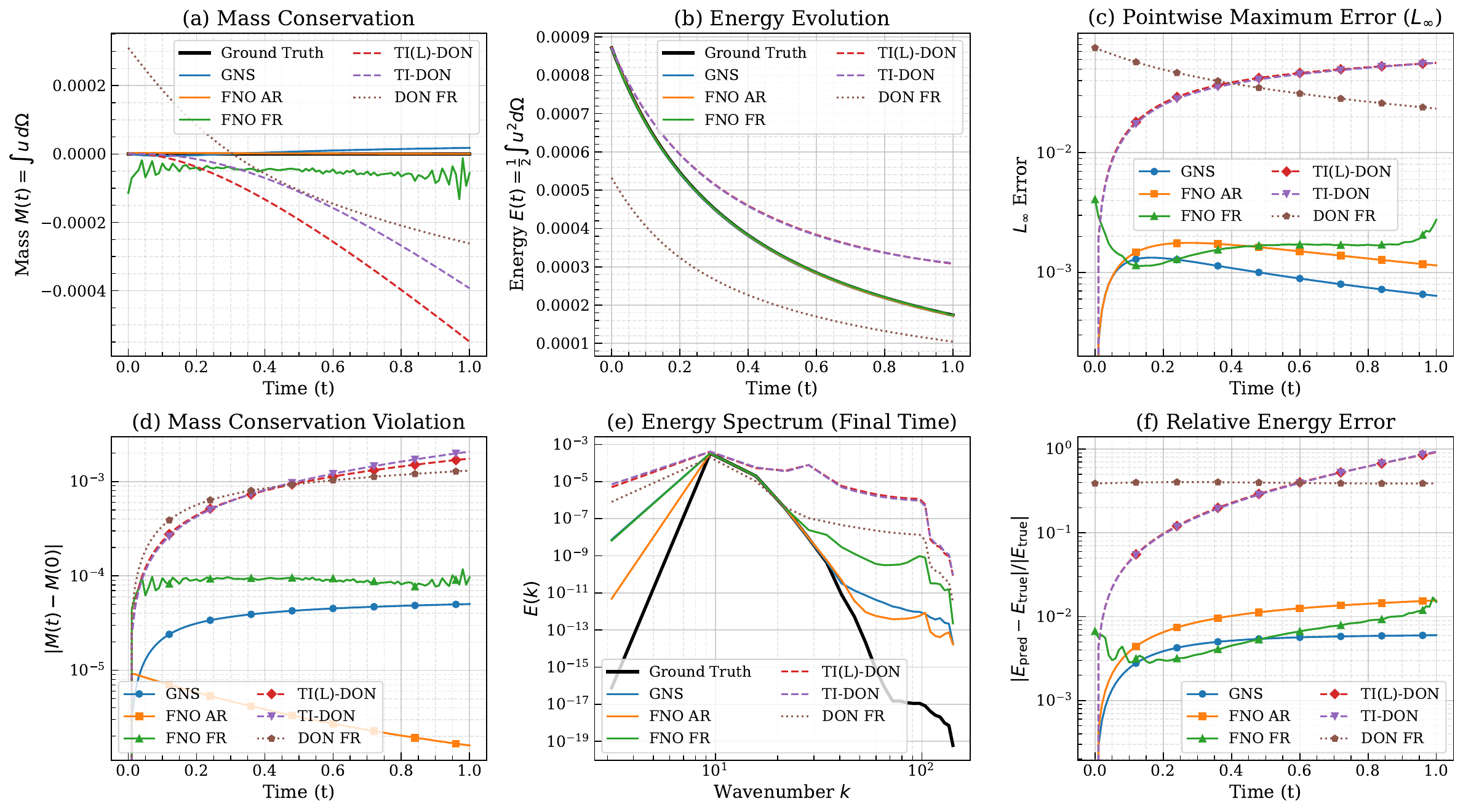}
    \caption{\rev{2D Burgers' equation with scalar output field: (a) Mass evolution over time, (b) energy evolution over time, (c) pointwise maximum $L_\infty$ error growth over time, (d) mass conservation violation error, (e) spectral energy density $E(k)$ as a function of wavenumber $k$ at the final timestep, and (f) relative energy error over time, compared across different frameworks.}}
    \label{fig:2D_Burgers_scalar_conservation_metrics}
\end{figure}

We next provide a brief description of the learning strategies employed by the various frameworks. GNS learns a one-timestep mapping $u_t \rightarrow u_{t+1}$ by estimating the time derivative conditioned on the current state, $\frac{\partial u}{\partial t}\big|_{u(t, x, y)}$, and then applies an explicit Euler update to predict the next state $u(t+1, x, y)$ from the input state $u(t, x, y)$. The full rollout methods, such as FNO-FR and DON-FR, learn the solution operator that maps the initial condition to the entire spatiotemporal output field, \textit{\textit{i.e.}}, $\mathcal{G} \sim \mathcal{G}_{\theta}: u_0(x, y) \rightarrow u(t, x, y)$. In contrast, the autoregressive frameworks, such as FNO-AR and DON-AR, directly learn the one-timestep mapping operator that advances the solution state forward in time, \textit{\textit{i.e.}}, $\mathcal{G} \sim \mathcal{G}_{\theta}: u(t, x, y) \rightarrow u(t+1, x, y)$. Finally, the time integrator (TI)-based frameworks (see~\cite{nayak2026ti} for more details) learn the instantaneous time-derivative operator, \textit{\textit{i.e.}}, $\mathcal{G} \sim \mathcal{G}_{\theta}: u(t, x, y) \rightarrow \frac{\partial u}{\partial t}\big|_{u(t, x, y)}$. It is noteworthy that the learning strategies of GNS and the TI-based frameworks are analogous, with the latter having more relevance in an operator learning setting.

Figure~\ref{fig:2D_Burgers_contours} presents a comparison of the absolute error in predictions for a representative test trajectory across all frameworks. Among the DeepONet-based models, the autoregressive variant (DON-AR) incurs errors nearly an order of magnitude higher than the others. This can be attributed to its greater sensitivity to error accumulation during inference and the limited generalization ability of DeepONet in data-scarce regimes. The full rollout DeepONet (DON-FR) and the time-integrator (TI)-based variants yield comparable errors, with DON-FR performing qualitatively worse than the TI-based frameworks during the early timesteps ($t = \{0.2, 0.4, 0.5\}$). Between the TI-based models, the learnable version, TI(L)-DeepONet, consistently outperforms its vanilla counterpart, \textit{\textit{i.e.},} TI-DeepONet. In contrast, the Fourier Neural Operator (FNO)-based frameworks demonstrate markedly better performance. Both the full rollout (FNO-FR, using FNO-3D) and autoregressive (FNO-AR, using FNO-2D) variants achieve lower prediction errors than all DeepONet-based models, with errors marginally below those of the GNS. Finally, GNS achieves the best overall performance, significantly reducing generalization error and producing predictions that closely match the ground truth contours, even when trained on only 30 trajectories. This superior accuracy can be attributed to the strong inductive bias encoded in the GNS architecture: by explicitly modeling local spatial interactions between nodes in the computational graph, GNS effectively mirrors the underlying physical dynamics, thereby enabling robust learning in data-scarce regimes.
\begin{table}[htbp!]
    \renewcommand{\arraystretch}{1.25}
    \setlength{\tabcolsep}{4pt}
    \centering
    \caption{\rev{2D Burgers' equation with scalar output field: Inference errors (mean $\pm$ std over 5 seeds) for $u$ field across different methods and training trajectory counts. At inference, held-out set of 200 test trajectories are used, with a timestep of $\Delta t = 0.01$ for the time integrator-based frameworks (GNS, TI-DON, TI(L)-DON).}}
    \color{black}{
    \begin{tabular}{|>{\centering\arraybackslash}m{1.75cm}|
                    *{4}{>{\centering\arraybackslash}m{2.5cm}|}}
        \hline
        \multirow{2}{*}{Method} 
        & \multicolumn{4}{c|}{Training Trajectories} \\
        \cline{2-5}
         & 30 & 50 & 70 & 100 \\
        \hline
        GNS  
           & \bf{0.0054$\pm$0.0004} 
           & \bf{0.0043$\pm$0.0001} 
           & \bf{0.0038$\pm$0.0003} 
           & \bf{0.0036$\pm$0.0005}\\
        \hline
        DON FR  
           & 0.513$\pm$0.024 
           & 0.284$\pm$0.010 
           & 0.197$\pm$0.008 
           & 0.139$\pm$0.004\\
        \hline
        DON AR  
           & 5.01$\pm$2.64 
           & 4.63$\pm$1.72 
           & 7.59$\pm$4.43 
           & 2.33$\pm$1.60\\
        \hline
        TI-DON  
           & 0.449$\pm$0.018 
           & 0.282$\pm$0.013 
           & 0.225$\pm$0.004 
           & 0.181$\pm$0.003\\
        \hline
        TI(L)-DON  
           & 0.457$\pm$0.027 
           & 0.283$\pm$0.008 
           & 0.223$\pm$0.009 
           & 0.177$\pm$0.002\\
        \hline
        FNO AR  
           & 0.0143$\pm$0.0024 
           & 0.0084$\pm$0.0014 
           & 0.0068$\pm$0.0011 
           & 0.0054$\pm$0.0009\\
        \hline
        FNO FR  
           & 0.0174$\pm$0.0011 
           & 0.0131$\pm$0.0012 
           & 0.0118$\pm$0.0010 
           & 0.0060$\pm$0.0005\\
        \hline
    \end{tabular}
    }
    \label{tab:2D_Burgers_results_summary}
\end{table}
\begin{figure}[htbp!]
    \centering
    \begin{subfigure}[b]{0.495\linewidth}
        \centering
        \includegraphics[width=\linewidth]{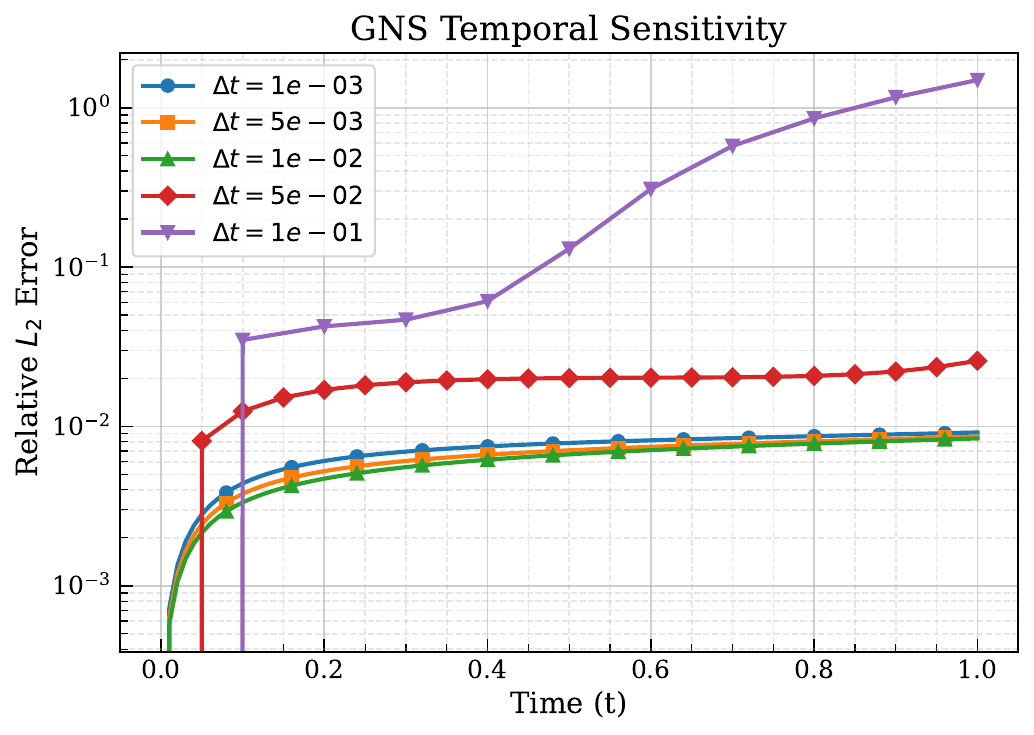}
        \caption{}
        \label{fig:dt_sensitivity}
    \end{subfigure}
    \hfill
    \begin{subfigure}[b]{0.495\linewidth}
        \centering
        \includegraphics[width=\linewidth]{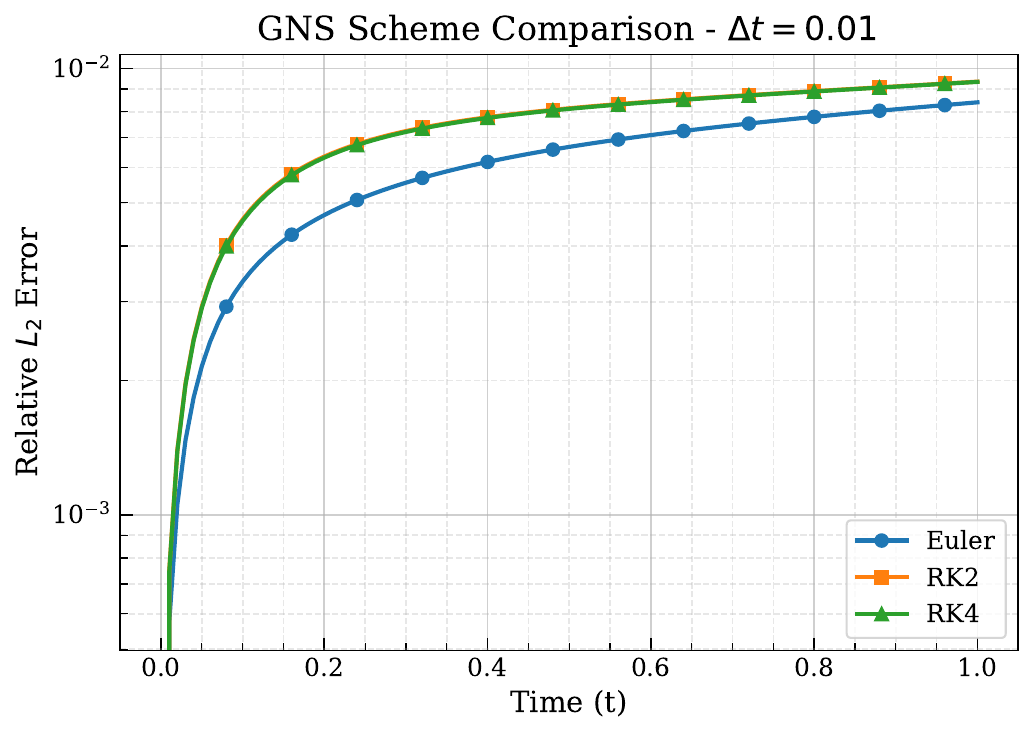}
        \caption{}
        \label{fig:scheme_comparison}
    \end{subfigure}
    \caption{\rev{2D Burgers' equation with scalar output field: (a) Temporal evolution of the relative $L_2$ error during inference of the GNS model for different timestep sizes $\Delta t \in \{0.001, 0.005, 0.01, 0.05, 0.1\}$, with a training timestep of $\Delta t = 0.01$. (b) Temporal evolution of the relative $L_2$ error across three time integration schemes (explicit Euler, RK2, and RK4) for a fixed inference timestep of $\Delta t = 0.01$.}}
    \label{fig:temporal_sensitivity_analysis}
\end{figure}

Next, we quantitatively assess the generalization performance of all frameworks by computing the overall relative $L_2$ test error on \rev{200} unseen \rev{test} trajectories, with models trained on subsets of 30, 50, 70, and 100 trajectories. The summarized results are reported in Table~\ref{tab:2D_Burgers_results_summary}. These observations are consistent with the qualitative comparisons presented in Figure~\ref{fig:2D_Burgers_contours}. As expected, increasing the number of training trajectories reduces inference error across all frameworks, consistent with the well-established fact that larger training datasets help narrow the generalization gap in data-driven models. Among the DeepONet-based frameworks, DON-AR consistently exhibits the largest relative $L_2$ errors, nearly an order of magnitude higher than all other methods, confirming its larger sensitivity to error accumulation during rollout. This is followed by DON-FR, TI-DON, and TI(L)-DON. For smaller training sets (30 and 50 trajectories), TI(L)-DON outperforms TI-DON and DON-FR, likely due to its ability to learn adaptive integration weights. At 70 training trajectories, these three methods perform comparably, while with 100 training trajectories, DON-FR slightly surpasses the TI-based variants. This improvement can be attributed to DON-FR’s more accurate basis functions and coefficients representative of the output function space, which better captures the PDE dynamics at higher data availability. In contrast, the FNO-based frameworks (FNO-FR and FNO-AR) perform significantly better than their DeepONet counterparts. This can be attributed to two main reasons. First, DeepONet operates in physical space and being based on the universal approximation theorem for operators, is inherently more data-hungry. Second, learning the solution operator in spectral space provides a distinct advantage: for dissipative systems such as Burgers’, FNO captures global features (dominant modes) in the spatiotemporal solution, resulting in stronger generalization capacity. While FNO is itself data-hungry, the dynamics of the Burgers’ PDE favors spectral learning, thereby enhancing its performance.  

Finally, the GNS framework outperforms all other methods by a substantial margin. Remarkably, GNS achieves a relative $L_2$ error as low as $0.88\%$ when trained on only 30 trajectories. This superior performance aligns with our earlier analysis: by explicitly modeling spatial interactions through its graph-based architecture, GNS embeds a strong physics-aware inductive bias, allowing it to naturally encode the underlying dynamical laws of the system and generalize robustly even in the data-scarce regime. \rev{While relative $L_2$ error is a standard metric to assess model performance in ML modeling, in physical problems, especially in CFD, a surrogate model must also respect the governing physics, especially the conservation laws that underpin the dynamical system. To this end, we visualize the conservation violation errors for mass and energy, the pointwise $L_{\infty}$ error over time, and the spectral energy density $E(k)$ as a function of wavenumber $k$ in Fig.~\ref{fig:2D_Burgers_scalar_conservation_metrics}. Under periodic boundary conditions, the total mass $M(t) = \int u \, d\Omega$ is a strict invariant of the continuous PDE, since both the viscous term and the nonlinear advection term integrate to zero over the domain; any drift in panels~(a) and~(d) therefore reflects a spurious source or sink introduced by the surrogate itself. The DeepONet variants exhibit a smooth monotonic drift of $\mathcal{O}(10^{-3})$, while FNO-FR shows smaller non-monotonic fluctuations of $\mathcal{O}(10^{-4})$. FNO-AR achieves the tightest mass conservation of all frameworks, driving the violation down to $\mathcal{O}(10^{-6})$ by the final time, with GNS second at $\mathcal{O}(10^{-5})$.}

\rev{The kinetic energy $E(t) = \tfrac{1}{2}\int u^2 \, d\Omega$ decays at the rate $\nu \int |\nabla u|^2 \, d\Omega$, since the nonlinear advection term integrates to zero over the periodic domain and only viscosity dissipates energy. Panel~(b) shows the TI-DON variants under-dissipate, while DON-FR starts from an already corrupted initial energy and over-dissipates thereafter, yielding relative energy errors of $\mathcal{O}(10^{-1})$--$\mathcal{O}(10^{0})$ (panel~(f)). GNS attains the lowest relative energy error of all frameworks, settling at $\mathcal{O}(10^{-3})$, below both FNO-AR and FNO-FR, which sit at $\mathcal{O}(10^{-2})$ by the final time. The final-time spectrum in panel~(e) refines this comparison. The spectrum peaks near $k \approx 10$ and decays on either side; the ground truth falls off sharply for $k \gtrsim 10$, characteristic of viscous dissipation acting on the small scales. In the low-wavenumber range below the peak, GNS follows the ground-truth spectrum closely, whereas the FNO variants fall well below it, underrepresenting the large-scale energy content. In the high-wavenumber range above the peak, all surrogates overestimate the spectral energy relative to the sharply decaying ground truth, but by markedly different margins: GNS and FNO-AR remain closest to the ground truth, FNO-FR sits higher, and the DeepONet variants remain nearly flat far into the high-$k$ range, indicating substantial spurious energy content at small scales rather than physical dynamics. This spectral picture is consistent with the pointwise $L_\infty$ error (panel~(c)), where GNS maintains $\mathcal{O}(10^{-3})$ throughout the rollout while the FNO variants settle an order of magnitude higher at $\mathcal{O}(10^{-2})$. The overall picture is nuanced: FNO-AR conserves mass most tightly (panel~(d)), whereas GNS is superior on energy conservation, pointwise $L_\infty$ error, and the fidelity of the spectral energy distribution across scales.}

\rev{Finally, we assess the temporal robustness of the GNS rollout in Fig.~\ref{fig:temporal_sensitivity_analysis}. Panel~(a) shows that the inference error is essentially insensitive to $\Delta t$ for $\Delta t \le \Delta t_{\mathrm{train}} = 10^{-2}$, since the one-step approximation error of the learned model dominates the temporal
discretization error; the rollout remains stable at $\Delta t = 5 \times 10^{-2}$ but diverges at $\Delta t = 10^{-1}$, indicating the empirical stability boundary of the explicit Euler update. Panel~(b) compares Euler, RK2, and RK4 at the training $\Delta t$: the higher-order schemes yield no improvement and in fact perform marginally worse, because the training objective supervises only the one-step Euler update, so evaluating the learned RHS at intermediate RK stages probes states outside the training distribution and inherits the model's fixed one-step bias at each sub-stage. In effect, the current formulation makes GNS a learned Euler integrator rather than a learned continuous-time RHS operator; lifting this restriction would require training against a multi-stage or continuous-in-time loss and is left to future work.}

\subsection{Two-Dimensional Coupled Burgers’ Equation with Vector Output Field}
\label{subsec:2d-burgers-vector}
The second example we consider is the two-dimensional coupled Burgers’ equation with vector-valued outputs, representing the velocity components $u(x, y, t)$ and $v(x, y, t)$. This system can be regarded as a simplification of the incompressible Navier–Stokes equations that does not account for pressure effects, extending the scalar Burgers’ equation by introducing two interacting velocity fields. In doing so, it captures richer transport phenomena by balancing nonlinear cross-advection with viscous diffusion. The system is defined on the spatial domain $(x, y) \in [0, 1]^2$ over the time interval $t \in [0, 1]$, and its governing equations are:
\begin{equation}
    \dfrac{\partial u}{\partial t} + u\dfrac{\partial u}{\partial x} + v\dfrac{\partial u}{\partial y} 
    = \mu \left( \dfrac{\partial^2 u}{\partial x^2} + \dfrac{\partial^2 u}{\partial y^2} \right),
\end{equation}
\begin{equation}
    \dfrac{\partial v}{\partial t} + u\dfrac{\partial v}{\partial x} + v\dfrac{\partial v}{\partial y} 
    = \mu \left( \dfrac{\partial^2 v}{\partial x^2} + \dfrac{\partial^2 v}{\partial y^2} \right),
\end{equation}
where $\mu = 0.01$ denotes the viscosity coefficient. The initial conditions $u(x, y, 0) = u_0(x, y)$ and $v(x, y, 0) = v_0(x, y)$ are sampled from two-dimensional Gaussian random fields, while periodic Dirichlet and Neumann boundary conditions are imposed in both $x$ and $y$ directions.

Mathematically, this system is a nonlinear parabolic PDE with coupled advection–diffusion dynamics. The convective terms jointly transport both velocity components, while the viscous terms act to dissipate energy and smooth the solution. Physically, this makes the system a closer analog to the Navier–Stokes equations, albeit without the pressure term and incompressibility constraint. The coupling between $u$ and $v$ gives rise to complex spatiotemporal structures, including interacting shear layers and vortex-like features, which are absent in the scalar case. From a surrogate modeling perspective, additional challenges arise due to (i) the cross-advection terms, which tightly couple the dynamics of $u$ and $v$, and (ii) the increased dimensionality of the output field, requiring the model to simultaneously learn and generalize across both velocity components. Compared to the scalar case, this coupled vector formulation represents a more challenging benchmark for evaluating the generalization capabilities of all the surrogate models considered in this study. The ground truth dataset is generated following a similar approach as the previous case, where the initial conditions $u_0(x, y)$ and $v_0(x, y)$ are sampled from periodic Mat\'ern-type Gaussian random fields (GRFs) (see Table~\ref{tab:dataset_generation_details} for specific details). For training trajectory selection, we employ the PCA-KMeans procedure described in Section~\ref{subsec:2d-burgers-scalar}, retaining the first 50 principal components during the PCA dimensionality reduction step.
\begin{figure}[htbp!]
    \centering
    \includegraphics[width=\linewidth]{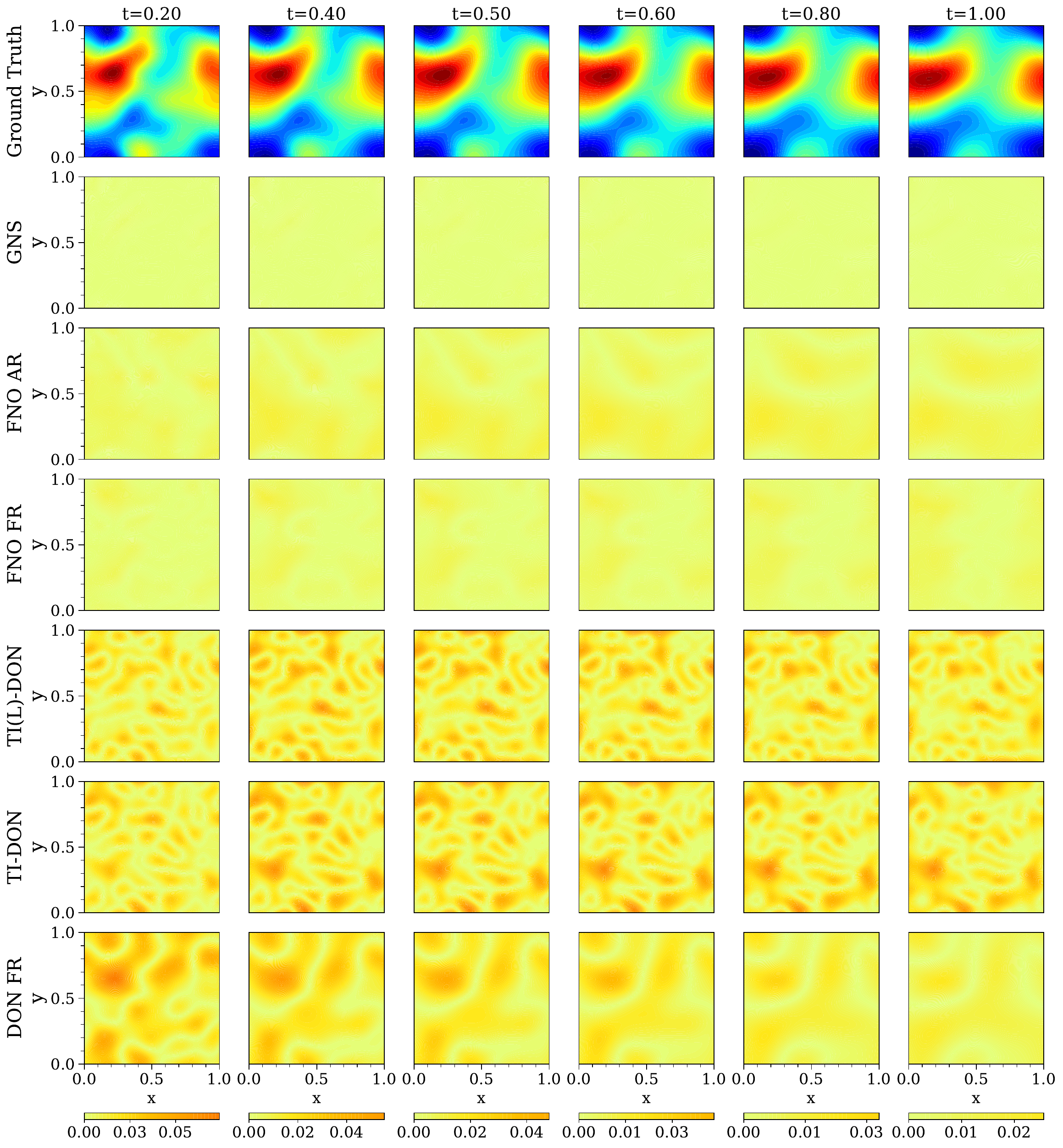}
    \includegraphics[width=\linewidth]{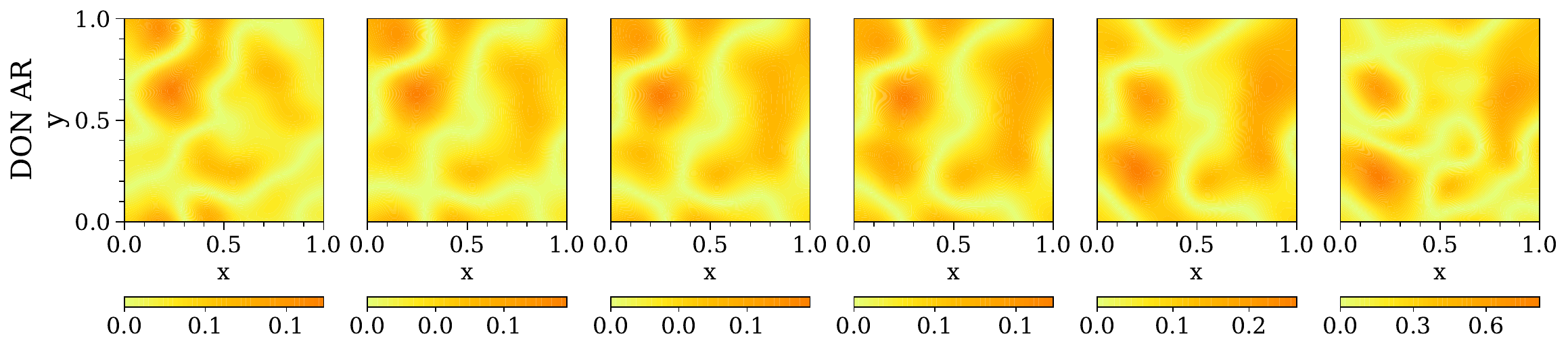}
    \caption{\rev{2D coupled Burgers' equation with vector output field: Performance of all frameworks trained on 30 trajectories and evaluated on a held-out set of 100 trajectories for predicting the $u$-velocity field. Top row: ground truth solution contours. Subsequent rows: absolute error contours for each framework, shown for a representative sample.}}
    \label{fig:2D_coupled_Burgers_contours_u}
\end{figure}
\begin{figure}[htbp!]
    \centering
    \includegraphics[width=\linewidth]{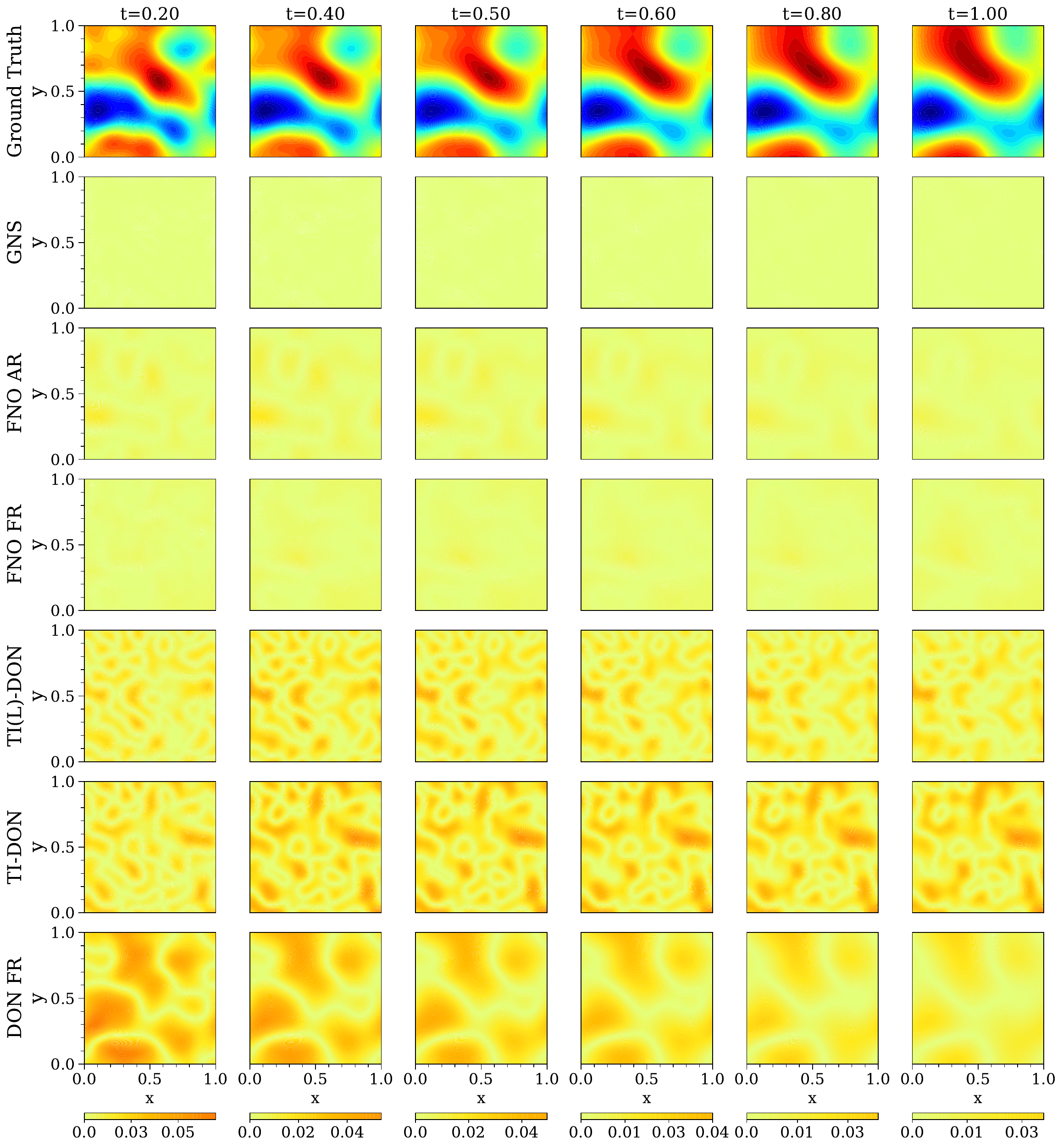}
    \includegraphics[width=\linewidth]{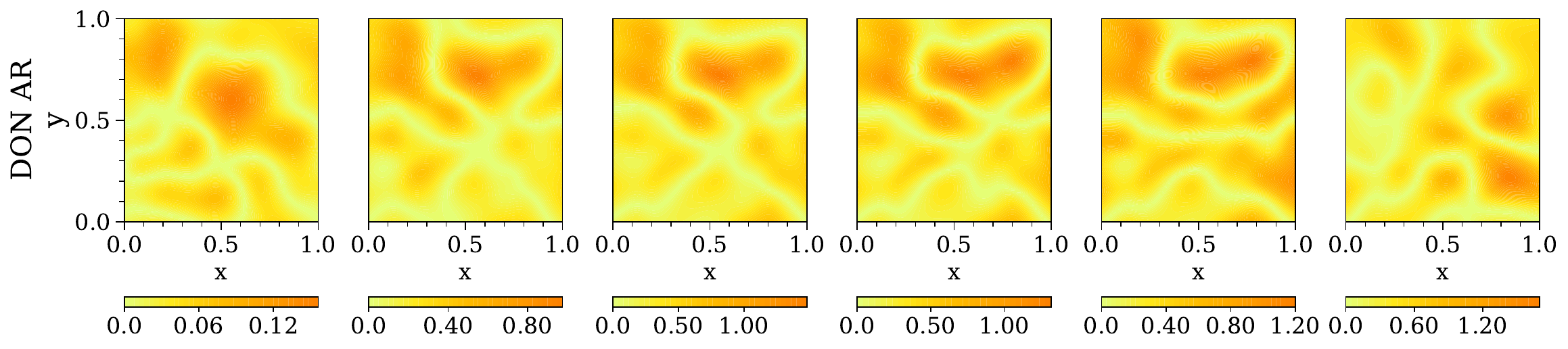}
    \caption{\rev{2D coupled Burgers' equation with vector output field: Performance of all frameworks trained on 30 trajectories and evaluated on a held-out set of 100 trajectories for predicting the $v$-velocity field. Top row: ground truth solution contours. Subsequent rows: absolute error contours for each framework, shown for a representative sample.}}
    \label{fig:2D_coupled_Burgers_contours_v}
\end{figure}
\begin{figure}[htbp!]
    \centering
    \includegraphics[width=\linewidth]{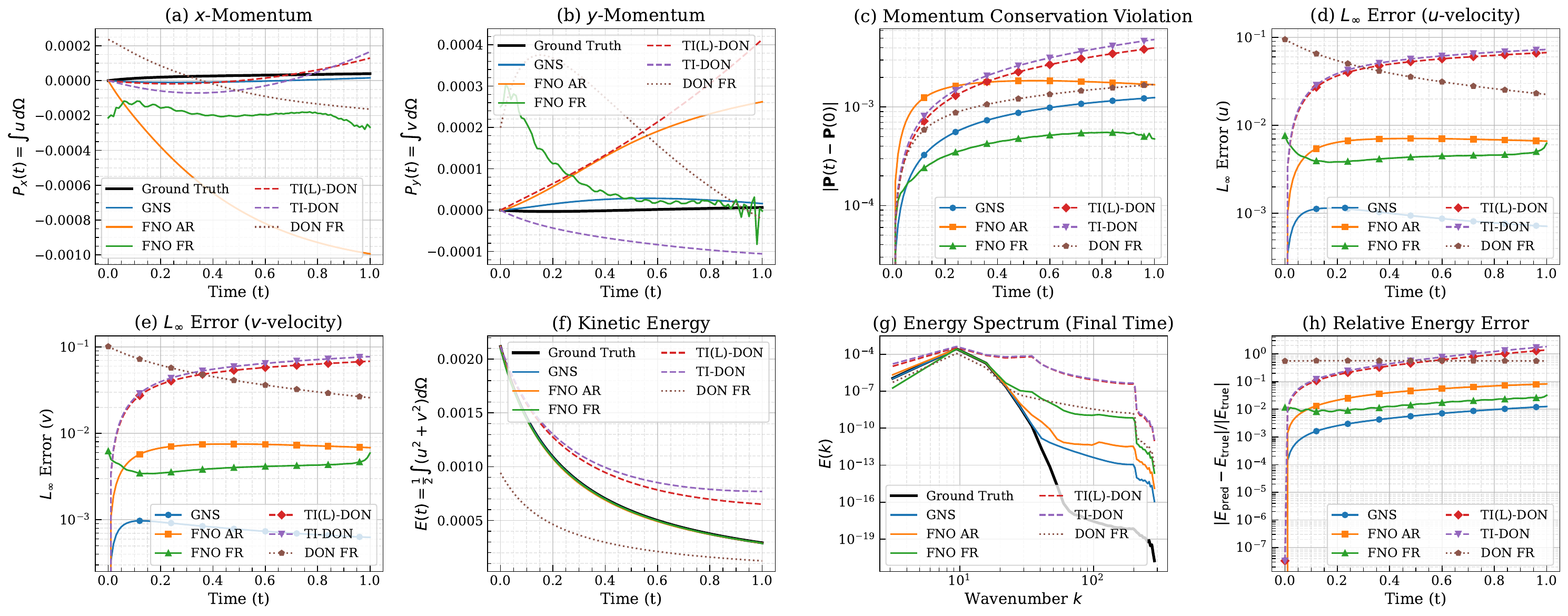}
    \caption{\rev{2D coupled Burgers' equation with vector output field: (a) Evolution of $x$-momentum $P_x$, (b) evolution of $y$-momentum $P_y$, (c) momentum conservation violation error, (d) pointwise maximum $L_\infty$ error for $u$-velocity, (e) pointwise maximum $L_\infty$ error for $v$-velocity, (f) kinetic energy evolution over time, (g) spectral energy density $E(k)$ as a function of wavenumber $k$ at the final timestep, and (h) energy conservation violation error.}}
    \label{fig:2D_Burgers_vector_conservation_metrics}
\end{figure}

In a similar spirit, we first present a qualitative comparison of the prediction test errors incurred during inference on \rev{100 unseen test} trajectories, trained on 30 trajectories. The results are shown in Figs.~\ref{fig:2D_coupled_Burgers_contours_u} and \ref{fig:2D_coupled_Burgers_contours_v} for the $u$ and $v$ velocity fields, respectively. While the contours are displayed separately for the two fields, the overall trends remain consistent across both. As before, DON-AR exhibits errors an order of magnitude larger than all other methods, which can be attributed to its high sensitivity to error accumulation, exacerbated by the limited availability of training data. Among the other DeepONet-based variants, the TI-based frameworks (TI(L)-DON and TI-DON) outperform DON-FR during the initial rollout (up to $t = 0.4$). Beyond this temporal horizon, DON-FR surpasses the TI-based methods and achieves lower test errors. Within the TI family, TI(L)-DON consistently outperforms the vanilla TI-DON due to its ability to adapt to local solution features, an advantage that becomes especially useful when training data is scarce. 

In contrast, the FNO-based approaches achieve significantly lower errors than the DeepONet-based methods, with FNO-FR marginally outperforming FNO-AR in this case. This behavior is expected, as the coupled Burgers’ PDE is more amenable to learning in spectral space, where dominant frequency modes of the solution dynamics can be effectively captured even with limited data. Nevertheless, it is worth noting that while FNO outperforms DeepONet here, neural operators in general remain data-hungry, and we hypothesize that FNO would require substantially more training data for PDEs with richer frequency spectra. Finally, compared to all operator-learning methods, GNS once again delivers the best performance, achieving an almost perfect match with the ground truth for both the $u$ and $v$ velocity fields.
\begin{table}[htbp!]
    \renewcommand{\arraystretch}{1.25}
    \setlength{\tabcolsep}{4pt}
    \centering
    \caption{\rev{2D coupled Burgers' equation with vector output field: Inference errors (mean $\pm$ std) for $u$ and $v$ fields across different methods and training trajectory counts. At inference, held-out set of 100 trajectories are used, with a timestep of $\Delta t = 0.01$ for the time-integrator-based frameworks (GNS, TI-DON, TI(L)-DON).}}
    \color{black}{
    \begin{tabular}{|>{\centering\arraybackslash}m{1.75cm}|
                    *{4}{>{\centering\arraybackslash}m{2.5cm}|}}
        \hline
        \multirow{2}{*}{Method} & \multicolumn{4}{c|}{Training Trajectories} \\
        \cline{2-5}
         & 30 & 50 & 70 & 100 \\
        \hline
        \multicolumn{5}{|c|}{Field $u$} \\
        \hline
        GNS & \bf{0.0068$\pm$0.001} & \bf{0.0047$\pm$0.0003} & \bf{0.0040$\pm$0.0002} & \bf{0.0035$\pm$0.0001} \\
        \hline
        DON FR & 0.721$\pm$0.037 & 0.448$\pm$0.043 & 0.338$\pm$0.016 & 0.243$\pm$0.007 \\
        \hline
        DON AR & 4.59$\pm$5.04 & 3.01$\pm$2.19 & 4.81$\pm$3.62 & 8.07$\pm$6.75 \\
        \hline
        TI-DON & 0.572$\pm$0.015 & 0.441$\pm$0.010 & 0.378$\pm$0.008 & 0.336$\pm$0.006 \\
        \hline
        TI(L)-DON & 0.579$\pm$0.014 & 0.439$\pm$0.004 & 0.380$\pm$0.006 & 0.337$\pm$0.009 \\
        \hline
        FNO AR & 0.102$\pm$0.019 & 0.047$\pm$0.011 & 0.028$\pm$0.005 & 0.018$\pm$0.003 \\
        \hline
        FNO FR & 0.067$\pm$0.010 & 0.037$\pm$0.005 & 0.027$\pm$0.002 & 0.017$\pm$0.001 \\
        \hline
        \multicolumn{5}{|c|}{Field $v$} \\
        \hline
        GNS & \bf{0.0063$\pm$0.0007} & \bf{0.0048$\pm$0.0004} & \bf{0.0040$\pm$0.0001} & \bf{0.0035$\pm$0.0002} \\
        \hline
        DON FR & 0.740$\pm$0.045 & 0.449$\pm$0.022 & 0.327$\pm$0.012 & 0.238$\pm$0.009 \\
        \hline
        DON AR & 2.15$\pm$1.08 & 7.18$\pm$4.05 & 5.81$\pm$3.94 & 6.41$\pm$6.62 \\
        \hline
        TI-DON & 0.540$\pm$0.012 & 0.422$\pm$0.011 & 0.349$\pm$0.006 & 0.310$\pm$0.008 \\
        \hline
        TI(L)-DON & 0.543$\pm$0.024 & 0.418$\pm$0.012 & 0.349$\pm$0.006 & 0.306$\pm$0.005 \\
        \hline
        FNO AR & 0.099$\pm$0.023 & 0.046$\pm$0.008 & 0.030$\pm$0.002 & 0.019$\pm$0.003 \\
        \hline
        FNO FR & 0.061$\pm$0.006 & 0.037$\pm$0.005 & 0.027$\pm$0.003 & 0.017$\pm$0.001 \\
        \hline
    \end{tabular}
    }
    \label{tab:2D_coupled_Burgers_results_summary}
\end{table}
\begin{figure}[htbp!]
    \centering
    \begin{subfigure}[b]{0.495\linewidth}
        \centering
        \includegraphics[width=\linewidth]{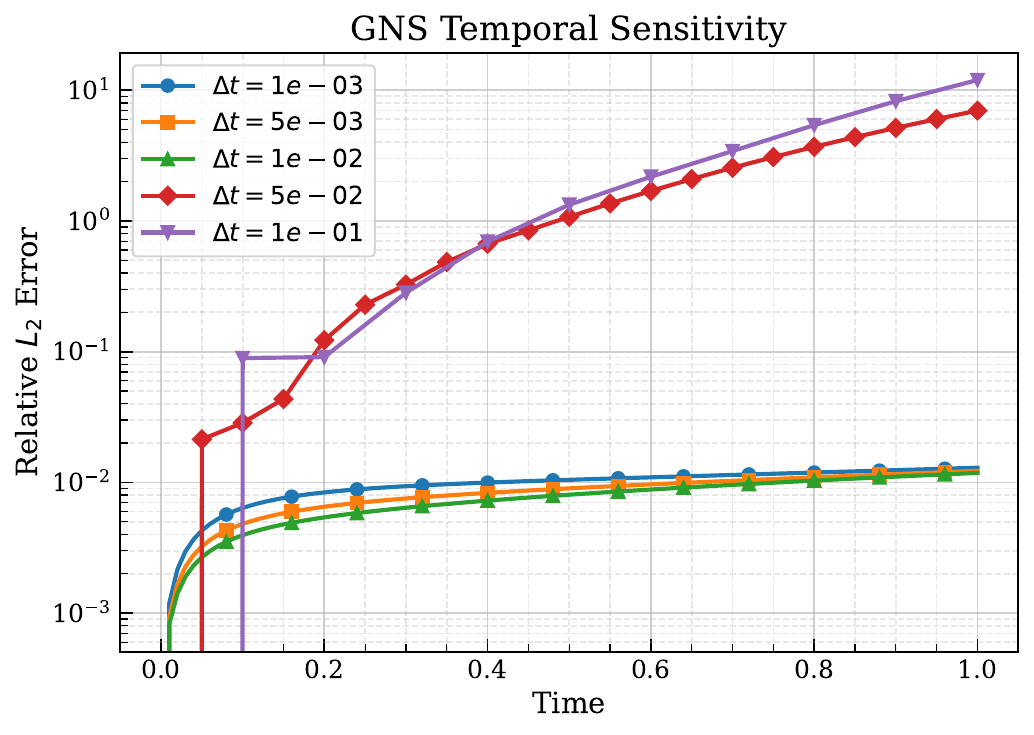}
        \caption{}
        \label{fig:2D_coupled_Burgers_dt_sensitivity}
    \end{subfigure}
    \hfill
    \begin{subfigure}[b]{0.495\linewidth}
        \centering
        \includegraphics[width=\linewidth]{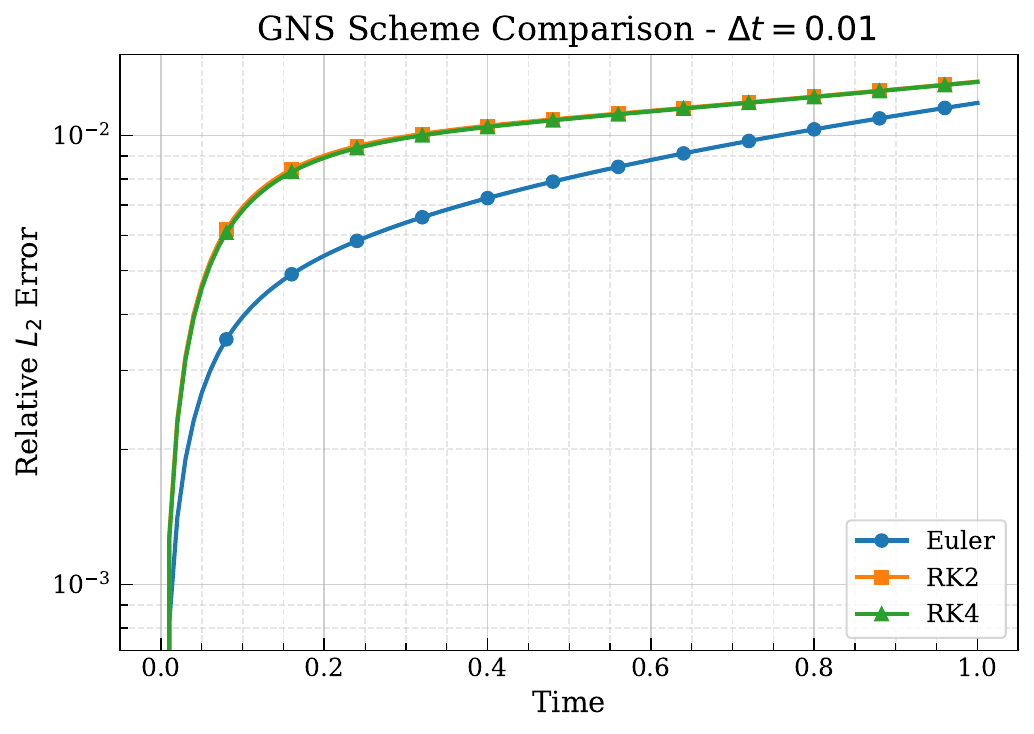}
        \caption{}
        \label{fig:2D_coupled_Burgers_scheme_comparison}
    \end{subfigure}
    \caption{\rev{2D coupled Burgers' equation with vector output field: (a) Temporal evolution of the relative $L_2$ error during inference of the GNS model for different timestep sizes $\Delta t \in \{0.001, 0.005, 0.01, 0.05, 0.1\}$, with a training timestep of $\Delta t = 0.01$. (b) Temporal evolution of the relative $L_2$ error across three time integration schemes (explicit Euler, RK2, and RK4) for a fixed inference timestep of $\Delta t = 0.01$.}}
    \label{fig:2D_coupled_Burgers_temporal_sensitivity_analysis}
\end{figure}

Table~\ref{tab:2D_coupled_Burgers_results_summary} summarizes the overall relative $L_2$ inference errors for predicting the $u$ and $v$ velocity fields using 30, 50, 70, and 100 training trajectories. The observed trends align closely with the qualitative comparisons in Figs.~\ref{fig:2D_coupled_Burgers_contours_u} and \ref{fig:2D_coupled_Burgers_contours_v}. As expected, DON-AR performs the worst, with relative $L_2$ errors exceeding $100\%$ across all cases, indicating that its predicted solutions deviate entirely from the ground truth. Among the remaining DeepONet-based frameworks, the TI variants (TI-DON and TI(L)-DON) achieve slightly lower errors than DON-FR when trained with only 30 trajectories and perform comparably for 50 and 70 trajectories. However, when sufficient training data is provided, DON-FR yields more accurate basis function and coefficient representations of the output solution space, ultimately surpassing the TI variants in predictive accuracy. 

The FNO-based approaches consistently outperform all DeepONet variants, achieving substantially lower test errors. In particular, FNO-FR demonstrates superior performance to FNO-AR for 30 and 50 training trajectories, while their accuracies remain similar at 70 and 100 trajectories. Finally, GNS achieves the best generalization performance overall, with errors below $1\%$ (0.9\% for $u$ and 0.78\% for $v$) even when trained on only 30 trajectories. This highlights GNS's inherent ability to naturally incorporate multiple interacting fields as node features, thereby capturing spatial interactions and coupling dynamics in a physically intuitive manner. Such an inductive bias makes GNS more physics-aware, enabling it to learn nonlinear, complex, and coupled dynamics effectively even in the low-data regime. \rev{The physics-based diagnostics for the 2D coupled Burgers' system are presented in Fig.~\ref{fig:2D_Burgers_vector_conservation_metrics}. This system carries two momentum quantities, $P_x = \int u \, d\Omega$ and $P_y = \int v \, d\Omega$, which remain essentially constant for the ground-truth trajectories (panels~(a) and~(b)), so their drift over the rollout serves as a conservation diagnostic. Panel~(c) shows that FNO-FR achieves the tightest momentum conservation at $\mathcal{O}(10^{-4})$, GNS follows at $\mathcal{O}(10^{-3})$, and FNO-AR together with the DeepONet variants sit at $\mathcal{O}(10^{-3})$ or worse. GNS's momentum violation is roughly two orders of magnitude larger than the mass violation in the scalar case, indicating that the coupled system is measurably harder for the one-step formulation to preserve.}

\rev{The kinetic energy $E(t) = \tfrac{1}{2}\int (u^2+v^2) \, d\Omega$ (panel~(f)) shows a distinct failure mode for DON-FR: its predicted energy is already incorrect at $t=0$, indicating that the fixed basis functions cannot faithfully represent the initial coupled velocity field. The TI-DON variants systematically under-dissipate, while GNS attains the lowest relative energy error, settling at $\mathcal{O}(10^{-2})$ below both FNO variants, which sit at $\mathcal{O}(10^{-2})$--$\mathcal{O}(10^{-1})$ (panel~(h)). The final-time energy spectrum (panel~(g)) exhibits the same structure as the scalar case: it peaks near $k \approx 10$, and the ground truth decays sharply on either side. In the low-wavenumber range below the peak, GNS follows the ground-truth spectrum closely on both fields, whereas the FNO variants fall below it, underrepresenting the large-scale energy content. In the high-wavenumber range, all surrogates overestimate the spectral energy relative to the sharply decaying ground truth, but GNS and FNO-AR remain closest, while the DeepONet variants stay nearly flat far into the high-$k$ range, indicating substantial spurious small-scale content. The pointwise $L_\infty$ errors on the $u$ and $v$ fields (panels~(d),~(e)) reinforce this ordering: GNS maintains $\mathcal{O}(10^{-3})$ throughout the rollout, the FNO variants settle at $\mathcal{O}(10^{-3})$--$\mathcal{O}(10^{-2})$, and the DeepONet variants sit two orders of magnitude higher. The overall trade-off mirrors the scalar case: FNO-FR conserves momentum most tightly (panel~(c)), whereas GNS is superior on energy conservation, pointwise $L_\infty$ error, and the fidelity of the spectral energy density across scales.}

\rev{The temporal sensitivity analysis in Fig.~\ref{fig:2D_coupled_Burgers_temporal_sensitivity_analysis}(a) shows a tighter empirical stability boundary than the scalar case, with the rollout diverging already at $\Delta t = 5 \times 10^{-2}$. The scheme comparison in Fig.~\ref{fig:2D_coupled_Burgers_temporal_sensitivity_analysis}~(b) reproduces the scalar-case observation that Euler outperforms RK2 and RK4 at the training $\Delta t$, for the same reason discussed earlier: the training objective supervises only the one-step Euler update, so higher-order stages inherit the model's one-step bias without offering compensating accuracy gains.}

\subsection{Two-Dimensional Allen–Cahn Equation}
\label{subsec:2d-allen-cahn}
The third example that we consider is the two-dimensional Allen–Cahn equation, which serves as a prototype model for phase separation and interface dynamics. It describes the evolution of an order parameter by balancing diffusion-driven smoothing with a nonlinear reaction term that enforces bistable equilibrium states. The governing equation is:
\begin{equation}
\dfrac{\partial u}{\partial t}
= \epsilon^2 \left( \dfrac{\partial^2 u}{\partial x^2} + \dfrac{\partial^2 u}{\partial y^2}\right) - (u^3 - u),
~~~~ \forall ~ (x,y,t) \in [0,1]^2 \times [0,1],
\end{equation}
where $u(x,y,t)$ is a scalar order parameter field, and $\epsilon = 0.05$ controls the interfacial width (or diffusion length). The initial condition, $u(x,y,0) = u_0(x,y)$, is sampled from two-dimensional Gaussian random fields, and periodic Dirichlet and Neumann boundary conditions are imposed along both spatial directions.

Mathematically, this is a nonlinear parabolic reaction–diffusion equation. The diffusion term tends to smooth spatial variations, while the cubic reaction term, $-(u^3 - u)$, drives the system toward the two stable equilibrium states $u = \pm 1$, leading to phase separation. The resulting dynamics are characterized by the formation and motion of interfaces, which evolve approximately by mean curvature flow and give rise to metastable domain patterns. For surrogate modeling, the Allen–Cahn equation poses two key challenges: (i) the strong nonlinearity of the cubic term amplifies prediction errors in long-horizon rollouts, and (ii) the interplay between sharp interfacial layers (set by $\epsilon$) and large-scale domain coarsening requires models to resolve both fine-scale structures and large-scale phase separation dynamics, demanding robustness across multiple spatial and temporal scales.

The dataset is generated by solving the two-dimensional Allen--Cahn equation on a periodic spatial domain $[0,1] \times [0,1]$, discretized using a uniform $32 \times 32$ grid. Temporal evolution is computed up to $T_{\mathrm{final}} = 1.0$ using an exponential time-differencing fourth-order Runge-Kutta (ETDRK4) scheme with a coarse time step $\Delta t = 0.01$, refined by a factor of 200 for numerical stability. The total number of coarse snapshots is $n_{\mathrm{coarse}} = 101$. Initial conditions are sampled from Gaussian-filtered random fields and scaled to the range $[-1,1]$. Specific implementation details are provided in Table~\ref{tab:dataset_generation_details}. Training trajectory selection follows the PCA-KMeans procedure employed in all other cases.
\begin{figure}[htbp!]
    \centering
    \includegraphics[width=\linewidth]{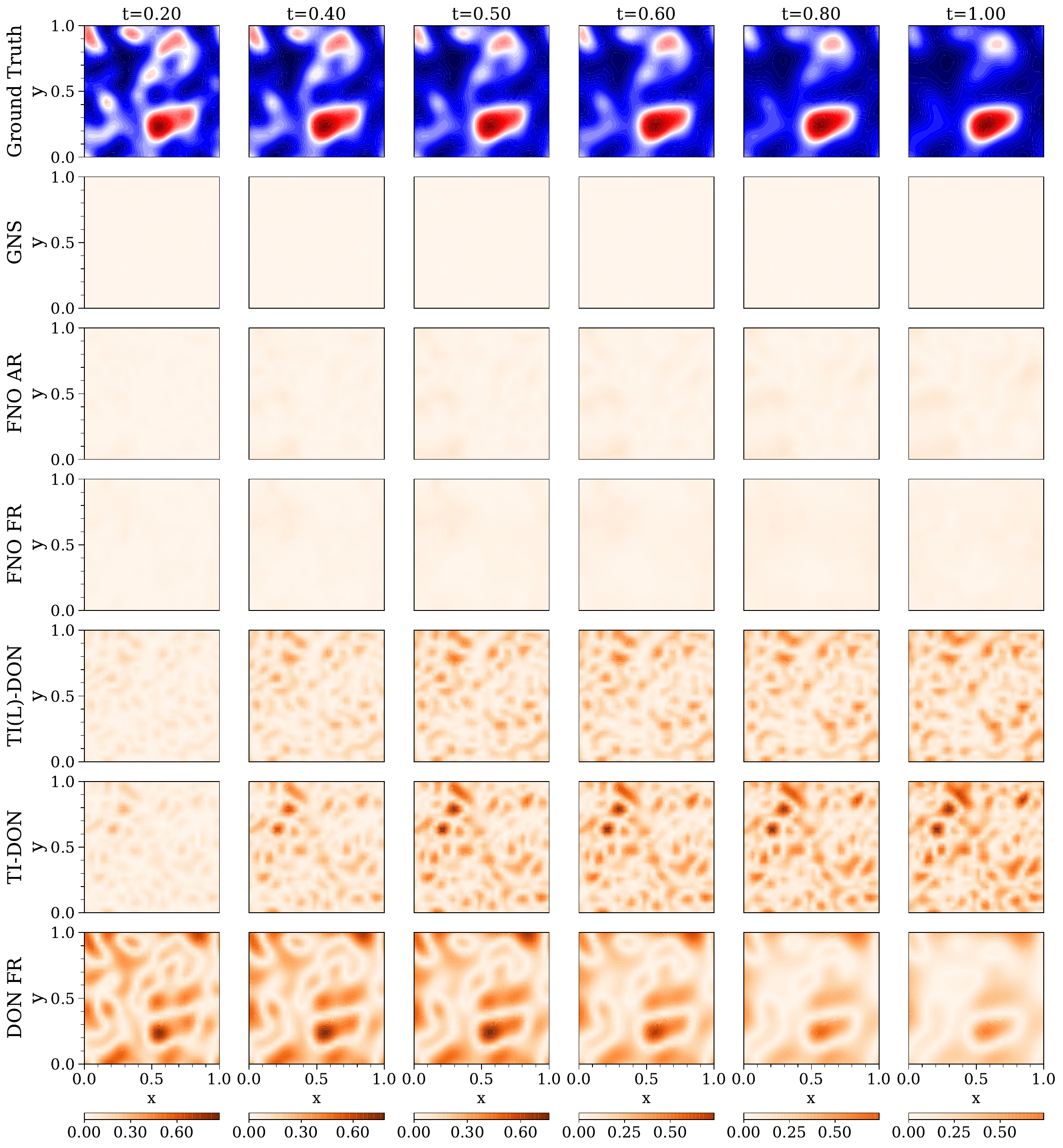}
    \includegraphics[width=\linewidth]{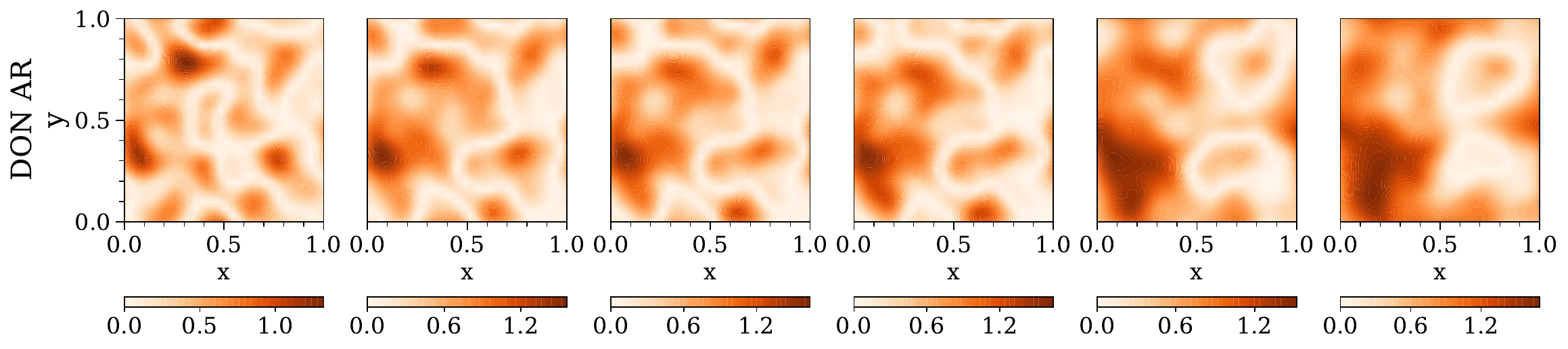}
    \caption{\rev{2D Allen-Cahn equation: Performance of all frameworks trained on 30 trajectories and evaluated on a held-out set of 200 trajectories. Top row: ground truth solution contours. Subsequent rows: absolute error contours for each framework, shown for a representative sample.}}
   \label{fig:2D_Allen_Cahn_contours}
\end{figure}

Figure~\ref{fig:2D_Allen_Cahn_contours} depicts the comparative performance of all frameworks for different time snapshots of a representative sample, using 30 training trajectories. The qualitative trends observed here are consistent with the previous two cases. DON-AR exhibits significantly larger errors compared to all other frameworks, highlighting its sensitivity to the accumulation of model approximation errors at each timestep, which is particularly pronounced in the low-data regime. Among the remaining DeepONet-based methods, the TI variants outperform DON-FR for timesteps up to $t = 0.6$, after which DON-FR surpasses TI(L)-DON and TI-DON, achieving lower absolute errors. Across all timesteps, the learnable TI variant, TI(L)-DON, consistently yields lower errors due to its adaptive numerical time-stepping scheme conditioned on the solution state at each timestep. In comparison, FNO significantly outperforms all DeepONet-based frameworks, producing far lower errors and closely matching the ground truth. Between the FNO variants, FNO-FR marginally outperforms FNO-AR and approaches the high accuracy of GNS. Finally, GNS delivers solution profiles that almost perfectly match the ground truth, even with a very limited number of training trajectories (\textit{e.g.} 30). This superior performance arises because GNS naturally captures spatial interactions and interface dynamics inherent to the Allen–Cahn equation, effectively encoding the phase separation and interfacial motion patterns as relational node features, which provides a strong inductive bias for learning complex, nonlinear dynamics in the low-data regime.
\begin{table}[htbp!]
    \renewcommand{\arraystretch}{1.25}
    \setlength{\tabcolsep}{4pt}
    \centering
    \caption{\rev{2D Allen-Cahn equation: Inference errors (mean $\pm$ std over 5 seeds) for $u$ field across different methods and training trajectory counts. At inference, held-out set of 200 test trajectories are used, with a timestep of $\Delta t = 0.01$ for the time-integrator-based frameworks (GNS, TI-DON, TI(L)-DON).}}
    \color{black}{
    \begin{tabular}{|>{\centering\arraybackslash}m{1.75cm}|
                    *{4}{>{\centering\arraybackslash}m{2.5cm}|}}
        \hline
        \multirow{2}{*}{Method} 
        & \multicolumn{4}{c|}{Training Trajectories} \\
        \cline{2-5}
         &  30 & 50 & 70 & 100 \\
        \hline
        \multirow{1}{*}{GNS}  
        & \bf{0.0035$\pm$0.00002} & \bf{0.0034$\pm$0.00001} & \bf{0.0034$\pm$0.000005} & \bf{0.0034$\pm$0.000001} \\
        \hline
        \multirow{1}{*}{DON FR}  
            & 0.884$\pm$0.042 & 0.771$\pm$0.017 & 0.731$\pm$0.017 & 0.679$\pm$0.018 \\
        \hline
        \multirow{1}{*}{DON AR}  
            & 2.13$\pm$0.629 & 2.38$\pm$0.451 & 3.07$\pm$0.388 & 3.35$\pm$0.394 \\
        \hline
        \multirow{1}{*}{TI-DON}  
            & 0.455$\pm$0.013 & 0.394$\pm$0.014 & 0.341$\pm$0.011 & 0.340$\pm$0.129 \\
        \hline
        \multirow{1}{*}{TI(L)-DON}  
            & 0.402$\pm$0.013 & 0.354$\pm$0.006 & 0.311$\pm$0.004 & 0.262$\pm$0.004 \\
        \hline
        \multirow{1}{*}{FNO AR}  
            & 0.040$\pm$0.008 & 0.016$\pm$0.002 & 0.010$\pm$0.001 & 0.006$\pm$0.0006 \\
        \hline
        \multirow{1}{*}{FNO FR}  
            & 0.020$\pm$0.003 & 0.011$\pm$0.002 & 0.007$\pm$0.0009 & 0.004$\pm$0.0003 \\
        \hline
    \end{tabular}
    }
    \label{tab:2D_Allen_Cahn_results_summary}
\end{table}
\begin{figure}[htbp!]
    \centering
    \begin{subfigure}[b]{0.495\linewidth}
        \centering
        \includegraphics[width=\linewidth]{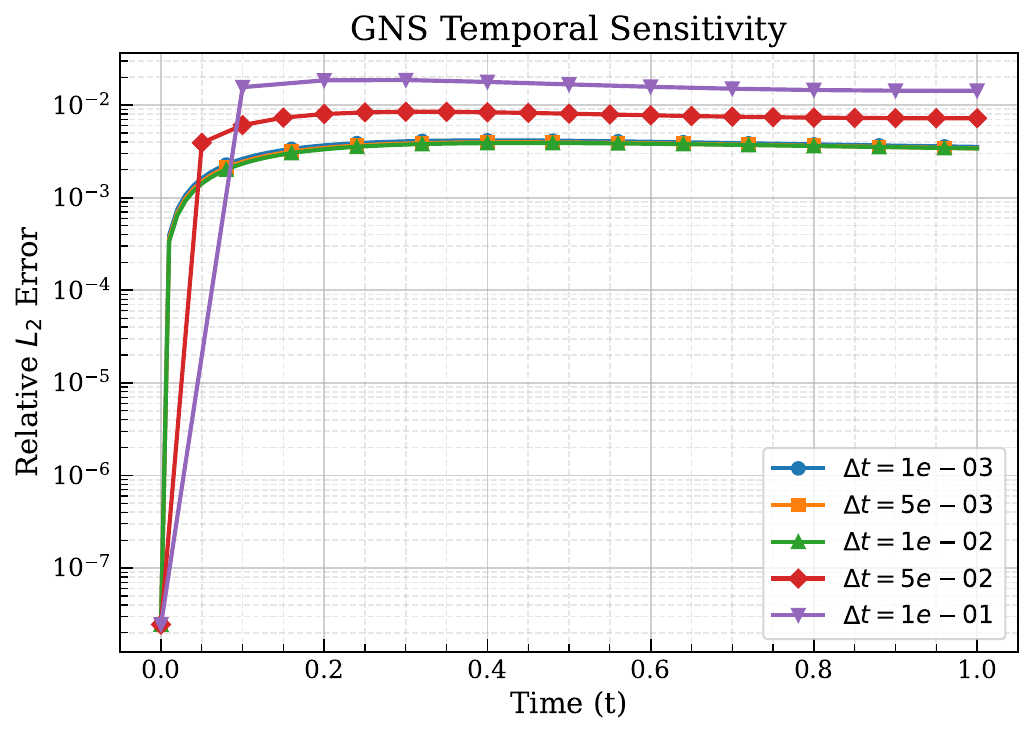}
        \caption{}
        \label{fig:2D_allen_cahn_dt_sensitivity}
    \end{subfigure}
    \hfill
    \begin{subfigure}[b]{0.495\linewidth}
        \centering
        \includegraphics[width=\linewidth]{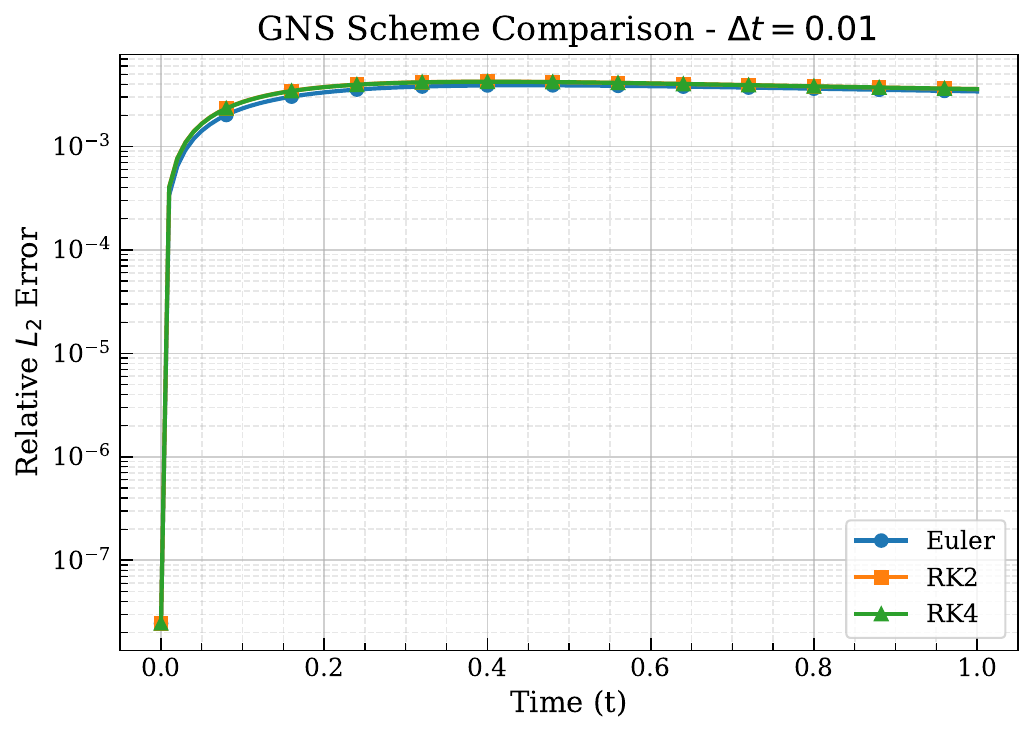}
        \caption{}
        \label{fig:2D_allen_cahn_scheme_comparison}
    \end{subfigure}
    \caption{\rev{2D Allen-Cahn equation: (a) Temporal evolution of the relative $L_2$ error during inference of the GNS model for different timestep sizes $\Delta t \in \{0.001, 0.005, 0.01, 0.05, 0.1\}$, with a training timestep of $\Delta t = 0.01$. (b) Temporal evolution of the relative $L_2$ error across three time integration schemes (explicit Euler, RK2, and RK4) for a fixed inference timestep of $\Delta t = 0.01$.}}
    \label{fig:2D_allen_cahn_temporal_sensitivity_analysis}
\end{figure}

Lastly, in line with the findings from the previous two cases, we quantitatively evaluate the relative $L_2$ test errors at inference on \rev{200 unseen test} trajectories, training on subsets of 30, 50, 70, and 100 trajectories (see Table~\ref{tab:2D_Allen_Cahn_results_summary}). As expected, DON-AR incurs the highest errors; however, compared to the earlier cases, the performance gap between DON-AR and the other frameworks is smaller, albeit its errors still exceed $100\%$. DON-FR follows with slightly better accuracy, though it remains inferior to the TI-based frameworks, \textit{\textit{i.e.}}, TI-DON and TI(L)-DON. Among these, the learnable counterpart consistently outperforms its vanilla version across all training subsets. Consistent with earlier observations, FNO significantly outperforms the DeepONet-based frameworks, achieving much lower test errors for all training subsets. Notably, both FNO-AR and FNO-FR nearly match the performance of GNS when trained on 100 trajectories. Finally, GNS once again establishes itself as the most robust and accurate framework, delivering superior performance over all others with relative errors as low as 0.4\%-even with scarce training data (30 trajectories).

\rev{The temporal sensitivity analysis for the 2D Allen-Cahn equation is shown in Fig.~\ref{fig:2D_allen_cahn_temporal_sensitivity_analysis}. Panel~(a) shows that the rollout error is essentially insensitive to $\Delta t$ for $\Delta t \le 10^{-2}$, degrades by roughly a factor of two at $\Delta t = 5 \times 10^{-2}$ and $10^{-1}$, but remains stable across the entire tested range without divergence. Panel~(b) shows that Euler, RK2, and RK4 yield nearly indistinguishable errors at the training $\Delta t$, indicating that the temporal integration scheme is not the bottleneck here and that the one-step model approximation error dominates.}

\subsection{Two-Dimensional Nonlinear Shallow Water Equations}
\label{subsec:2d-nonlin-swe}
The \rev{fourth} example that we consider is the two-dimensional nonlinear shallow water equations with three coupled fields: fluid column height $\eta(x, y, t)$ and velocity components $u(x, y, t)$ and $v(x, y, t)$. These equations are derived by depth-integrating the Navier-Stokes equations under the assumption that horizontal length scales greatly exceed vertical scales. This leads to nearly hydrostatic vertical pressure gradients and depth-constant horizontal velocities, allowing the vertical velocity to be eliminated through integration. While vertical velocity is removed from the equations, it can be recovered from the continuity equation once the horizontal solution is obtained. The governing equations are:
\begin{equation}
    \dfrac{\partial(\eta)}{\partial t} + \dfrac{\partial (\eta u)}{\partial x} + \dfrac{\partial (\eta v)}{\partial y} = 0,
\end{equation}
\begin{equation}
    \dfrac{\partial (\eta u)}{\partial t} + \dfrac{\partial}{\partial x}\left( \eta u^2 + \dfrac{1}{2} g\eta^2 \right) + \dfrac{\partial (\eta u v)}{\partial y} = \nu (u_{xx} + u_{yy}),
\end{equation}
\begin{equation}
    \dfrac{\partial (\eta v)}{\partial t} + \dfrac{\partial (\eta u v)}{\partial x} + \dfrac{\partial}{\partial y}\left(\eta v^2 + \dfrac{1}{2} g\eta^2 \right) = \nu (v_{xx} + v_{yy}).
\end{equation}

The system is solved on the spatiotemporal domain defined as: $(x, y) \in [0, 1] \times [0, 1]$ and $t \in [0, 1]$. The initial conditions are defined as follows. The total fluid column height $\eta(x, y, t)$, is given by a mean value of 1 plus some initial perturbation around a unit mean depth, $\eta_0(x, y)$. The initial velocity fields $u(x, y, t)$ and $v(x, y, t)$ are zero. We set the gravitational acceleration $g = 1$ and the viscosity coefficient $\nu = 0.002$. This problem presents a particularly challenging benchmark that combines multiple coupled fields with complex nonlinear dynamics. The shallow water equations serve as an excellent test case due to their physical relevance in modeling tsunami propagation and other geophysical flows. Further, these equations share structural similarities with compressible fluid dynamics, differing primarily in the absence of an energy equation dependent on the equation of state.

Mathematically, the shallow water equations form a nonlinear hyperbolic system of conservation laws. The advective fluxes capture nonlinear momentum transport, while the pressure-gradient and source terms govern gravity-driven wave propagation and topographic interactions. The dynamics are characterized by the coexistence of smooth wave-like solutions and sharp gradients, including hydraulic jumps, shocks, and bore formations. This multiscale behavior leads to complex flow patterns that are highly sensitive to initial and boundary conditions. For surrogate modeling, the shallow water equations pose two key challenges: (i) the strong nonlinearity of the advective terms amplifies errors over long-horizon rollouts, particularly in the presence of shocks and discontinuities, and (ii) the simultaneous need to capture fine-scale features such as bores and fronts, alongside basin-scale wave propagation and circulation, demands models that are robust across multiple spatial and temporal scales.
\begin{figure}[htbp!]
    \centering
    \includegraphics[width=\linewidth]{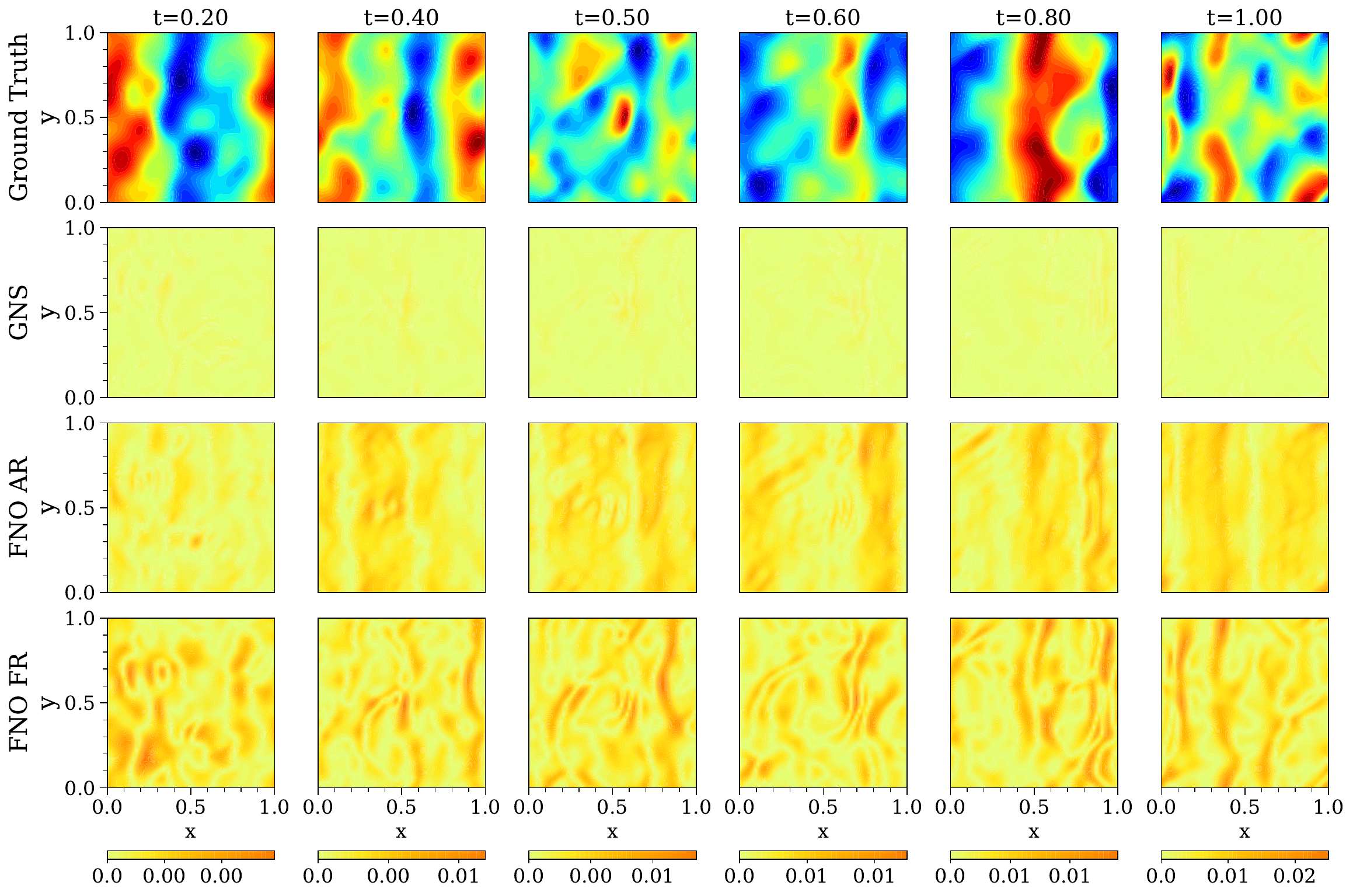}
    \caption{\rev{2D nonlinear shallow water equations: Performance of all frameworks trained on 50 trajectories and evaluated on a held-out set of 100 trajectories for predicting the $u$-velocity field. Top row: ground truth solution contours. Subsequent rows: absolute error contours for each framework, shown for a representative sample.}}
    \label{fig:2D_nonlinear_SWE_contours_u}
\end{figure}
\begin{figure}[htbp!]
    \centering
    \includegraphics[width=\linewidth]{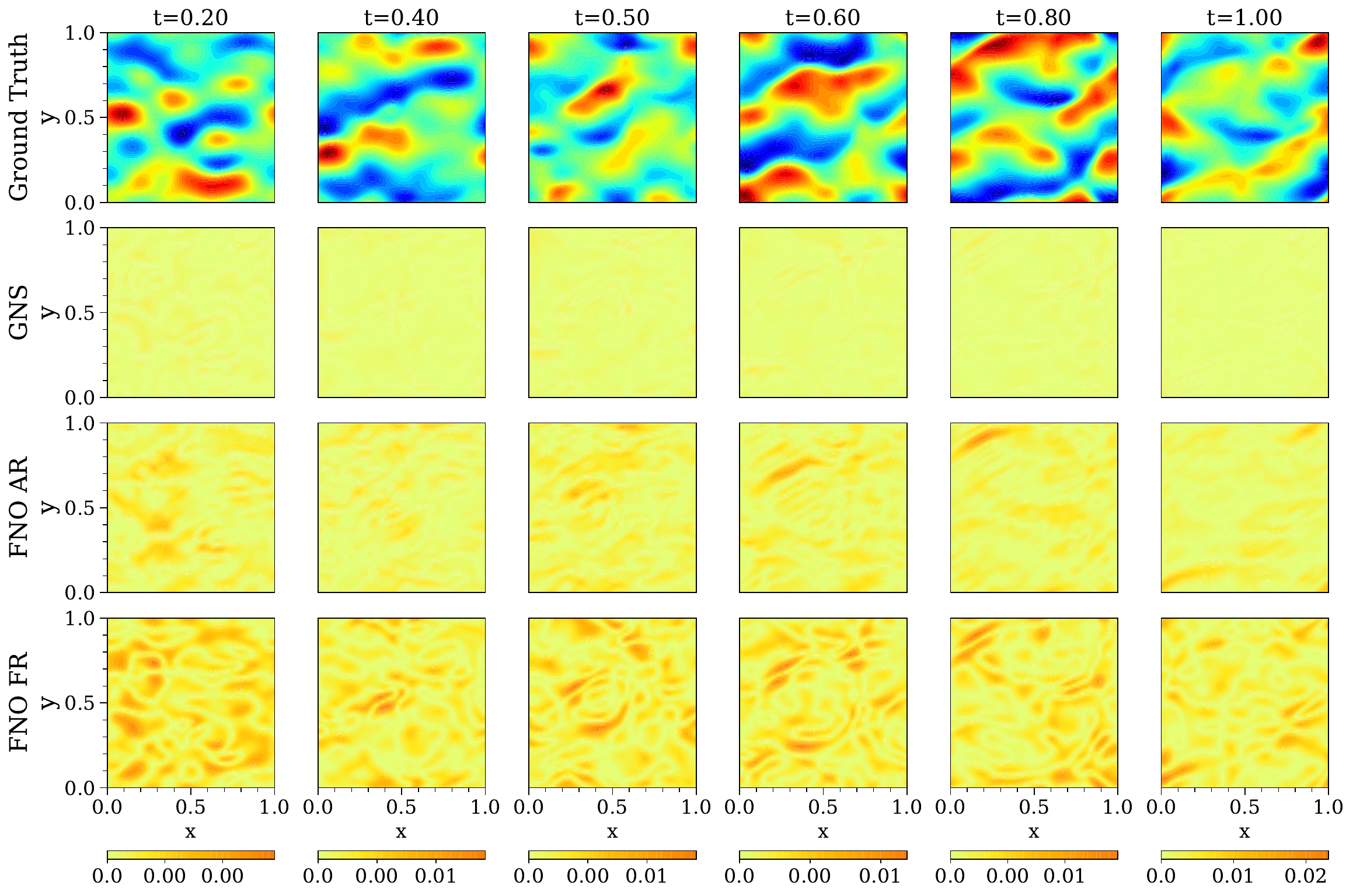}
    \caption{\rev{2D nonlinear shallow water equations: Performance of all frameworks trained on 50 trajectories and evaluated on a held-out set of 100 trajectories for predicting the $v$-velocity field. Top row: ground truth solution contours. Subsequent rows: absolute error contours for each framework, shown for a representative sample.}}
    \label{fig:2D_nonlinear_SWE_contours_v}
\end{figure}
\begin{figure}[htbp!]
    \centering
    \includegraphics[width=\linewidth]{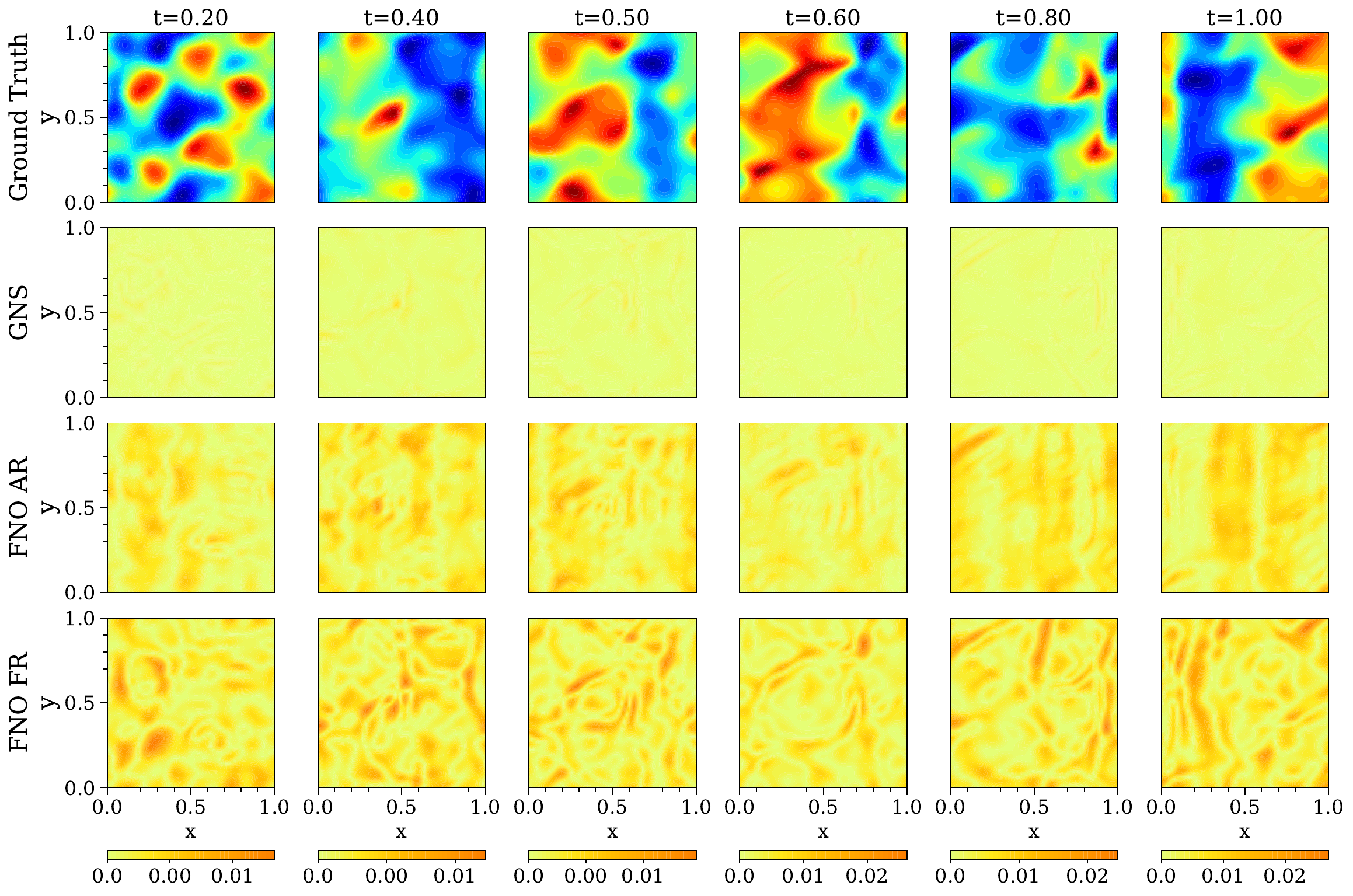}
    \caption{\rev{2D nonlinear shallow water equations: Performance of all frameworks trained on 50 trajectories and evaluated on a held-out set of 100 trajectories for predicting the $\eta$-height field. Top row: ground truth solution contours. Subsequent rows: absolute error contours for each framework, shown for a representative sample.}}
    \label{fig:2D_nonlinear_SWE_contours_h}
\end{figure}
\begin{figure}[htbp!]
    \centering
    \includegraphics[width=\linewidth]{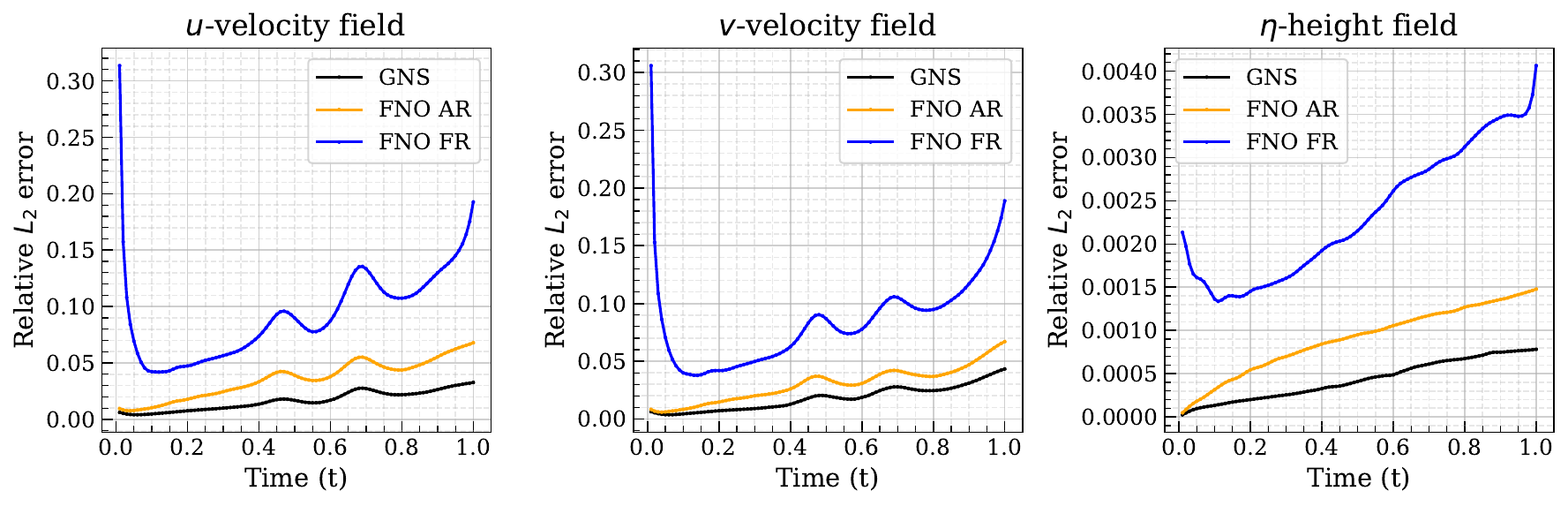}
    \caption{2D nonlinear shallow water equations: Temporal evolution of the relative $L_2$ error across GNS (black), FNO AR (orange), and FNO FR (blue).}
    \label{fig:err_acc_2D_SWE}
\end{figure}
\begin{figure}[htbp!]
    \centering
    \includegraphics[width=\linewidth]{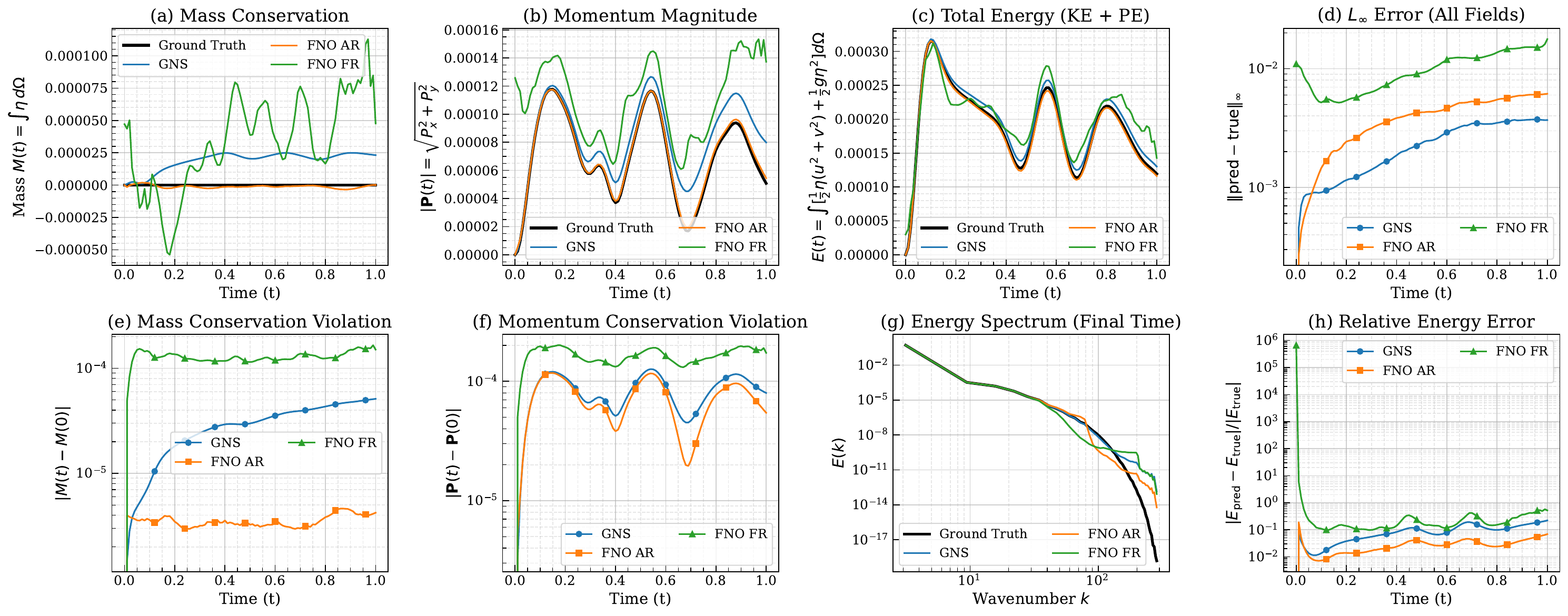}
    \caption{\rev{2D nonlinear shallow water equations: (a) Mass evolution over time, (b) momentum magnitude evolution over time, (c) total energy (kinetic and potential) evolution over time, (d) overall pointwise maximum $L_\infty$ error across all three fields, (e) mass conservation violation error, (f) momentum conservation violation error, (g) spectral energy density $E(k)$ as a function of wavenumber $k$ at the final timestep, and (h) energy conservation violation error, compared across different frameworks.}}
    \label{fig:2D_nonlinear_SWE_conservation_metrics}
\end{figure}
\begin{figure}[htbp!]
    \centering
    \includegraphics[width=\linewidth]{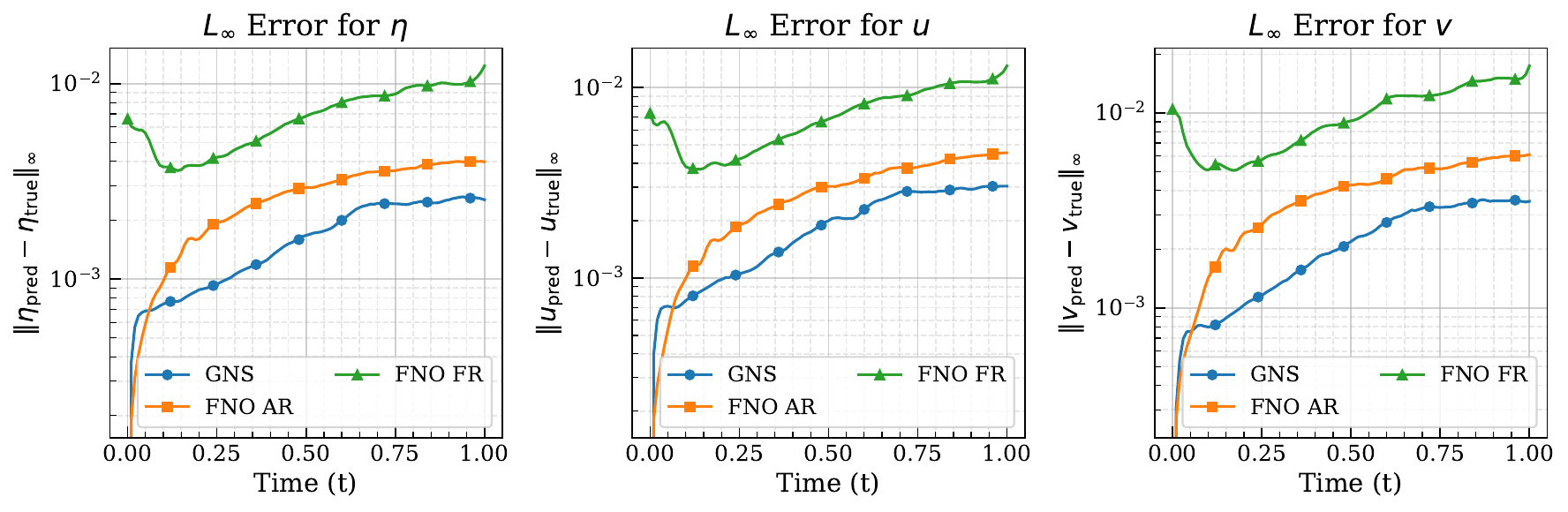}
    \caption{\rev{2D nonlinear shallow water equations: Temporal evolution of the pointwise maximum $L_\infty$ error for GNS (blue), FNO AR (orange), and FNO FR (green) across the height field $\eta$, $u$-velocity field, and $v$-velocity field (from left to right).}}
    \label{fig:2D_nonlinear_SWE_l_infinity_error}
\end{figure}
The ground truth dataset is generated following a strategy similar to those in cases~\ref{subsec:2d-burgers-scalar} and \ref{subsec:2d-burgers-vector}. As before, the initial conditions are sampled from smooth Gaussian random fields with Matérn-type kernels and periodic boundary conditions. Once again, to pick the most informative subset of training trajectories, the PCA-KMeans procedure was employed (see Section~\ref{subsec:2d-burgers-scalar}) by retaining the first 50 dominant principal vectors following the PCA dimensionality reduction step.

We now examine the performance of different surrogates for the three coupled fields: the velocity components $u$ and $v$, and the height field $\eta$. Figures~\ref{fig:2D_nonlinear_SWE_contours_u}, \ref{fig:2D_nonlinear_SWE_contours_v}, and \ref{fig:2D_nonlinear_SWE_contours_h} present a qualitative comparison of prediction errors for GNS, FNO AR, and FNO FR. All methods were trained on 50 trajectories and evaluated on \rev{100 unseen} test trajectories. Notably, the DeepONet variants exhibited poor performance on this problem, producing errors several orders of magnitude larger than the FNO variants and GNS. This degraded performance can be attributed to the initial conditions where both velocity fields $u$ and $v$ are exactly zero, creating a potentially challenging scenario for DeepONet's operator learning framework. Hence, for brevity, we focus our analysis only on the comparison between GNS and the two FNO variants (FNO AR and FNO FR). Between the FNO variants, FNO FR incurs larger errors than FNO AR. This can be attributed to the inherent difficulty of the FNO FR framework in trying to learn the solution operator from near-zero initial conditions (exactly zero $u$ and $v$ velocity fields). GNS performs the best out of all three frameworks, with the errors in a very small range and the predictions showing a near perfect match, with a modest training set of 50 trajectories. The rationale is that in spite of the near zero initial conditions, the evolution dynamics of $u$ and $v$ velocity fields is effortlessly handled by GNS. Primarily in the SWE case, the initial perturbation provided to the height field is what sets up or induces the velocity fields and these nonlinear interactions are extremely well captured by GNS owing to its ability to model such interactions in a more physically-consistent manner that closely mimics the true dynamics.
\begin{table}[htbp!]
    \renewcommand{\arraystretch}{1.2}
    \setlength{\tabcolsep}{6pt}
    \centering
    \caption{\rev{2D nonlinear shallow water equations: Inference errors (mean $\pm$ std over 5 seeds) for $u$, $v$, and $\eta$ fields across different methods and training trajectory counts. At inference, held-out set of 100 test trajectories are used, with a timestep of $\Delta t = 0.01$ for GNS.}}
    
    \color{black}{
    \begin{tabular}{|>{\centering\arraybackslash}m{1.4cm}|
                    *{4}{>{\centering\arraybackslash}m{2.55cm}|}}
        \hline
        \multirow{2}{*}{Method} & \multicolumn{4}{c|}{Training Trajectories} \\
        \cline{2-5}
         & 30 & 50 & 70 & 100 \\
        \hline
        \multicolumn{5}{|c|}{Field $u$} \\
        \hline
        GNS & \bf{0.0279$\pm$0.0061}  & \bf{0.0205$\pm$0.0154}  & \bf{0.0146$\pm$0.0040} & \bf{0.0113$\pm$0.0044}  \\
        \hline
        FNO AR & 0.0297$\pm$0.0030 & 0.0279$\pm$0.0072  & 0.0239$\pm$0.0030 & 0.0215$\pm$0.0025 \\
        \hline
        FNO FR & 0.1628$\pm$0.0126 & 0.1033$\pm$0.0090 & 0.0693$\pm$0.0049 & 0.0496$\pm$0.0037 \\ 
        \hline
        \multicolumn{5}{|c|}{Field $v$} \\
        \hline
        GNS & \bf{0.0250$\pm$0.0052}  & \bf{0.0156$\pm$0.0079}  & \bf{0.0132$\pm$0.0047} & \bf{0.0099$\pm$0.0060} \\
        \hline
        FNO AR & 0.0281$\pm$0.0019 & 0.0256$\pm$0.0063 & 0.0214$\pm$0.0018 & 0.0192$\pm$0.0020 \\
        \hline
        FNO FR & 0.1477$\pm$0.0106 & 0.0898$\pm$0.0074 & 0.0604$\pm$0.0040 & 0.0437$\pm$0.0026 \\
        \hline
        \multicolumn{5}{|c|}{Field $\eta$} \\
        \hline
        GNS & \bf{0.00060$\pm$0.000098}  & \bf{0.00036$\pm$0.00012}  & \bf{0.00031$\pm$0.000075} & \bf{0.00024$\pm$0.00013} \\
        \hline
        FNO AR & 0.00083$\pm$0.000065 & 0.00077$\pm$0.00019 & 0.00065$\pm$0.000061 & 0.00058$\pm$0.000065 \\
        \hline
        FNO FR & 0.00456$\pm$0.00032 & 0.00282$\pm$0.00023 & 0.00186$\pm$0.00014 & 0.00133$\pm$0.000095 \\
        \hline
    \end{tabular}
    }
    \label{tab:2D_nonlinear_SWE_results_summary}
\end{table}
\begin{figure}[htbp!]
    \centering
    \begin{subfigure}[b]{0.495\linewidth}
        \centering
        \includegraphics[width=\linewidth]{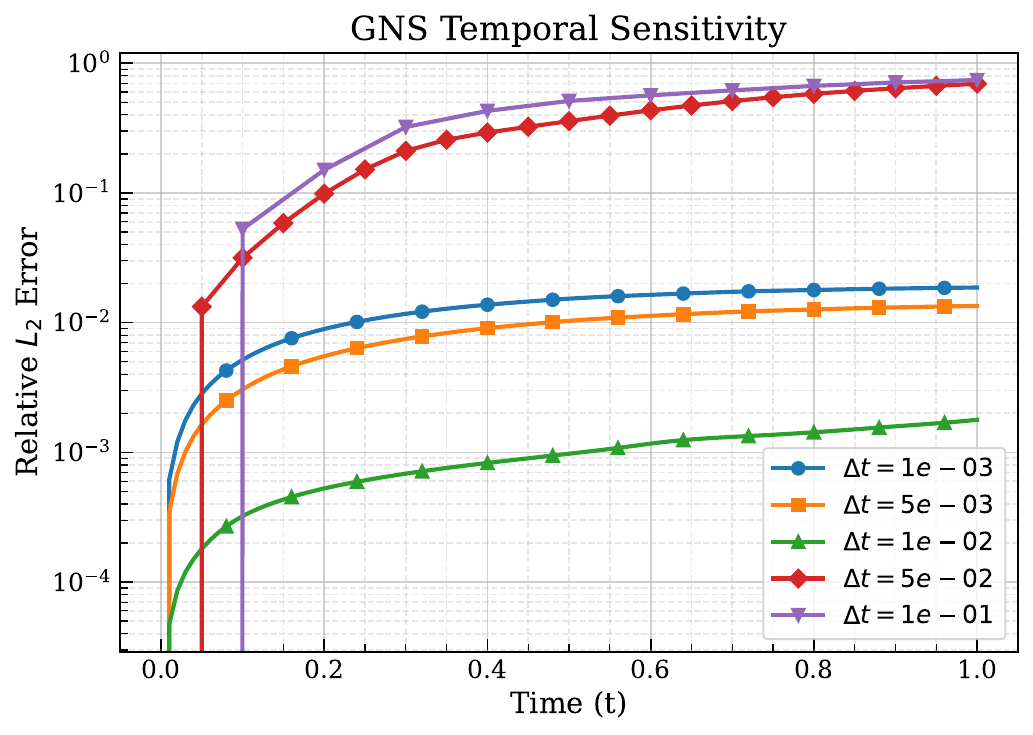}
        \caption{}
        \label{fig:2D_nonlinear_SWE_dt_sensitivity}
    \end{subfigure}
    \hfill
    \begin{subfigure}[b]{0.495\linewidth}
        \centering
        \includegraphics[width=\linewidth]{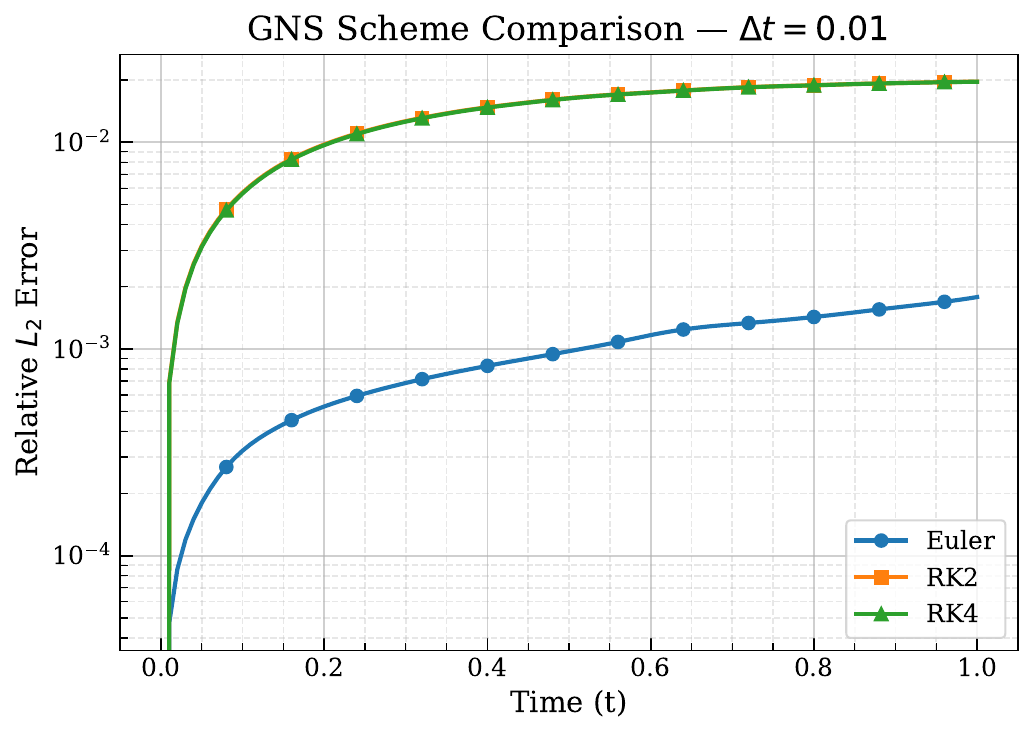}
        \caption{}
        \label{fig:2D_nonlinear_SWE_comparison}
    \end{subfigure}
    \caption{\rev{2D nonlinear shallow water equations: (a) Temporal evolution of the relative $L_2$ error during inference of the GNS model for different timestep sizes $\Delta t \in \{0.001, 0.005, 0.01, 0.05, 0.1\}$, with a training timestep of $\Delta t = 0.01$. (b) Temporal evolution of the relative $L_2$ error across three time integration schemes (explicit Euler, RK2, and RK4) for a fixed inference timestep of $\Delta t = 0.01$.}}
    \label{fig:2D_nonlinear_SWE_temporal_sensitivity_analysis}
\end{figure}

Table~\ref{tab:2D_nonlinear_SWE_results_summary} presents a comprehensive evaluation of the relative $L_2$ test errors for GNS and FNO-based variants trained with subsets of 30, 50, 70, and 100 trajectories. The quantitative error values exhibit strong agreement with the qualitative observations in Figs.~\ref{fig:2D_nonlinear_SWE_contours_u}, \ref{fig:2D_nonlinear_SWE_contours_v}, and \ref{fig:2D_nonlinear_SWE_contours_h}. Across all training subsets, GNS consistently delivers the best performance, achieving the lowest errors for all three coupled fields, $u$, $v$, and $\eta$. This is closely followed by FNO AR, which has slightly higher errors than GNS but performs substantially better than FNO FR, which exhibits the highest errors across all training subsets. The primary reason for the deterioration in FNO FR performance is the underlying operator learning problem it faces: mapping near-zero initial conditions (due to initial zero velocities) to the full spatiotemporal solution field, which, as established earlier, is an especially challenging task. 

We also assess the temporal error growth for the GNS and FNO frameworks in Fig.~\ref{fig:err_acc_2D_SWE}. In addition to exhibiting higher generalization accuracy in the limited-data regime, GNS (black) demonstrates stable and controlled error growth, indicating robustness and reliability in predicting accurate solutions over extended time horizons. FNO AR (orange) follows closely, while FNO FR (blue) shows comparatively higher error accumulation, consistent with the results in Table~\ref{tab:2D_nonlinear_SWE_results_summary}. Finally, the relatively large $L_2$ errors for the $u$ and $v$ velocity fields in Fig.~\ref{fig:err_acc_2D_SWE} for FNO FR can be attributed to the zero initial conditions and the difficulty FNO FR encounters in learning the coupling between the initial height perturbation and the subsequent evolution of the velocity fields. By contrast, GNS is able to effectively capture this coupling despite the zero initial velocity fields. \rev{The physics-based diagnostics for the 2D nonlinear SWE system are presented in Fig.~\ref{fig:2D_nonlinear_SWE_conservation_metrics}. This system carries the total mass $M = \int \eta \, d\Omega$ and the momentum components $P_x = \int \eta u \, d\Omega$, $P_y = \int \eta v \, d\Omega$, alongside the total energy $E = \int \left[\tfrac{1}{2}\eta(u^2+v^2) + \tfrac{1}{2}g\eta^2\right] d\Omega$ which combines kinetic and potential contributions. Given the poor performance of the DeepONet-based frameworks documented in the earlier benchmarks, this analysis focuses on GNS and the FNO variants. Panels~(a),~(e) show that FNO-AR achieves the tightest mass conservation at $\mathcal{O}(10^{-6})$, one order of magnitude better than GNS at $\mathcal{O}(10^{-5})$, while FNO-FR sits at $\mathcal{O}(10^{-4})$ with pronounced oscillations. For momentum (panels~(b),~(f)), all three frameworks remain at $\mathcal{O}(10^{-4})$, with FNO-AR the tightest followed by GNS and FNO-FR comparable. The total energy trajectory (panel~(c)) is tracked closely by all three frameworks, and the relative energy error (panel~(h)) places FNO-AR first at $\mathcal{O}(10^{-2})$, followed by GNS at $\mathcal{O}(10^{-2})$--$\mathcal{O}(10^{-1})$ and FNO-FR at $\mathcal{O}(10^{-1})$.}

\rev{The final-time energy spectrum (panel~(g)) shows that all three frameworks reproduce the ground-truth spectral decay closely across the entire resolved wavenumber range, with only minor deviations near the Nyquist limit; unlike the Burgers' cases, the SWE spectrum exhibits no pronounced high-$k$ plateau, and the FNO variants track the small-scale content nearly as well as GNS. The frameworks are instead separated by the pointwise $L_\infty$ error (panel~(d)), which confirms the same hierarchy as the earlier cases: GNS maintains $\mathcal{O}(10^{-3})$ throughout the rollout, FNO-AR sits at $\mathcal{O}(10^{-3})$--$\mathcal{O}(10^{-2})$, and FNO-FR an order of magnitude higher. The field-wise $L_\infty$ errors on $\eta$, $u$, and $v$ (Fig.~\ref{fig:2D_nonlinear_SWE_l_infinity_error}) exhibit the same ordering across all three fields. The overall picture for the SWE system is therefore mixed on the integral diagnostics: FNO-AR conserves mass most tightly and attains the lowest relative energy error, while all three frameworks reproduce the spectrum comparably; GNS retains its advantage on the pointwise $L_\infty$ error across every field.}

\rev{The temporal sensitivity analysis in Fig.~\ref{fig:2D_nonlinear_SWE_temporal_sensitivity_analysis}(a) shows that $\Delta t = 10^{-2}$ (the training step size) yields the lowest error. Rollouts at $5 \times 10^{-2}$ and $10^{-1}$ diverge outright, while the finer step sizes $10^{-3}$ and $5 \times 10^{-3}$ remain stable but yield larger errors than at the training $\Delta t$, indicating that departing from the training step size in either direction is detrimental. The scheme comparison in panel~(b) reproduces the pattern from the Burgers' cases: Euler outperforms RK2 and RK4 at the training $\Delta t$ by roughly an order of magnitude, for the same reason discussed earlier -- the training objective supervises only the one-step Euler update, so higher-order stages inherit the model's one-step bias without offering compensating accuracy gains.}

\subsection{\rev{Two-Dimensional Flow Past a Cylinder}}
\label{sec:cylinder_flow}
\rev{The preceding examples were defined on uniform Cartesian grids, which are compatible with the structured-input requirements of FNO and convolutional architectures. To evaluate the models on problems involving complex geometry and unstructured spatial discretizations, where graph-based surrogates are expected to hold a structural advantage, we consider viscous flow past a circular cylinder, a canonical benchmark in computational fluid dynamics.} \rev{The flow is governed by the two-dimensional incompressible Navier-Stokes equations,
\begin{align}
    \frac{\partial \mathbf{u}}{\partial t} + (\mathbf{u} \cdot \nabla)\mathbf{u} &= -\nabla p + \frac{1}{\mathrm{Re}}\nabla^2 \mathbf{u}, \label{eq:ns_momentum} \\
    \nabla \cdot \mathbf{u} &= 0, \label{eq:ns_continuity}
\end{align}
where $\mathbf{u} = (u, v)$ denotes the velocity field, $p$ is the pressure, and $\mathrm{Re}$ is the Reynolds number. The momentum equation~\eqref{eq:ns_momentum} is of advection-diffusion type, with the nonlinear advection term $(\mathbf{u} \cdot \nabla)\mathbf{u}$ and the viscous diffusion term $\mathrm{Re}^{-1}\nabla^2 \mathbf{u}$. Pressure does not evolve via its own time-dependent equation; instead, it is determined instantaneously at each time step by solving the elliptic Poisson equation that arises from enforcing the divergence-free constraint~\eqref{eq:ns_continuity}. This distinction is important for the surrogate formulation: velocity is the dynamical state that must be evolved autoregressively, while pressure is a diagnostic variable that can be predicted alongside velocity but should not be fed back as an input to the model.}

\rev{We use the \textit{CylinderFlow} dataset introduced by Pfaff et al.~\cite{pfaff2020learning}, generated using the COMSOL multiphysics solver. The dataset comprises 1000 training trajectories and 100 test trajectories, each containing 600 time steps at a temporal resolution of $\Delta t = 0.01$\,s. The computational domain is a rectangular channel with a circular cylinder obstacle. Crucially, the cylinder radius $R$, center position $(c_x, c_y)$, and mean inflow velocity $U_\mathrm{mean}$ vary across trajectories, producing a range of Reynolds numbers spanning both steady and unsteady (vortex-shedding) regimes. Because each trajectory has a different obstacle geometry, the triangular mesh is regenerated for each sample, resulting in meshes with approximately 1{,}886 nodes and 10{,}848 edges on average. The mesh resolution is finer near the cylinder surface, where boundary-layer gradients are steepest and coarser in the far field. Each node carries a velocity vector $(u, v)$ and a scalar pressure $p$ at every time step, along with a node-type label indicating whether it lies in the interior, on a wall, or at the inflow/outflow boundary.}

\rev{This dataset poses two challenges that the uniform-grid examples did not: (i) the spatial discretization is an unstructured triangular mesh that varies in both topology and node count across samples, and (ii) the geometry of the obstacle changes across trajectories, so the model must generalize across different flow configurations rather than learning a single fixed domain. These properties make CylinderFlow a natural testbed for evaluating the geometric flexibility of graph-based surrogates against neural operators that assume a fixed discretization.} \rev{For GNS, the unstructured mesh maps directly to a graph: mesh nodes become graph nodes and mesh edges become graph edges. The input node features consist of the normalized velocity $(u, v)$ at the current time step along with a one-hot encoding of the node type (interior, wall, inflow, outflow). Edge features consist of the relative position vector and Euclidean distance between connected nodes, normalized to zero mean and unit variance. Pressure is excluded from the input features and instead appears only as a target, consistent with the physics discussed above. The model predicts a normalized velocity increment $\Delta \mathbf{u}$ and the normalized next-step pressure $p(t+1)$. At inference, the velocity is updated as $\mathbf{u}(t+1) = \mathbf{u}(t) + \Delta \hat{\mathbf{u}}$ and fed back autoregressively; pressure is recorded but not recirculated. Boundary nodes are reset to their ground-truth values at each step.}

\rev{To improve rollout stability, we adopt the training-noise injection strategy of Pfaff et al.~\cite{pfaff2020learning}: during training, zero-mean Gaussian noise is added to the input velocity on interior nodes, and the same noise is subtracted from the velocity-increment target. This exposes the model to the type of perturbed inputs it will encounter during autoregressive rollout, where its own prediction errors accumulate over time.} \rev{For GINO, the same data pipeline is used (identical trajectory selection, normalization, and training noise), but the model architecture replaces the graph message-passing processor with a GNO encoder-FNO-GNO decoder chain~\cite{li2023geometry}. The GNO input encoder maps function values from the unstructured mesh nodes to a regular $128 \times 32$ latent grid via a radius-based integral transform, the FNO processes this grid in spectral space, and the GNO output decoder maps back to the mesh nodes. This architecture can handle unstructured inputs but relies on the intermediate regular grid for its spectral processing, in contrast to GNS which operates entirely on the native mesh topology.}

\rev{Training trajectories are selected from the 1000-trajectory training pool using the PCA-KMeans clustering strategy described before. Because the meshes vary across trajectories (training samples), we construct a mesh-invariant feature vector for each trajectory by computing per-timestep summary statistics (mean and standard deviation of $u$, $v$, and $p$), yielding a fixed-length representation suitable for PCA regardless of node count. The same selected trajectories are used for all models to ensure a controlled comparison. We evaluate on a held-out set of 20 unseen test trajectories drawn from the 100-trajectory test pool, using $T = 100$ time steps of autoregressive rollout, reporting the relative $L_2$ error on interior nodes for each field ($u$, $v$, $p$) separately.}
\begin{figure}[htbp!]
    \centering
    \includegraphics[width=\linewidth]{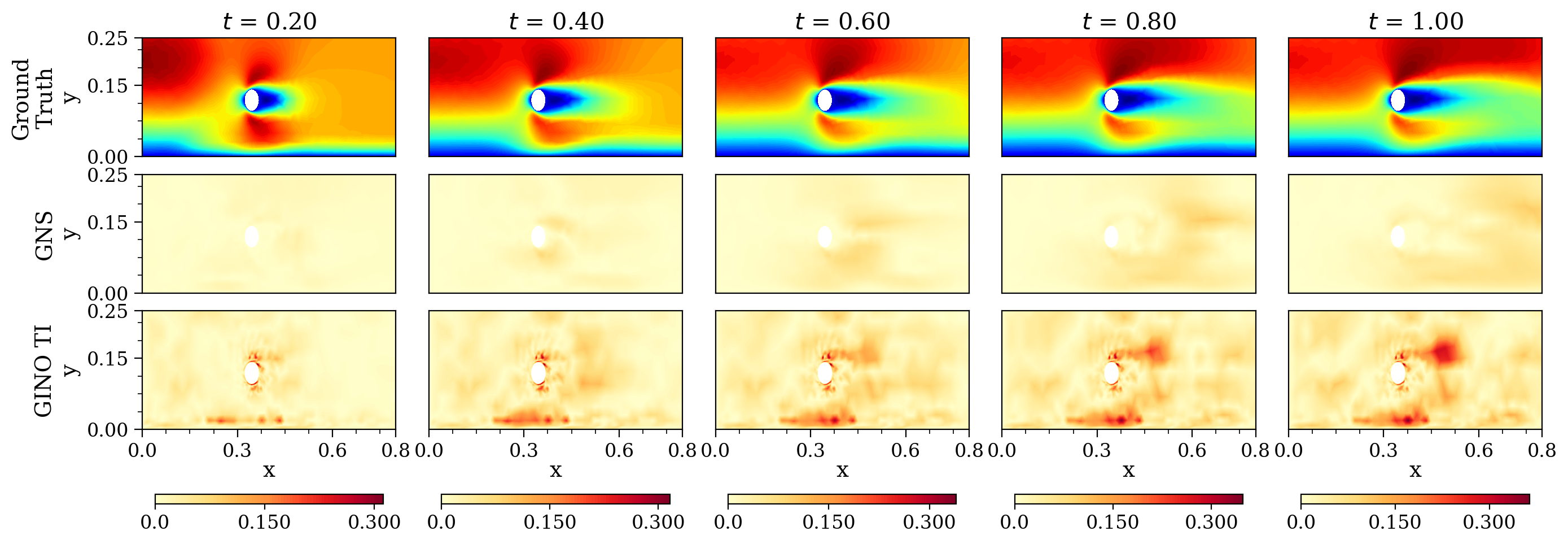}
    \caption{\rev{2D flow past a cylinder: Performance of all frameworks on 200 training trajectories, evaluated on 20 unseen test trajectories for predicting the $u$-velocity field of a representative sample. Top row: ground truth solution contours; subsequent rows: absolute error contours for each framework.}}
    \label{fig:2D_flow_past_cylinder_error_contours_u}
\end{figure}
\begin{figure}[htbp!]
    \centering
    \includegraphics[width=\linewidth]{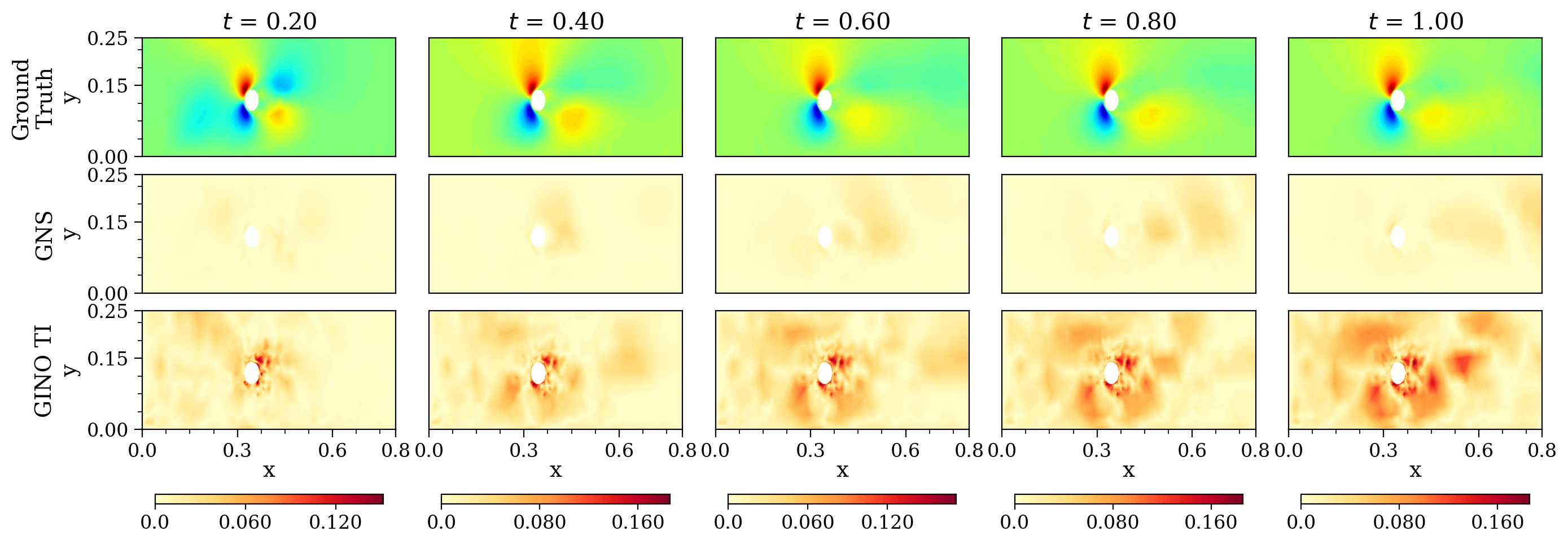}
    \caption{\rev{2D flow past a cylinder: Performance of all frameworks on 200 training trajectories, evaluated on 20 unseen test trajectories for predicting the $v$-velocity field of a representative sample. Top row: ground truth solution contours; subsequent rows: absolute error contours for each framework.}}
    \label{fig:2D_flow_past_cylinder_error_contours_v}
\end{figure}
\begin{figure}[htbp!]
    \centering
    \includegraphics[width=\linewidth]{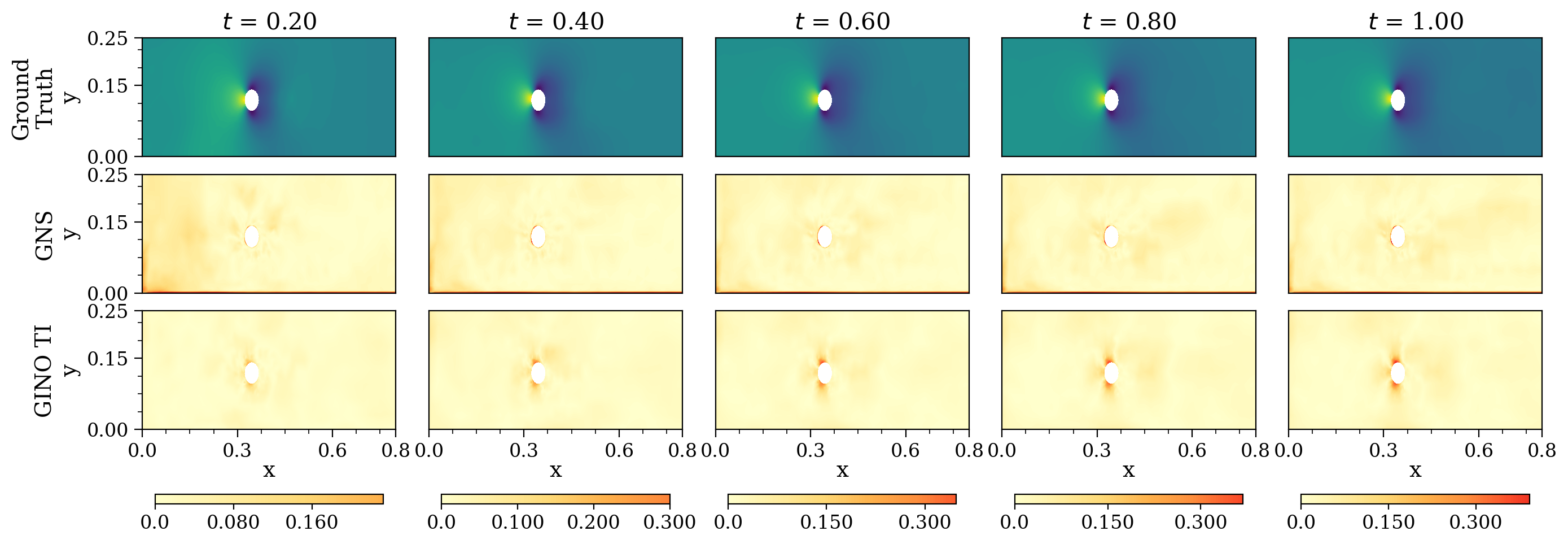}
    \caption{\rev{2D flow past a cylinder: Performance of all frameworks on 200 training trajectories, evaluated on 20 unseen test trajectories for predicting the pressure field, $p$, of a representative sample. Top row: ground truth solution contours; subsequent rows: absolute error contours for each framework.}}
    \label{fig:2D_flow_past_cylinder_error_contours_p}
\end{figure}
\begin{figure}[htbp!]
    \centering
    \includegraphics[width=\linewidth]{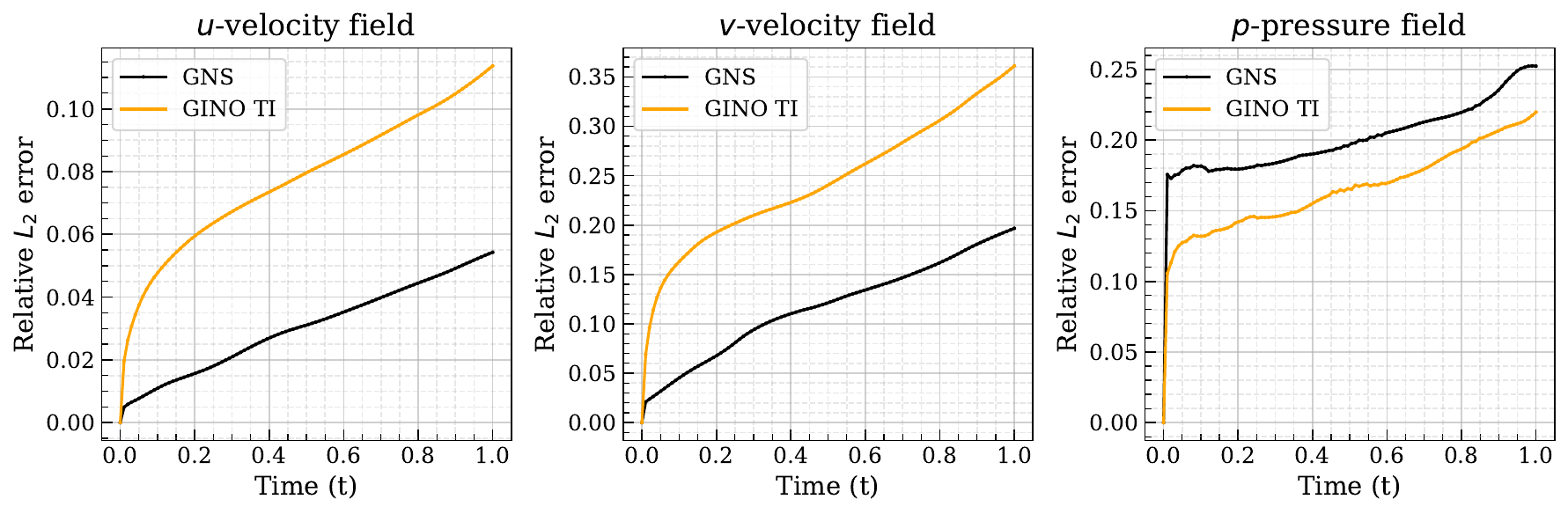}
    \caption{\rev{2D flow past a cylinder: Temporal evolution of the relative $L_2$ error across GNS (black) and GINO TI (orange).}}
    \label{fig:err_acc_2D_flow_past_cylinder}
\end{figure}
\begin{figure}[htbp!]
    \centering
    \includegraphics[width=\linewidth]{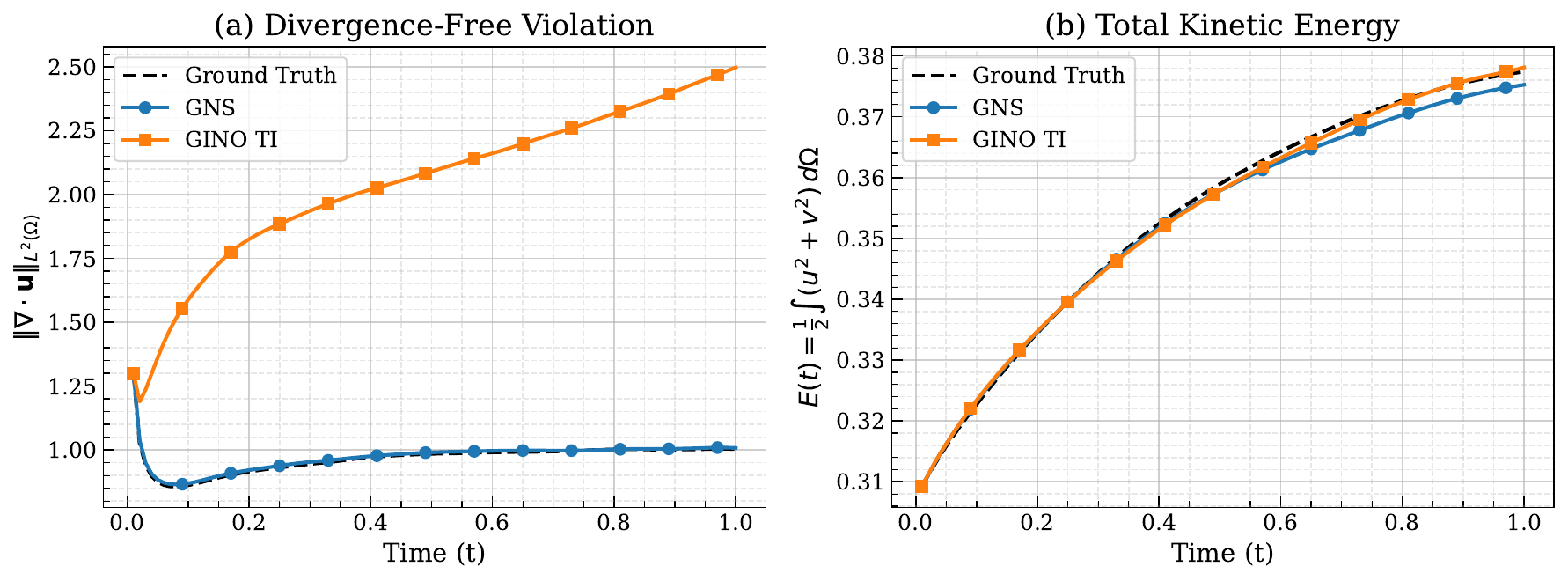}
    \caption{\rev{2D flow past a cylinder: (a) Violation of the incompressibility (divergence-free) condition and (b) total kinetic energy evolution over time, compared across GNS (blue) and GINO TI (orange) frameworks.}}
    \label{fig:violation_errs_2D_flow_past_cylinder}
\end{figure}
\begin{figure}[htbp!]
    \centering
    \includegraphics[width=\linewidth]{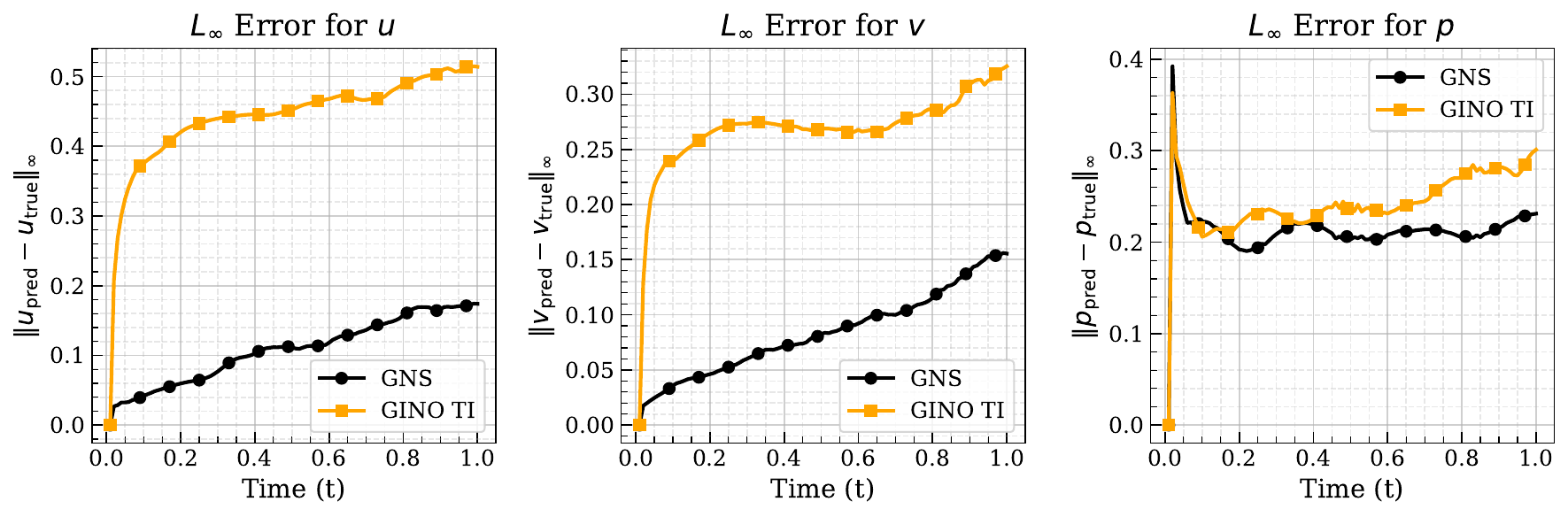}
    \caption{\rev{2D flow past a cylinder: Temporal evolution of the pointwise maximum $L_\infty$ error for GNS (black) and GINO TI (orange) across the $u$-velocity field, $v$-velocity field, and pressure field $p$ (from left to right).}}
    \label{fig:linf_err_acc_2D_flow_past_cylinder}
\end{figure}
\begin{table}[htbp!]
    \renewcommand{\arraystretch}{1.2}
    \setlength{\tabcolsep}{6pt}
    \centering
    \caption{\rev{2D flow past a cylinder: Inference errors (mean $\pm$ std over 5 seeds) for $u$, $v$, and $p$ fields across different methods and training trajectory counts. At inference, held-out set of 20 unseen test trajectories are used, with a timestep of $\Delta t = 0.01$ for GNS and GINO TI.}}
    \color{black}{
    \begin{tabular}{|>{\centering\arraybackslash}m{1.4cm}|
                    *{4}{>{\centering\arraybackslash}m{2.5cm}|}}
        \hline
        \multirow{2}{*}{Method} & \multicolumn{4}{c|}{Training Trajectories} \\
        \cline{2-5}
         & 80 & 100 & 200 & 500 \\
        \hline
        \multicolumn{5}{|c|}{Field $u$} \\
        \hline
        GNS & \bf{0.0661 $\pm$ 0.0023} & \bf{0.0644 $\pm$ 0.0105} & \bf{0.0382 $\pm$ 0.0029} & \bf{0.0244 $\pm$ 0.0032} \\
        \hline
        GINO TI & 0.1479 $\pm$ 0.02 & 0.1272 $\pm$ 0.0169 & 0.0797 $\pm$ 0.0043 & 0.0551 $\pm$ 0.001 \\
        \hline
        \multicolumn{5}{|c|}{Field $v$} \\
        \hline
        GNS & \bf{0.2503 $\pm$ 0.0326} & \bf{0.2117 $\pm$ 0.0438} & \bf{0.1267 $\pm$ 0.0055} & \bf{0.0832 $\pm$ 0.0134} \\
        \hline
        GINO TI & 0.4108 $\pm$ 0.0283 & 0.3492 $\pm$ 0.0342 & 0.2552 $\pm$ 0.0123 & 0.1801 $\pm$ 0.0057 \\
        \hline
        \multicolumn{5}{|c|}{Field $p$} \\
        \hline
        GNS & 0.3186 $\pm$ 0.0158 & 0.2899 $\pm$ 0.0379 & 0.186 $\pm$ 0.0116 & 0.1214 $\pm$ 0.0086 \\
        \hline
        GINO TI & \bf{0.2514 $\pm$ 0.023} & \bf{0.2159 $\pm$ 0.0204} & \bf{0.1413 $\pm$ 0.0128} & \bf{0.0971 $\pm$ 0.0026} \\
        \hline
    \end{tabular}
    }
    \label{tab:2D_flow_past_cylinder_results_summary}
\end{table}

\rev{Figures~\ref{fig:2D_flow_past_cylinder_error_contours_u}, \ref{fig:2D_flow_past_cylinder_error_contours_v}, and \ref{fig:2D_flow_past_cylinder_error_contours_p} show the ground truth solution alongside the absolute error contours for GNS and GINO~TI at five representative time instants during the autoregressive rollout of a held-out test trajectory. The comparison reveals a consistent pattern across the two velocity components and a distinct picture for the pressure field. The two models embody fundamentally different inductive biases about how information propagates spatially. GNS learns through iterative message passing over the mesh, so information at any node is assembled from a series of local exchanges with its immediate neighbors. This is well aligned with the discretized structure of the PDE itself, in which the time evolution of any state variable depends on a local stencil of neighboring quantities. GINO, on the other hand, lifts the mesh solution onto a fixed regular grid via its GNO encoder and processes it in Fourier space, which is inherently a global operation: every mode couples every point in the domain. This mismatch between global processing and the local structure of the underlying physics is precisely the limitation we identified for neural operators in general, and it manifests clearly in the results that follow.}

\rev{For the $u$-velocity field (Figure~\ref{fig:2D_flow_past_cylinder_error_contours_u}), GNS maintains uniformly low error across the entire domain at all time instants. The residual error is diffuse and lacks any clear spatial structure, indicating that GNS captures both the free-stream flow and the cylinder wake with high fidelity. GINO~TI, by contrast, exhibits localized error concentrations that grow with time. The most pronounced errors appear near the cylinder surface (where the boundary layer is thinnest and gradients are steepest) and along the lower channel wall, with additional structure emerging in the far wake at $t = 0.80$ and $t = 1.00$. This spatial pattern is consistent with the mechanism outlined above: the velocity field is driven by local convection and viscous shear near the cylinder, and message passing on the native mesh directly follows these local pathways of information transfer. GINO's spectral processor, operating on a coarser regular latent grid, cannot represent the sharp gradients in these regions with comparable fidelity. The $v$-velocity field (Figure~\ref{fig:2D_flow_past_cylinder_error_contours_v}) shows the same qualitative picture, but the contrast between the two models is more pronounced. GNS again produces low, spatially unstructured error. GINO~TI shows substantially higher errors that concentrate around the cylinder and propagate into the near wake, with the affected region visibly growing over time. Because the $v$-velocity captures the transverse dynamics associated with vortex shedding, the errors here directly correspond to inaccuracies in the shedding pattern and its downstream evolution. Vortex shedding is a locally driven instability (born at the separation points on the cylinder and advected into the wake) and is exactly the kind of dynamics that local message passing handles well. GINO's global Fourier representation blurs the sharp localized structures that drive shedding, and because the field is oscillatory and phase-sensitive, these representational errors amplify over the autoregressive rollout.}

\rev{The pressure field (Figure~\ref{fig:2D_flow_past_cylinder_error_contours_p}) offers a useful sanity check on this mechanistic interpretation. Both models exhibit error concentrations near the cylinder surface, where the pressure gradient is sharpest and the ground-truth solver itself carries the greatest numerical uncertainty, but GINO~TI produces a modestly lower error in the surrounding wake and free-stream region. This is consistent with the physics of the pressure field: unlike the velocity components, pressure is not a dynamical state but a derived field, determined instantaneously from the velocity by an elliptic Poisson equation. Elliptic solves are inherently global, and a Fourier basis is naturally aligned with the eigenfunctions of the Laplacian, so on this one diagnostic quantity GINO's global processing is well matched to the physics. Importantly, this is the exception rather than the rule: the pressure is a derived field that does not evolve in time, and no autoregressive error accumulates on it in the way it does on the velocity components. On the fields that actually evolve in time, GNS's locality provides a decisive advantage.}

\rev{The temporal evolution of the relative $L_2$ error, shown in Figure~\ref{fig:err_acc_2D_flow_past_cylinder}, corroborates this picture. For both velocity components, GNS maintains a lower error than GINO~TI at every instant, with the gap widening as the rollout proceeds. GINO's error rises steeply within the first few time steps and then continues to grow, whereas GNS shows a slower, near-linear accumulation. The $v$-field curve makes the contrast most stark: by $t = 1.0$, GINO's error is roughly twice that of GNS. For the pressure field, the ordering flips, but both curves grow at similar rates and are separated by only a small offset; the pressure comparison is best characterized as a modest GINO advantage on a diagnostic quantity, in contrast to the wide margins that separate the models on the two fields governed by time-dependent evolution equations.}

\rev{Beyond the field-wise error contours and the relative $L_2$ growth, we assess the physics-based diagnostics for the flow past cylinder case in Figs.~\ref{fig:violation_errs_2D_flow_past_cylinder} and \ref{fig:linf_err_acc_2D_flow_past_cylinder}. Because the flow is incompressible, the governing physical constraint is the divergence-free condition $\nabla \cdot \mathbf{u} = 0$; we therefore track $\|\nabla \cdot \mathbf{u}\|_{L^2(\Omega)}$ over the rollout as the primary conservation diagnostic, alongside the total kinetic energy $E(t) = \tfrac{1}{2}\int (u^2+v^2) \, d\Omega$. Note that the ground-truth divergence measured on the discrete mesh is not identically zero but takes a small finite value, so the relevant question is whether each surrogate tracks the ground-truth level rather than departs from it. Panel~(a) of Fig.~\ref{fig:violation_errs_2D_flow_past_cylinder} shows that GNS follows the ground-truth divergence norm closely throughout the rollout, whereas GINO~TI drifts monotonically upward, reaching roughly $2.5\times$ the ground-truth value by $t=1$. This indicates that the GINO~TI rollout progressively violates the incompressibility constraint, injecting spurious compressibility into a flow that should remain divergence-free to within the discretization error. Since the divergence is computed directly from the predicted velocity components, this behavior follows naturally from GNS's higher accuracy on the $u$ and $v$ fields: more accurate velocities yield a divergence field that adheres more closely to the ground-truth constraint. The total kinetic energy (panel~(b)) is by contrast tracked closely by both frameworks, with only a marginal difference at late times; the integrated energy is thus a far less discriminating diagnostic than the divergence norm for this problem.}

\rev{The field-wise pointwise $L_\infty$ errors in Fig.~\ref{fig:linf_err_acc_2D_flow_past_cylinder} reinforce the ordering seen in the contour plots. For both velocity components, GNS maintains a substantially lower $L_\infty$ error than GINO~TI at every instant, with the gap widening over the rollout: by $t=1$, the GINO~TI error is roughly three times that of GNS on the $u$-field and twice that on the $v$-field. The pressure field is more subtle. As noted in the contour discussion, GINO~TI attains a lower relative $L_2$ error on pressure, yet in the $L_\infty$ sense the two frameworks are comparable, with GNS in fact holding a slight edge through most of the rollout. This reversal between the two metrics is itself informative: $L_2$ is averaged over the spatial domain and favors GINO~TI's global spectral representation, which captures the smooth large-scale structure of the pressure field, whereas $L_\infty$ is a pointwise worst-case measure that is amenable to GNS's stronger resolution of sharp local features. The residual GINO~TI error may be further compounded by the interpolation incurred in its GNO encode/decode between the unstructured mesh and the regular latent grid, which tends to smooth small-scale features. Taken together, GNS holds a decisive advantage on the velocity components that carry the time-dependent dynamics, while the pressure, a diagnostic field governed by a global elliptic constraint, is the one quantity on which the two frameworks are closely matched, each favored under a different error norm.}

\rev{Taken together, these observations reinforce the central thesis of this work. Physical systems governed by time-dependent PDEs have local structure at every step of their evolution, and surrogate architectures that respect this locality generalize more reliably than those that impose a global processing bias. GNS's mesh-native message passing directly mirrors the local stencils used in classical numerical solvers, and this alignment translates into markedly lower rollout errors on the velocity components that carry the actual dynamics of the flow past cylinder problem. The one field on which GINO holds an edge, pressure, is precisely the diagnostic quantity determined by a globally coupled elliptic equation, which is not representative of the kind of local, evolutionary dynamics that dominate scientific PDE simulation. This result extends the conclusions of the preceding uniform-grid benchmarks to the more challenging setting of unstructured meshes and complex geometry, and demonstrates that the data efficiency and rollout stability advantages of GNS are not tied to any particular discretization but stem from the graph-based locality inherent to the framework.}

\section{Conclusions}
\label{sec:conclusions}
In this work, we performed a systematic comparison of generalization performance of various NO frameworks, including DeepONet and FNO-based variants with Graph Neural Simulators (GNS), on \rev{five} representative two-dimensional PDEs: the viscous Burgers' equation, the coupled Burgers' system, the Allen-Cahn equation, the nonlinear shallow water equations, \rev{and the flow past cylinder case}. Across all cases, we evaluated both qualitative solution contours and quantitative relative $L_2$ errors under varying amounts of training data. \rev{To assess physical consistency beyond the averaged $L_2$ metric, we additionally examined a suite of physics-based diagnostics, including the temporal violation of mass, momentum, and energy conservation, the pointwise $L_\infty$ error, and the final-time kinetic energy spectrum $E(k)$ as a function of the wavenumber $k$.} Our results demonstrate that DeepONet-based methods are more sensitive to limited training data compared to other NOs, with DON-AR consistently exhibiting the largest errors due to the steep accumulation of approximation errors over time, which is further exacerbated in low-data regimes. TI-based DeepONet variants improve upon the generalization performance of DON-FR, particularly when training data is very limited. However, with sufficient training trajectories, DON-FR surpasses the TI variants due to a more accurate representation of the output solution space, enabled by the increased availability of data. FNO frameworks consistently outperform all DeepONet variants, especially for PDEs whose solutions are dominated by low-frequency modes, making them particularly amenable to spectral-space learning and enabling efficient capture of the dominant frequency modes that govern the PDE dynamics with relatively few training samples. \rev{The physics-based diagnostics, however, reveal a more nuanced picture among the best-performing frameworks: the autoregressive FNO variant is highly competitive on the integral quantities, and on some benchmarks conserves mass or energy better than GNS, whereas GNS more consistently reproduces the fine-scale structure of the solution, most visibly in the large-scale spectral content of the Burgers' cases and in the pointwise $L_{\infty}$ error across all benchmarks.}

Notably, GNS emerges as the most robust and accurate framework across all PDEs, even in low-data regimes. Its superior performance is attributed to the explicit modeling of spatial interactions and local dynamics through graph-based node and edge feature representations, allowing it to naturally capture complex, nonlinear, and coupled phenomena such as viscous dissipation, nonlinear cross-advection, interfacial evolution, and wave-height coupling with gravity-driven free-surface flow. This inductive bias enables GNS to learn physically consistent representations, making it particularly effective for long-horizon rollouts and low-data scenarios. \rev{The physics-based diagnostics reinforce this conclusion: GNS maintains the lowest pointwise $L_\infty$ error across every benchmark and reproduces the energy spectrum at least as faithfully as the neural operator baselines, indicating that its accuracy is distributed across spatial scales rather than confined to the large-scale structures that dominate the $L_2$ metric.} \rev{Importantly, the flow past cylinder benchmark extends these findings from uniform structured grids to unstructured, variable-geometry meshes: on this problem, GNS outperforms the geometry-aware GINO baseline on the dynamically evolving velocity fields and better respects the divergence-free constraint of the incompressible flow, while GINO retains a modest edge only on the pressure, a diagnostic field governed by a global elliptic equation. This confirms that the advantages of GNS stem from graph-based locality rather than from any particular discretization, and that they persist in the irregular-geometry setting for which the framework is designed.} \rev{We further note that GNS attains this performance as the most parameter-efficient framework considered (see Sec.~\ref{sec:computational_costs}), albeit at a higher per-iteration and inference cost, which we identify as a direction for future improvement.} Overall, our study highlights the strengths and limitations of current neural operator approaches, demonstrating that while kernel integral operators such as FNO, which operate in spectral space, are highly effective for certain PDEs, graph-based architectures like GNS offer a more universal, physics-aware framework for learning spatiotemporally complex dynamics, particularly in low-data regimes.

\section{Limitations and Future Work}
\label{sec:future_work}
While GNS emerged as the superior method in this study in terms of generalization performance for time-dependent PDEs because of its ability to learn the underlying dynamics through stronger physics-aware inductive biases, training such models comes with its own set of challenges. First, training a GNS (or more generally, a GNN) often incurs higher computational costs compared to operator learning frameworks. \rev{Specifically, while FNO benefits from $\mathcal{O}(N \log N)$ complexity per layer via the Fast Fourier Transform, GNS message-passing scales as $\mathcal{O}(Nk)$ per layer, where $N$ is the number of nodes and $k$ is the neighborhood size.} Furthermore, the inherent graph-based data structures required for representing inputs and outputs can impose significant memory demands, which may necessitate smaller batch sizes during training, especially under limited GPU capacity. This issue becomes more pronounced as the memory requirement scales with the number of nodal points, making highly resolved spatial domains quickly prohibitive. \rev{For high-resolution 3D CFD meshes (e.g., $N > 10^7$), the memory footprint for storing graph adjacency lists and edge features renders vanilla GNS frameworks impractical without additional strategies.} In this study, all GNS training was conducted on a multi-node, multi-GPU setup using PyTorch DDP~\cite{li2020pytorch}, underscoring the importance of distributed training for scaling to large domains. \rev{It is important to clarify that when we refer to GNS as ``scalable'' in this work, we specifically mean \emph{data scalability}, i.e., the ability to achieve accurate generalization with significantly fewer training trajectories. Our results demonstrate that GNS achieves less than 1\% relative $L_2$ error using only approximately 3\% of available trajectories, substantially outperforming neural operator baselines in this low-data regime. This data efficiency translates to reduced overall data generation and curation burden, which is a critical consideration when training data must be obtained from expensive high-fidelity simulations.} 

\rev{In this study, the GNS framework was applied to both uniform structured grids and unstructured meshes, the latter demonstrated through the 2D flow past cylinder benchmark featuring irregular triangular meshes with variable topology across trajectories. This validates the flexibility of GNS in handling non-Euclidean domains.} A natural direction for future research lies in mesh-based simulations where the mesh itself deforms at every timestep, necessitating the use of temporal graph networks~\cite{rossi2020temporal}. Beyond this, efforts could also focus on improving the scalability and efficiency of GNS-based approaches, for instance, through memory-efficient graph construction strategies, hierarchical or multiscale message-passing schemes~\cite{fortunato2022multiscale}, or hybrid architectures that combine the efficiency of neural operators with the physics-awareness of GNS, leading to more data-efficient models. \rev{One promising approach for addressing resolution scalability is graph partitioning, as employed in frameworks such as XMeshGraphNet~\cite{nabian2024x}, which decomposes large meshes into smaller subgraphs using algorithms like METIS, performs message-passing over partitions independently, synchronizes information through halo node extensions, and aggregates gradients across partitions. Such strategies enable GNS-based methods to scale to industrial-scale meshes while preserving their physics-aware inductive biases.} Furthermore, embedding physics-informed constraints more seamlessly within the GNS framework presents an exciting opportunity to enhance robustness and broaden its applicability to complex, real-world scientific problems.

\section*{CRediT authorship contribution statement}
\textbf{Dibyajyoti Nayak:} Conceptualization, Methodology, Software, Validation, Formal analysis, Investigation, Data Curation, Writing -- Original Draft, Writing -- Review \& Editing, Visualization; \textbf{Somdatta Goswami:} Conceptualization, Methodology, Validation, Formal analysis, Resources, Data Curation, Writing -- Review \& Editing, Visualization, Supervision, Project administration, Funding acquisition

\section*{Acknowledgments}

The authors' research efforts were partly supported by the National Science Foundation (NSF) under Grant No. 2438193 and 2436738. The authors would like to acknowledge computing support provided by the Advanced Research Computing at Hopkins (ARCH) core facility at Johns Hopkins University and the Rockfish cluster. ARCH core facility (\url{rockfish.jhu.edu}) is supported by the NSF grant number OAC1920103. Any opinions, findings, conclusions, or recommendations expressed in this material are those of the author(s) and do not necessarily reflect the views of the funding organizations.

\bibliographystyle{unsrt}
\bibliography{references}

\newpage
\renewcommand{\thetable}{A\arabic{table}}  
\renewcommand{\thefigure}{A\arabic{figure}} 
\makeatother
\setcounter{figure}{0}
\setcounter{table}{0}
\setcounter{section}{1}
\setcounter{page}{1}
\appendix
\section{\rev{Time-integrator embedded Deep Operator Networks}}
\label{sec:TI-based-NOs_details}

\subsection{\rev{TI-DeepONet (TI-DON)}}
\label{subsec:ti-don}
{\color{black}{
Standard operator-learning strategies for time-dependent PDEs fall into two categories: full-rollout models, which map an initial condition directly to the entire spatiotemporal solution and thereby ignore temporal causality, and autoregressive models, which map the state at time $t$ to the state at $t+\Delta t$ and roll this forward, accumulating error at every step. TI-DeepONet instead reformulates the learning target: rather than predicting the next state $u^{i+1}$ directly, the DeepONet learns the right-hand side of the governing PDE, i.e. the instantaneous time derivative
\begin{equation}
u_t = \mathcal{F}(t,\mathbf{x},u,u_{\mathbf{x}},u_{\mathbf{xx}},\dots).
\end{equation}
This learned derivative is then advanced in time by a classical numerical integrator embedded within the network. Concretely, the current state $u^i$ is passed to the DeepONet, whose output is taken to be the estimated derivative $\mathcal{F}(t^i,u^i)$. Using the fourth-order Runge-Kutta (RK4) scheme with timestep $\Delta t$, the intermediate slopes are computed as
\begin{align}
k_1 &= \mathcal{F}(t^i,u^i), \\
k_2 &= \mathcal{F}\left(t^i+\tfrac{\Delta t}{2},\,u^i+\tfrac{\Delta t}{2}k_1\right), \\
k_3 &= \mathcal{F}\left(t^i+\tfrac{\Delta t}{2},\,u^i+\tfrac{\Delta t}{2}k_2\right), \\
k_4 &= \mathcal{F}\left(t^i+\Delta t,\,u^i+\Delta t\,k_3\right),
\end{align}
and the next state is obtained by
\begin{equation}
\hat{u}^{i+1} = u^i + \Delta t\left(\tfrac{1}{6}k_1 + \tfrac{2}{6}k_2 + \tfrac{2}{6}k_3 + \tfrac{1}{6}k_4\right).
\label{eq:rk4-update}
\end{equation}
Because the integrator is differentiable, the entire pipeline is trained end-to-end: the loss is computed between the predicted state $\hat{u}^{i+1}$ and the ground truth $u^{i+1}$, and gradients are backpropagated through the integrator to update the operator parameters. The loss is summed over all $N = N_s \times N_t$ state pairs, where $N_s$ is the number of trajectories and $N_t$ the number of timesteps. This formulation preserves the Markovian, causal structure of the dynamics while reducing the error accumulation that is inherent to purely autoregressive rollouts.}}

\subsection{\rev{TI(L)-DeepONet (TI(L)-DON)}}
\label{subsec:til-don}
{\color{black}{According to the classical RK4 numerical scheme, the weights of the intermediate slopes $\{\tfrac{1}{6},\tfrac{2}{6},\tfrac{2}{6},\tfrac{1}{6}\}$ in Eq.~\eqref{eq:rk4-update} are fixed and are derived under the assumption of exact slope evaluations. Since here the slopes $k_j$ are produced by a neural approximation of the derivative and therefore carry approximation error, the optimal combination weights need not equal the classical values. TI(L)-DeepONet makes these weights learnable and state-dependent: an auxiliary network $\mathrm{NN}_{RK}$ maps the current state to four coefficients
\begin{equation}
\boldsymbol{\alpha} = [\alpha_1,\alpha_2,\alpha_3,\alpha_4] = \mathrm{NN}_{RK}(u^i),
\end{equation}
and the update becomes
\begin{equation}
u^{i+1} = u^i + \Delta t\,(\alpha_1 k_1 + \alpha_2 k_2 + \alpha_3 k_3 + \alpha_4 k_4).
\end{equation}
The coefficients are normalized with a softmax,
\begin{equation}
\tilde{\alpha}_j = \frac{\exp(\alpha_j)}{\sum_{\ell=1}^{4}\exp(\alpha_\ell)}, \qquad j \in \lbrace 1,2,3,4\rbrace,
\end{equation}
so that they remain positive and sum to one. Because $\boldsymbol{\alpha}$ depends on the evolving state $u^i$, the effective integration weights adapt at each timestep to the local solution behavior, allowing the scheme to compensate for the operator's approximation error, placing greater weight on later-stage slopes in regions of rapid change and reverting toward near-classical weighting in smoother regions.}}

\section{Computational Costs}
\label{sec:computational_costs}
\begin{table}[htbp!]
     \renewcommand{\arraystretch}{1.12}
    \centering
    \caption{Computational costs of training and inference for all examples across different frameworks. \rev{Training for all the cases was carried out on an NVIDIA A100 GPU with 80 GB memory, paired with an Intel Xeon Gold Ice Lake CPU. Note that the training for GNS was done using PyTorch DDP in a multiGPU setup with 2 GPUs. Therefore, the effective batch size for the GNS is batch size per GPU times the number of GPUs (i.e., 2).}}
    \begin{tabular}{|c|c|c|c|c|c|c|}
        \hline
        Problem & Method & \rev{$\#$Params} & \makecell{Batch \\ size} & \makecell{Training time\\(iter/sec)} & \textbf{$N_{test}$} & \makecell{Inference time \\(sec)} \\
        \hline
        \multirow{7}{*}{\shortstack[l]{2D Burgers' \\ scalar}} 
        & GNS   & \rev{168,198} & 4 & 72.87 & \multirow{7}{*}{\rev{200}} & \rev{40.741} \\
        & FNO FR & \rev{134,222,145} & 64  & 7.39 & & \rev{1.708} \\
        & FNO AR & \rev{4,198,689} & 64 & 110.13 & & \rev{1.803} \\
        & TI(L)-DON  & \rev{1,311,788} & 32 & 165.53 & & \rev{6.043} \\
        & TI-DON   & \rev{1,240,584} & 32 & 202.76 & & \rev{5.779} \\
        & DON FR   & \rev{1,240,712} & 32 & 91.35 & & \rev{1.230} \\
        & DON AR   & \rev{1,240,584} & 128 & 305.95 & & \rev{4.923} \\
        \hline
        \multirow{7}{*}{\shortstack[l]{2D Burgers' \\ vector}} 
        & GNS   & \rev{169,626}  & 4 & 45.63 & \multirow{7}{*}{\rev{100}}  & \rev{42.163} \\
        & FNO FR & \rev{134,222,210} & 64  & 2.37 & & \rev{2.384} \\
        & FNO AR & \rev{4,198,754} & 64 & 35.12 & & \rev{2.162} \\
        & TI(L)-DON & \rev{9,308,248} & 32 & 30.15 & & \rev{12.389} \\
        & TI-DON   & \rev{8,772,624} & 32 & 33.54 & & \rev{10.588} \\
        & DON FR   & \rev{8,772,880} & 32 & 21.34 & & \rev{1.569} \\
        & DON AR   & \rev{8,772,624} & 32 & 130.82 & & \rev{9.377} \\
        \hline
        \multirow{7}{*}{\shortstack[l]{2D Allen \\ Cahn}} 
        & GNS   & \rev{168,198} & 4 & 72.94 & \multirow{7}{*}{\rev{200}} & \rev{37.038} \\
        & FNO FR  & \rev{134,222,145} & 64  & 7.35 & & \rev{1.813} \\
        & FNO AR  & \rev{4,198,689} & 64 & 92.57 & & \rev{1.858} \\
        & TI(L)-DON  & \rev{1,311,788} & 64 & 127.26 & & \rev{5.502} \\
        & TI-DON   & \rev{1,240,584} & 128 & 123.05 & & \rev{6.779} \\
        & DON FR  & \rev{1,240,712} & 128 & 91.49 & & \rev{1.235} \\
        & DON AR & \rev{1,240,584} & 128 & 295.61 & & \rev{4.632} \\
        \hline
        \multirow{3}{*}{\shortstack[l]{2D nonlinear \\ SWE}} 
        & GNS   & \rev{171,054} & 2 & 45.79 & \multirow{3}{*}{\rev{100}} & \rev{67.712} \\
        & FNO FR  & \rev{805,331,971} & 16  & 1.88 & & \rev{4.554} \\
        & FNO AR  & \rev{25,191,363} & 16 & 14.79 & & \rev{3.943} \\
        \hline
        \multirow{2}{*}{\shortstack[l]{\rev{2D flow past} \\ \rev{cylinder}}} 
        & \rev{GNS}   & \rev{2,282,883} & \rev{128} & \rev{2.15} & \multirow{2}{*}{\rev{20}} & \rev{115.447} \\
        & \rev{GINO TI}  & \rev{4,877,957} & \rev{128}  & \rev{7.22} & & \rev{27.178} \\
        \hline
        
    \end{tabular}
    \label{tab:training-inference-times}
\end{table}

\section{Architecture Details}
\label{sec:architecture_details}
\subsection{Graph Neural Simulators (GNS)}
\label{subsec:gns_architecture}
\begin{table}[htbp!]
    \renewcommand{\arraystretch}{1.5}
    \centering
    \caption{GNS architecture and training settings across different PDE examples. Here, MP refers to message passing. Each MP layer within the processor block uses LayerNorm. The multilayer perceptrons (MLPs) used within a single MP layer have two hidden layers of 64 neurons each, which is kept consistent across all examples. Finally, a ``mean'' aggregation operation is used in the message aggregation step for all MP layers.}
    \begin{tabular}{|>{\RaggedRight\arraybackslash}p{1.85cm}|>{\centering\arraybackslash}p{2.6cm}|>{\centering\arraybackslash}p{1.25cm}|>{\centering\arraybackslash}p{1.45cm}|>{\centering\arraybackslash}p{1.4cm}|>{\centering\arraybackslash}p{1cm}|>{\centering\arraybackslash}p{1cm}|}
        \hline
        \multirow{2}{*}{Problem} & \multirow{2}{*}{Encoder} & \multirow{2}{*}{Processor} & \multirow{2}{*}{Decoder} & \multirow{2}{*}{Activation} & \multirow{2}{*}{Epochs} & Batch Size\\
        \hline
        2D Burgers' scalar & Node: [7,64,64] Edge: [7,64,64] & 6 MP layers & [64,64,1] & GELU & 600 & 4 \\
        \hline
        2D Burgers' vector & Node: [8,64,64] Edge: [8,64,64] & 6 MP layers & [64,64,2] & GELU & 400 & 4 \\
        \hline
        2D Allen Cahn & Node: [7,64,64] Edge: [7,64,64] & 6 MP layers & [64,64,1] & GELU & 500 & 4 \\
        \hline
        2D nonlinear SWE & Node: [9,64,64] Edge: [9,64,64] & 6 MP layers & [64, 64, 3] & GELU & 600 & 2 \\
        \hline
        \rev{2D flow past cylinder} & \rev{Node: [6,128,128] Edge: [3,128,128]} & \rev{15 MP layers} & \rev{[128,128,3]} & \rev{GELU} & \rev{600} & \rev{128} \\
        \hline
    \end{tabular}
    \label{tab:gns_architecture}
\end{table}

\newpage
\subsection{Fourier Neural Operator (FNO)-based variants}
\label{subsec:fno_architecture}

\begin{table}[htb!]
    \renewcommand{\arraystretch}{1.25}
    \centering
    \caption{FNO architecture and training settings across different PDE examples. Except for the 2D nonlinear SWE case, where 6 Fourier blocks are used, all other cases use 4 Fourier blocks.}
    \begin{tabular}{|>{\RaggedRight\arraybackslash}p{1.8cm}|>{\centering\arraybackslash}p{1.35cm}|>{\centering\arraybackslash}p{1.1cm}|>{\centering\arraybackslash}p{1.3cm}|>{\centering\arraybackslash}p{1.4cm}|>{\centering\arraybackslash}p{1.2cm}|>{\centering\arraybackslash}p{1.25cm}|}
        \hline
        Problem & Method & Modes & Hidden Dim & Activation & Epochs & Batch Size\\
        \hline
        \multirow{2}{*}{\shortstack[l]{2D Burgers' \\ scalar}}
        & FNO AR & 16 & 32 & GELU & 5000 & 64 \\
        \cline{2-7}
        & FNO FR & 16 & 32 & GELU & 5000 & 64 \\
        \hline
        \multirow{2}{*}{\shortstack[l]{2D Burgers' \\ vector}}
        & FNO AR & 16 & 32 & GELU & 2000 & 64 \\
        \cline{2-7}
        & FNO FR & 16 & 32 & GELU & 5000 & 64 \\
        \hline
        \multirow{2}{*}{\shortstack[l]{2D Allen \\ Cahn}} 
        & FNO AR & 16 & 32 & GELU & 4000 & 64 \\
        \cline{2-7}
        & FNO FR & 16 & 32 & GELU & 5000 & 64 \\
        \hline
        \multirow{2}{*}{\shortstack[l]{2D nonlinear \\ SWE}} 
        & FNO AR & 16 & 64 & GELU & 2000 & 16 \\
        \cline{2-7}
        & FNO FR & 16 & 64 & GELU & 6000 & 16 \\
        \hline
    \end{tabular}
    \label{tab:fno_architecture}
\end{table}
\begin{table}[htbp!]
    \renewcommand{\arraystretch}{1.2}
    \centering
    {\color{black}{
    \caption{\rev{GINO architecture and training settings for the 2D flow past cylinder case.}}
    \begin{tabular}{|>{\centering\arraybackslash}p{1.1cm}|>{\centering\arraybackslash}p{1cm}|>{\centering\arraybackslash}p{0.9cm}|>{\centering\arraybackslash}p{1cm}|>{\centering\arraybackslash}p{1.0cm}|>{\centering\arraybackslash}p{0.9cm}|>{\centering\arraybackslash}p{0.8cm}|>{\centering\arraybackslash}p{1.6cm}|>{\centering\arraybackslash}p{1.8cm}|}
        \hline
        Latent Grid & Modes & Hidden Dim & FNO Layers & GNO Radius & Epochs & Batch Size & Input GNO MLP & Output GNO MLP \\
        \hline
        128$\times$32 & (32,16) & 64 & 4 & 0.1 & 500 & 128 & [128]*3 & [256, 128] \\
        \hline
    \end{tabular}
    }}
    \label{tab:gino_architecture}
\end{table}

\subsection{Deep Operator Network (DeepONet)-based variants}
\label{subsec:deeponet_architecture}
\begin{table}[htb!]
    \renewcommand{\arraystretch}{1.15}
    \centering
    \caption{2D Burgers' scalar: DeepONet architecture and training settings across different methods.}
    \begin{tabular}{|
        >{\RaggedRight\arraybackslash}m{1.7cm}|  
        >{\RaggedRight\arraybackslash}m{3.85cm}|  
        >{\RaggedRight\arraybackslash}m{2.25cm}| 
        >{\centering\arraybackslash}m{1.3cm}|   
        >{\centering\arraybackslash}m{1.5cm}|    
        >{\centering\arraybackslash}m{0.75cm}|   
    }
        \hline
        Method & Branch Net & Trunk Net & Activation & Epochs & Batch Size \\
        \hline
        TI(L)-DON 
        & Conv2D(64, (3,3), 1), MaxPool((2,2), (2,2)), Conv2D(64, (2,2), 1), AvgPool((2,2), (2,2)), MLP([256, 128, 128, 100]) 
        & [128]*7 + [100] & Conv2D: GELU, MLP: Tanh & $2.25\times10^5$ & 32 \\
        \hline
        TI-DON 
        & Conv2D(64, (3,3), 1), MaxPool((2,2), (2,2)), Conv2D(64, (2,2), 1), AvgPool((2,2), (2,2)), MLP([256, 128, 128, 100]) 
        & [128]*7 + [100] & Conv2D: GELU, MLP: Tanh & $2\times10^5$ & 32 \\
        \hline
        DON FR 
        & Conv2D(64, (3,3), 1), MaxPool((2,2), (2,2)), Conv2D(64, (2,2), 1), AvgPool((2,2), (2,2)), MLP([256, 128, 128, 100]) 
        & [128]*7 + [100] & Conv2D: GELU, MLP: SiLU & $2\times10^5$ & 32 \\
        \hline
        DON AR
        & Conv2D(64, (3,3), 1), MaxPool((2,2), (2,2)), Conv2D(64, (2,2), 1), AvgPool((2,2), (2,2)), MLP([256, 128, 128, 100]) 
        & [128]*7 + [100] & Conv2D: GELU, MLP: Tanh & $2\times10^5$ & 128 \\
        \hline
    \end{tabular}
    \label{tab:2d_burgers_scalar_nn_architecture}
\end{table}

\begin{table}[htb!]
    \renewcommand{\arraystretch}{1.15}
    \centering
    \caption{2D Burgers' vector: DeepONet architecture and training settings across different methods.}
    \begin{tabular}{|
        >{\RaggedRight\arraybackslash}m{1.7cm}|  
        >{\RaggedRight\arraybackslash}m{3.85cm}|  
        >{\RaggedRight\arraybackslash}m{2.25cm}| 
        >{\centering\arraybackslash}m{1.3cm}|   
        >{\centering\arraybackslash}m{1.5cm}|    
        >{\centering\arraybackslash}m{0.75cm}|   
    }
        \hline
        Method & Branch Net & Trunk Net & Activation & Epochs & Batch Size \\
        \hline
        TI(L)-DON 
        & Conv2D(64, (3,3), 1), MaxPool((2,2), (2,2)), Conv2D(64, (2,2), 1), AvgPool((2,2), (2,2)), MLP([256, 128, 128, 100]) 
        & [128]*7 + [100] & Conv2D: GELU, MLP: Tanh & $1.75\times10^5$ & 32 \\
        \hline
        TI-DON 
        & Conv2D(64, (3,3), 1), MaxPool((2,2), (2,2)), Conv2D(64, (2,2), 1), AvgPool((2,2), (2,2)), MLP([256, 128, 128, 100]) 
        & [128]*7 + [100] & Conv2D: GELU, MLP: Tanh & $1.5\times10^5$ & 32 \\
        \hline
        DON FR 
        & Conv2D(64, (3,3), 1), MaxPool((2,2), (2,2)), Conv2D(64, (2,2), 1), AvgPool((2,2), (2,2)), MLP([256, 128, 128, 100]) 
        & [128]*7 + [100] & Conv2D: GELU, MLP: SiLU & $1.5\times10^5$ & 32 \\
        \hline
        DON AR
        & Conv2D(64, (3,3), 1), MaxPool((2,2), (2,2)), Conv2D(64, (2,2), 1), AvgPool((2,2), (2,2)), MLP([256, 128, 128, 100]) 
        & [128]*7 + [100] & Conv2D: GELU, MLP: Tanh & $1.5\times10^5$ & 32 \\
        \hline
    \end{tabular}
    \label{tab:2d_burgers_vector_nn_architecture}
\end{table}
\begin{table}[h!]
    \renewcommand{\arraystretch}{1.15}
    \centering
    \caption{2D Allen--Cahn equation: DeepONet architecture and training settings across different methods.}
    \begin{tabular}{|
        >{\RaggedRight\arraybackslash}m{1.7cm}|  
        >{\RaggedRight\arraybackslash}m{3.85cm}|  
        >{\RaggedRight\arraybackslash}m{2.25cm}| 
        >{\centering\arraybackslash}m{1.3cm}|   
        >{\centering\arraybackslash}m{1.5cm}|    
        >{\centering\arraybackslash}m{0.75cm}|   
    }
        \hline
        Method & Branch Net & Trunk Net & Activation & Epochs & Batch Size \\
        \hline
        TI(L)-DON
        & Conv2D(64, (3,3), 1), MaxPool((2,2), (2,2)), Conv2D(64, (2,2), 1), AvgPool((2,2), (2,2)), MLP([256, 128, 128, 100]) 
        & [128]*7 + [100] & Conv2D: GELU, MLP: Tanh & $2.25\times10^5$ & 64 \\
        \hline
        TI-DON 
        & Conv2D(64, (3,3), 1), MaxPool((2,2), (2,2)), Conv2D(64, (2,2), 1), AvgPool((2,2), (2,2)), MLP([256, 128, 128, 100]) 
        & [128]*7 + [100] & Conv2D: GELU, MLP: Tanh & $2\times10^5$ & 128 \\
        \hline
        DON FR 
        & Conv2D(64, (3,3), 1), MaxPool((2,2), (2,2)), Conv2D(64, (2,2), 1), AvgPool((2,2), (2,2)), MLP([256, 128, 128, 100]) 
        & [128]*7 + [100] & Conv2D: GELU, MLP: SiLU & $2\times10^5$ & 128 \\
        \hline
        DON AR
        & Conv2D(64, (3,3), 1), MaxPool((2,2), (2,2)), Conv2D(64, (2,2), 1), AvgPool((2,2), (2,2)), MLP([256, 128, 128, 100]) 
        & [128]*7 + [100] & Conv2D: GELU, MLP: Tanh & $2\times10^5$ & 128 \\
        \hline
    \end{tabular}
    \label{tab:allen_cahn_nn_architecture}
\end{table}

\clearpage
\section{Additional Results}
\label{sec:additional_results}
\subsection{Two-Dimensional Burgers' Equation with Scalar Output Field}
\label{subsec:2d_burgers_scalar_appendix}
\begin{figure}[htbp!]
    \centering
    \includegraphics[width=\linewidth]{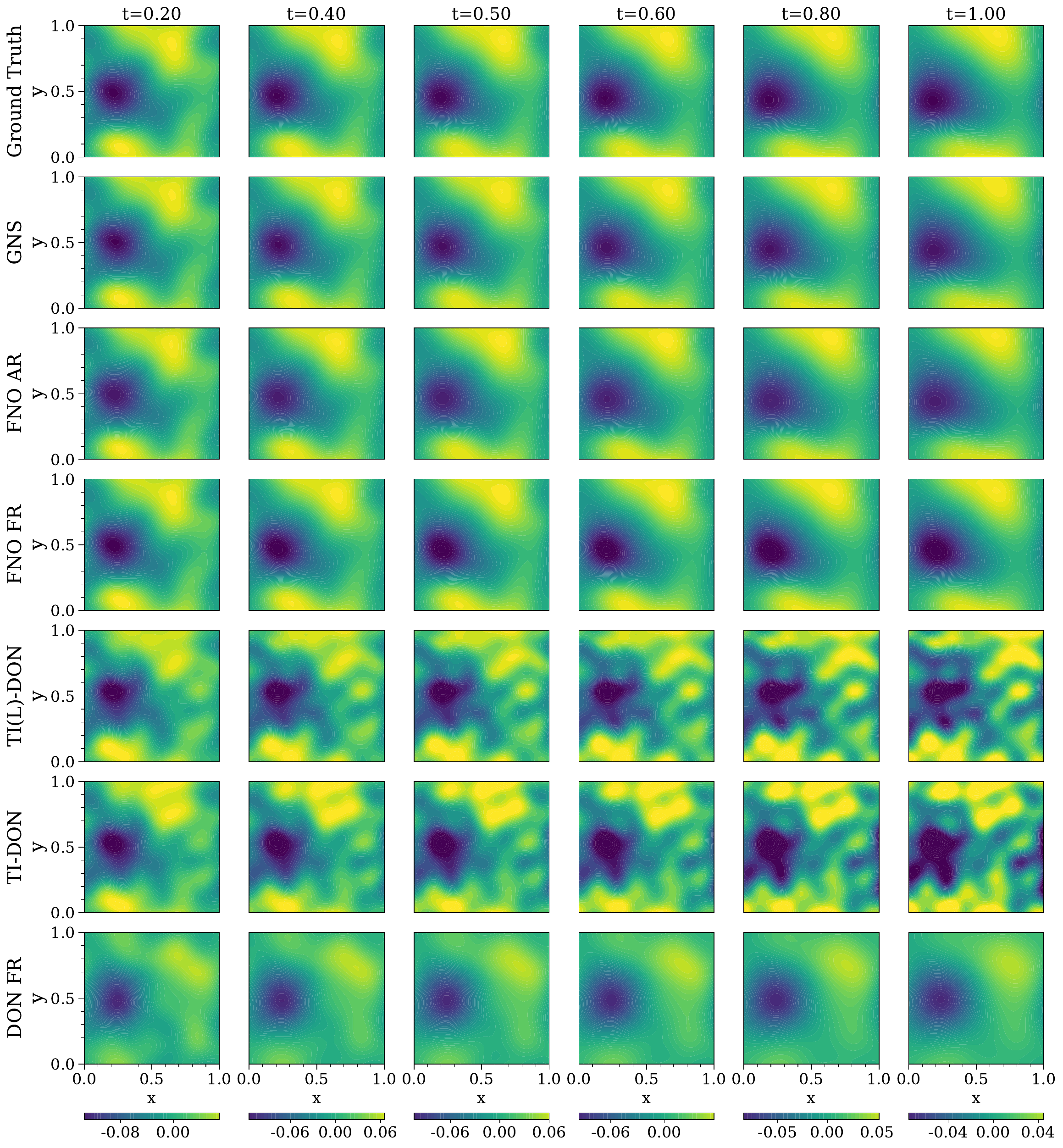}
    \includegraphics[width=\linewidth]{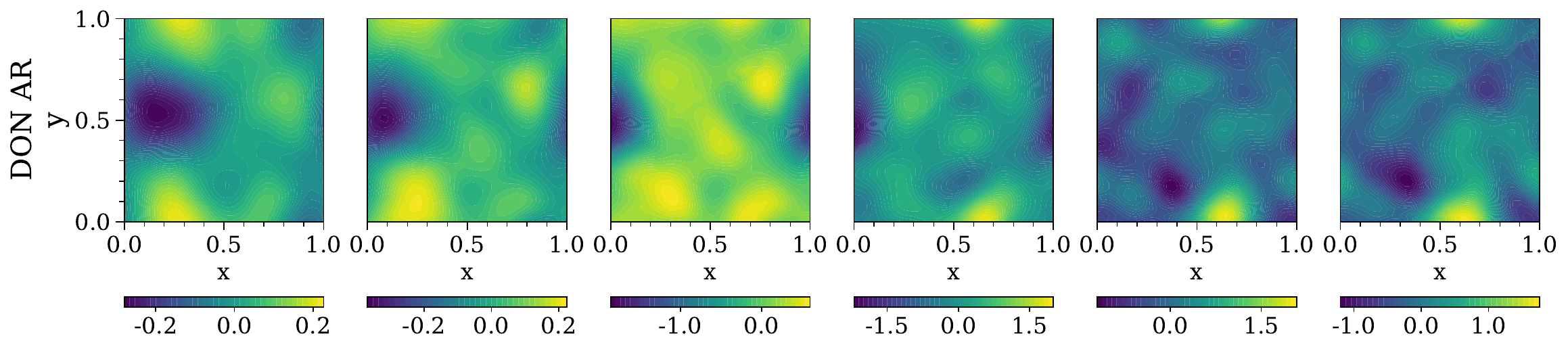}
    \caption{\rev{2D Burgers' equation with scalar output field: Predicted solution contours for all frameworks trained on 30 trajectories and evaluated on a held-out set of 200 trajectories, shown for a representative sample.}}
    \label{fig:2D_Burgers_sol_contours}
\end{figure}

\newpage
\subsection{Two-Dimensional Coupled Burgers' Equation with Vector Output Field}
\label{subsec:2d_burgers_vector_appendix}
\begin{figure}[htbp!]
    \centering
    \includegraphics[width=\linewidth]{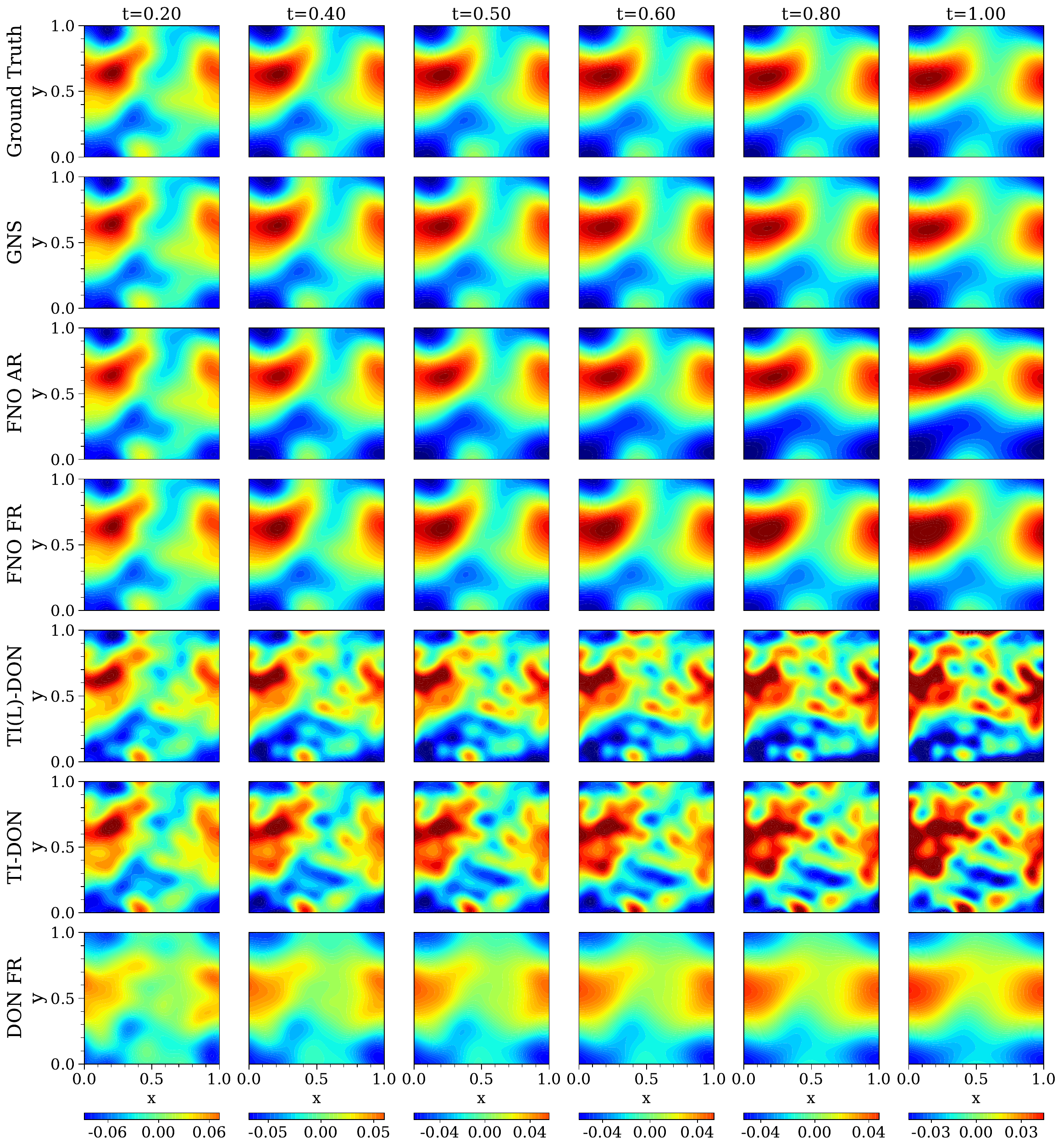}
    \includegraphics[width=\linewidth]{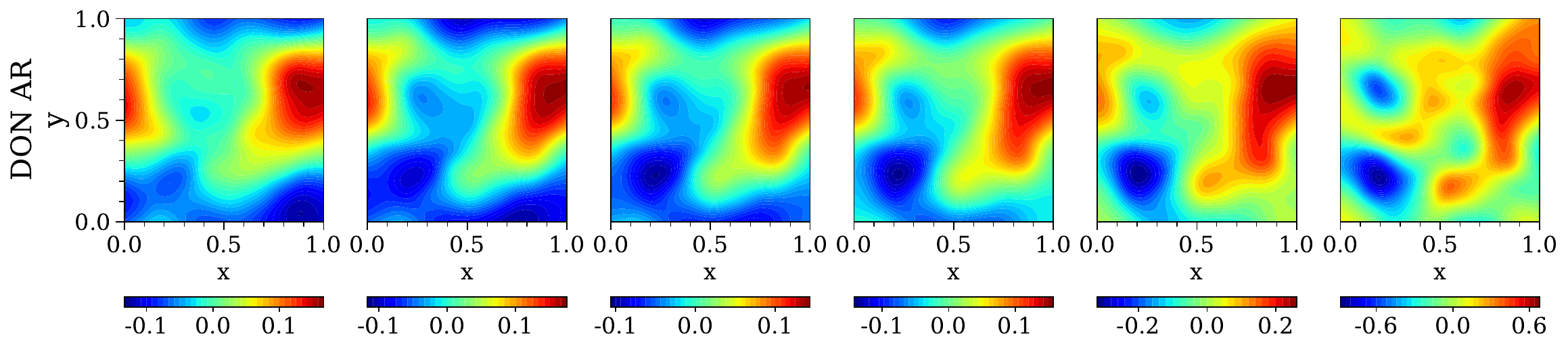}
    \caption{\rev{2D coupled Burgers' equation with vector output field: Predicted solution contours for all frameworks trained on 30 trajectories and evaluated on a held-out set of 100 trajectories for predicting the $u$-velocity field, shown for a representative sample.}}
    \label{fig:2D_coupled_Burgers_sol_contours_u}
\end{figure}
\begin{figure}[htbp!]
    \centering
    \includegraphics[width=\linewidth]{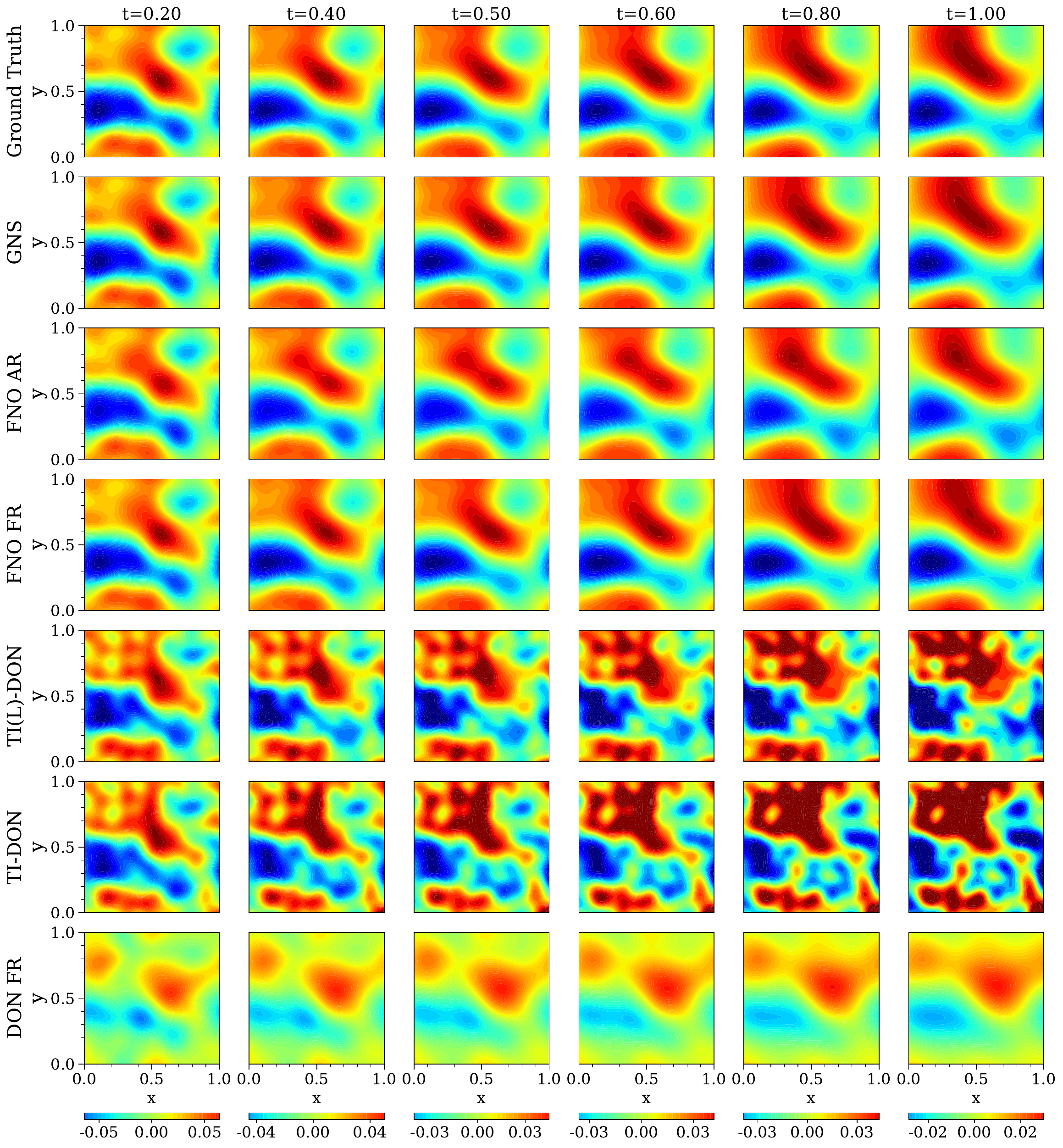}
    \includegraphics[width=\linewidth]{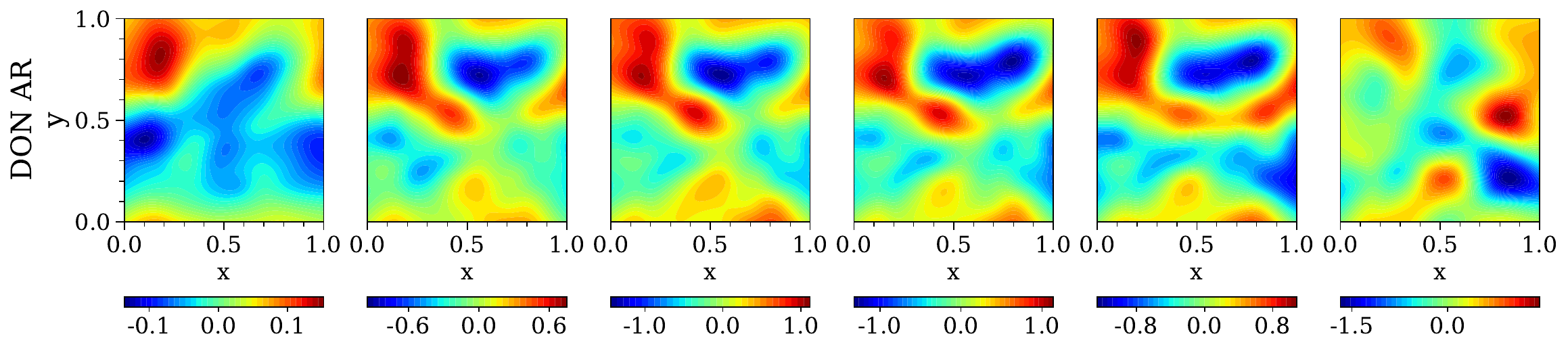}
    \caption{\rev{2D coupled Burgers' equation with vector output field: Predicted solution contours for all frameworks trained on 30 trajectories and evaluated on a held-out set of 100 trajectories for predicting the $v$-velocity field, shown for a representative sample.}}
    \label{fig:2D_coupled_Burgers_sol_contours_v}
\end{figure}

\newpage
\subsection{Two-Dimensional Allen-Cahn Equation}
\label{subsec:2d_allen_cahn_appendix}
\begin{figure}[htbp!]
    \centering
    \includegraphics[width=\linewidth]{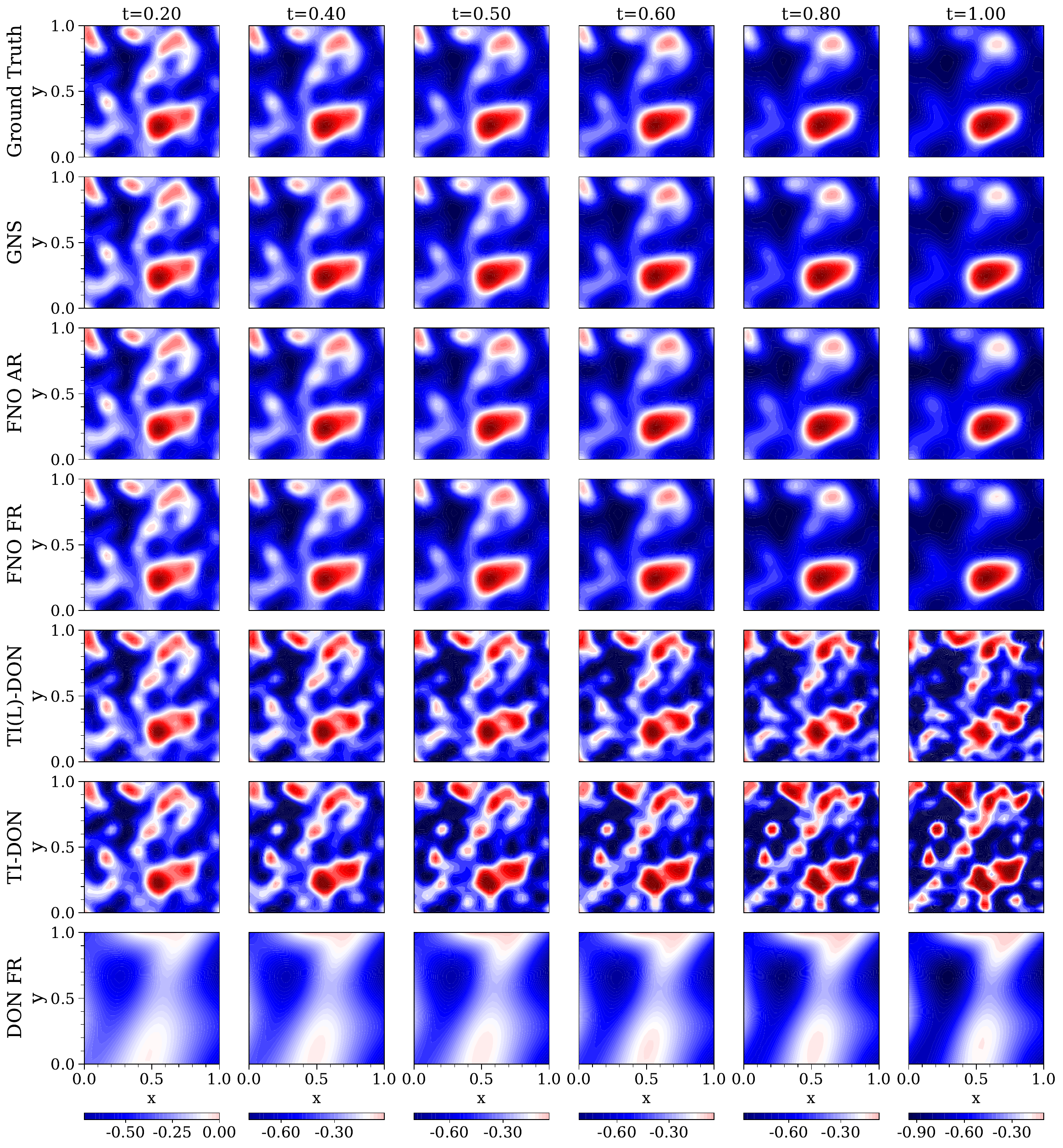}
    \includegraphics[width=\linewidth]{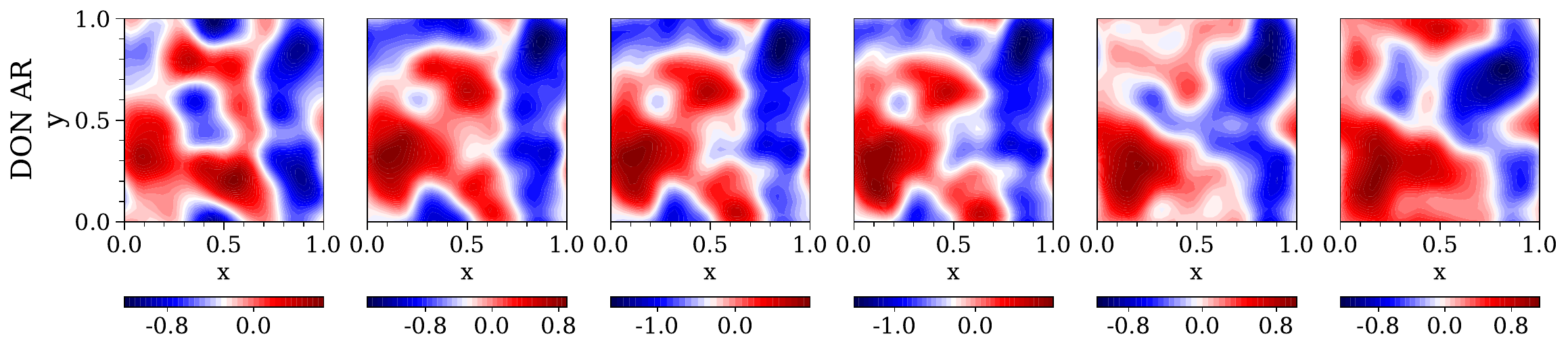}
    \caption{\rev{2D Allen-Cahn Equation: Predicted solution contours for all frameworks trained on 30 trajectories and evaluated on a held-out set of 200 trajectories, shown for a representative sample.}}
   \label{fig:2D_Allen_Cahn_sol_contours}
\end{figure}

\newpage
\subsection{Two-Dimensional Nonlinear Shallow Water Equations}
\label{subsec:2d_nonlinear_SWE_appendix}
\begin{figure}[htbp!]
    \centering
    \includegraphics[width=\linewidth]{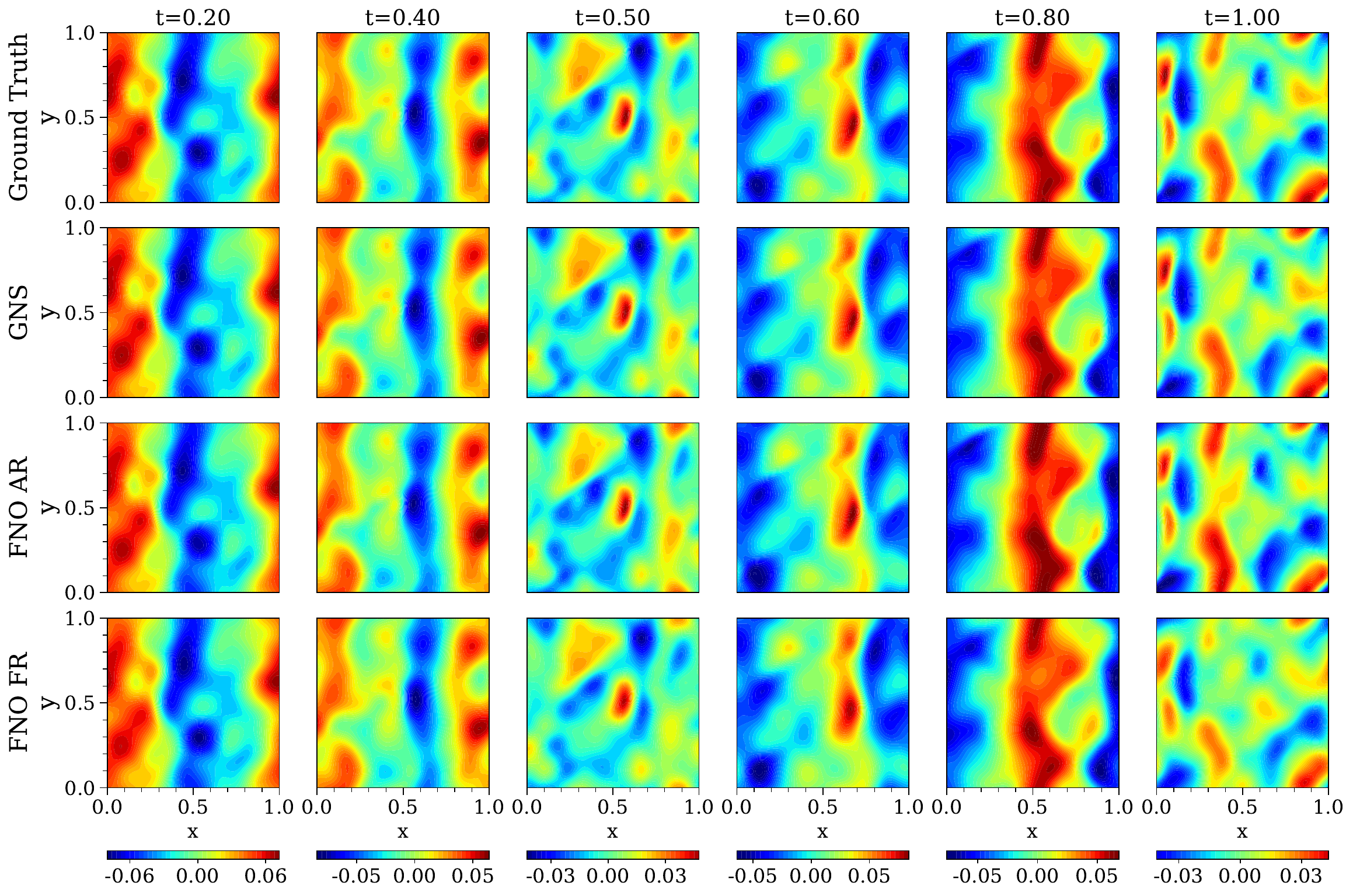}
    \caption{\rev{2D nonlinear shallow water equations: Predicted solution contours for all frameworks trained on 50 trajectories and evaluated on a held-out set of 100 trajectories for predicting the $u$-velocity field, shown for a representative sample.}}
    \label{fig:2D_nonlinear_SWE_sol_contours_u}
\end{figure}
\begin{figure}[htbp!]
    \centering
    \includegraphics[width=\linewidth]{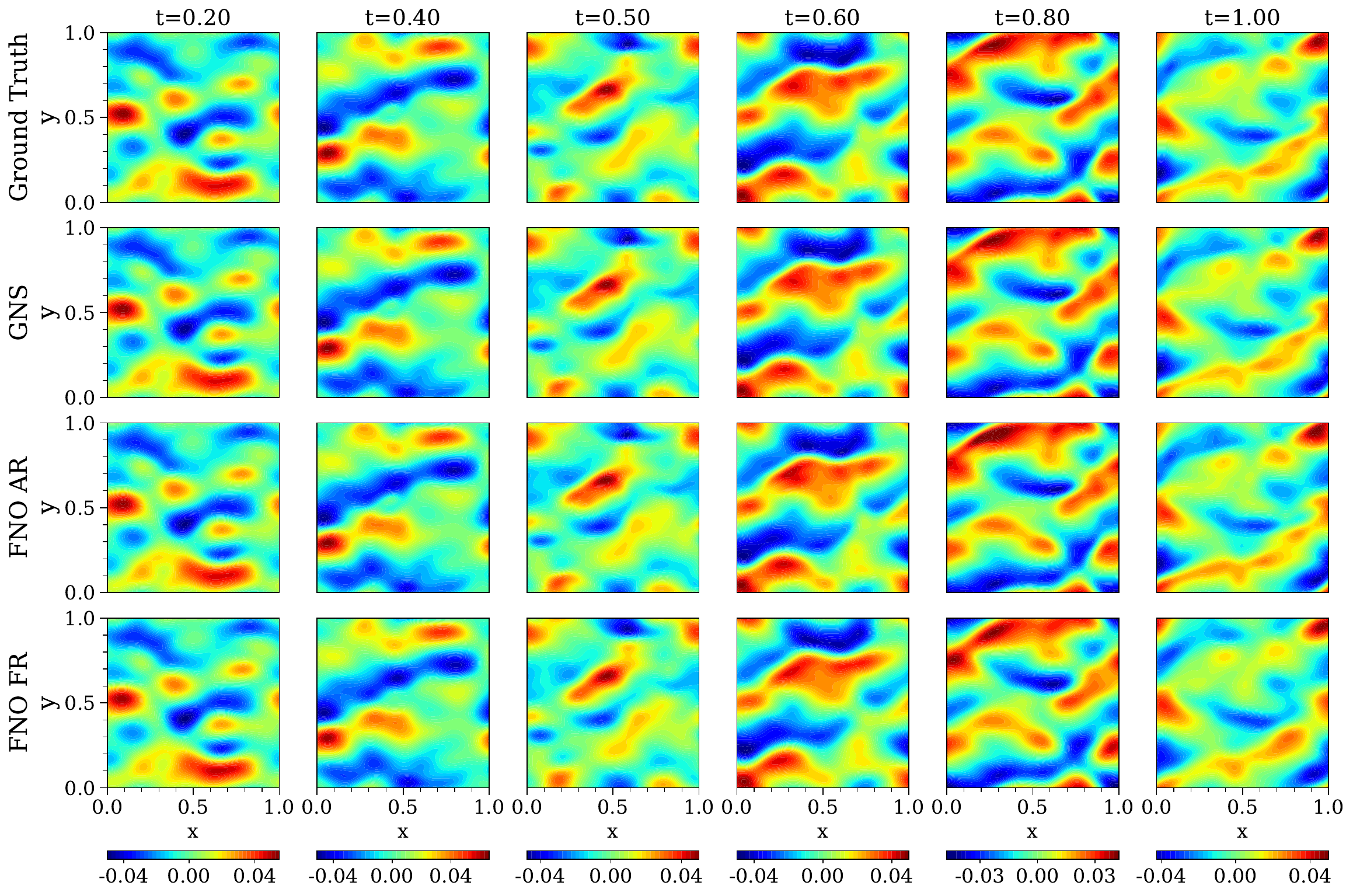}
    \caption{\rev{2D nonlinear shallow water equations: Predicted solution contours for all frameworks trained on 50 trajectories and evaluated on a held-out set of 100 trajectories for predicting the $v$-velocity field, shown for a representative sample.}}
    \label{fig:2D_nonlinear_SWE_sol_contours_v}
\end{figure}
\begin{figure}[htbp!]
    \centering
    \includegraphics[width=\linewidth]{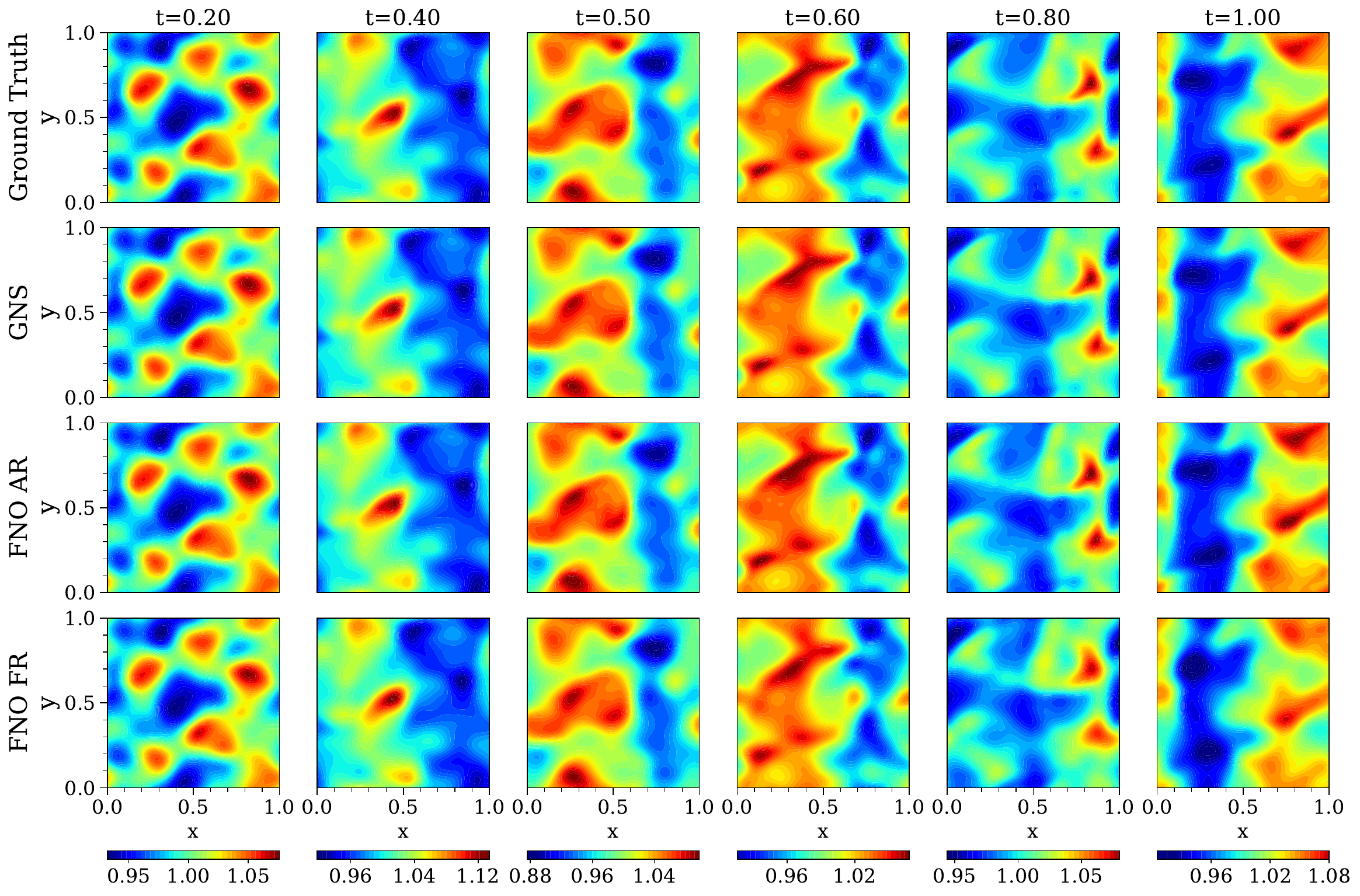}
    \caption{\rev{2D nonlinear shallow water equations: Predicted solution contours for all frameworks trained on 50 trajectories and evaluated on a held-out set of 100 trajectories for predicting the $\eta$-height field, shown for a representative sample.}}
    \label{fig:2D_nonlinear_SWE_sol_contours_h}
\end{figure}

\clearpage
\subsection{\rev{Two-Dimensional Flow Past a Cylinder}}
\label{subsec:cylinderflow_appendix}
\begin{figure}[htbp!]
    \centering
    \includegraphics[width=\linewidth]{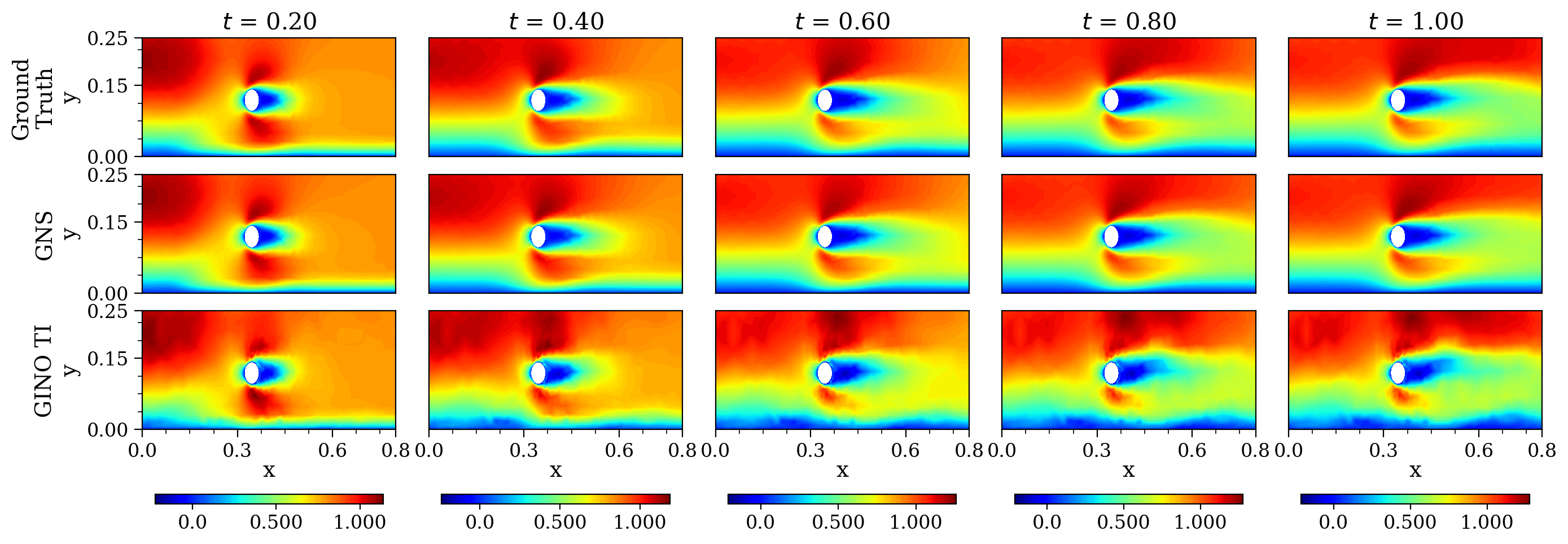}
    \caption{\rev{2D flow past a cylinder: Predicted solution contours for all frameworks trained on 200 trajectories and evaluated on 20 unseen test trajectories for predicting the $u$-velocity field, shown for a representative sample.}}
    \label{fig:2D_flow_past_cylinder_sol_contours_u}
\end{figure}
\begin{figure}[htbp!]
    \centering
    \includegraphics[width=\linewidth]{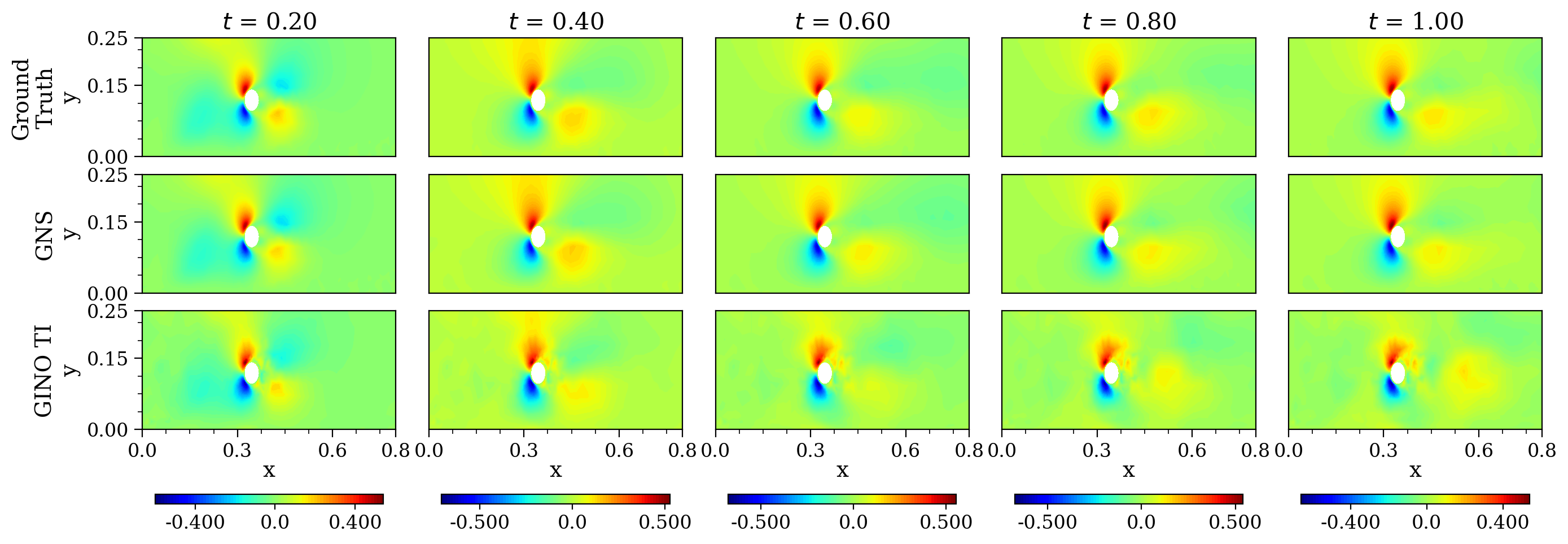}
    \caption{\rev{2D flow past a cylinder: Predicted solution contours for all frameworks trained on 200 trajectories and evaluated on 20 unseen test trajectories for predicting the $v$-velocity field, shown for a representative sample.}}
    \label{fig:2D_flow_past_cylinder_sol_contours_v}
\end{figure}
\begin{figure}[htbp!]
    \centering
    \includegraphics[width=\linewidth]{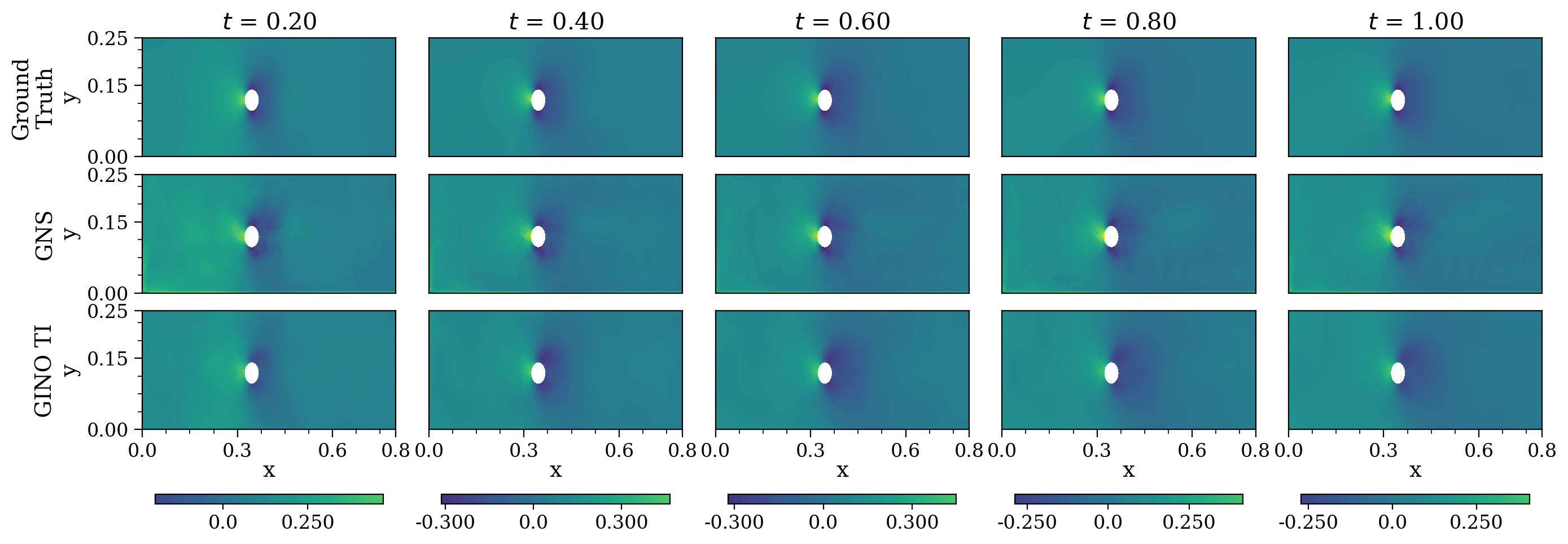}
    \caption{\rev{2D flow past a cylinder: Predicted solution contours for all frameworks trained on 200 trajectories and evaluated on 20 unseen test trajectories for predicting the pressure field $p$, shown for a representative sample.}}
    \label{fig:2D_flow_past_cylinder_sol_contours_p}
\end{figure}

\end{document}